\documentclass{article}

\PassOptionsToPackage{numbers, compress}{natbib}
\usepackage[final]{neurips_2025}
\usepackage[utf8]{inputenc}
\usepackage[T1]{fontenc}
\usepackage{hyperref}
\usepackage{url}
\usepackage{booktabs}
\usepackage{pgfplots}
\usepackage{amsfonts}
\usepackage{mathtools}
\usepackage{nicefrac}
\usepackage{microtype}
\usepackage{xcolor}
\usepackage{graphicx}
\usepackage{amsmath}
\usepackage{amssymb}
\usepackage{enumitem}
\usepackage{pifont}
\usepackage{colortbl, multirow}
\usepackage{pgfplots}\usepackage{pgfplotstable}
\usepackage{tikz}
\usepackage{algorithm}
\usepackage{algpseudocode}
\usepackage{bm}
\usepackage{caption}
\usepackage{subcaption}     
\usepackage{wrapfig}  
\pgfplotsset{compat=1.18}
\usepackage{soul}
\usepackage{xcolor}
\definecolor{highlight}{RGB}{245,245,200}  
\sethlcolor{highlight}
\usepackage{tablefootnote}

\newcommand{\prevmodel}{DG\textsuperscript{2}CD-Net}
\newcommand{\ourmodel}{\textsc{HiDISC}}

\title{\textsc{HiDISC}: A Hyperbolic Framework for Domain Generalization with Generalized Category Discovery}

\author{
  Vaibhav Rathore\textsuperscript{1} \quad Divyam Gupta\textsuperscript{1} \quad Biplab Banerjee\textsuperscript{1} \\
  Indian Institute of Technology Bombay\textsuperscript{1} \\
  \texttt{\{vaibhav.rathor.in, divs25.iitb, getbiplab\}@gmail.com}
}

\begin{document}

\maketitle

\begin{abstract}

Generalized Category Discovery (GCD) aims to classify test-time samples into either seen categories—available during training—or novel ones, without relying on label supervision. Most existing GCD methods assume simultaneous access to labeled and unlabeled data during training and arising from the same domain, limiting applicability in open-world scenarios involving distribution shifts. Domain Generalization with GCD (DG-GCD) lifts this constraint by requiring models to generalize to unseen domains containing novel categories, without accessing target-domain data during training.
The only prior DG-GCD method, DG$^2$CD-Net~\cite{dg2net}, relies on episodic training with multiple synthetic domains and task vector aggregation, incurring high computational cost and error accumulation. We propose \textsc{HiDISC}, a hyperbolic representation learning framework that achieves domain and category-level generalization without episodic simulation. To expose the model to minimal but diverse domain variations, we augment the source domain using GPT-guided diffusion, avoiding overfitting while maintaining efficiency.
To structure the representation space, we introduce \emph{Tangent CutMix}, a curvature-aware interpolation that synthesizes pseudo-novel samples in tangent space, preserving manifold consistency. A unified loss—combining penalized Busemann alignment, hybrid hyperbolic contrastive regularization, and adaptive outlier repulsion—facilitates compact, semantically structured embeddings. A learnable curvature parameter further adapts the geometry to dataset complexity.
\textsc{HiDISC} achieves state-of-the-art results on PACS~\cite{pacs}, Office-Home~\cite{officehome}, and DomainNet~\cite{domainnet}, consistently outperforming the existing Euclidean and hyperbolic (DG)-GCD baselines.\footnote{Project Page : https://vaibhavrathore1999.github.io/HiDISC/}

\end{abstract}

\section{Introduction}

Deep neural networks have achieved impressive success in visual recognition~\cite{voulodimos2018deep, chai2021deep}, yet typically assume a shared domain and label space between training and test data. This assumption breaks down in real-world applications such as autonomous driving~\cite{sun2022shift} and medical diagnostics~\cite{yoon2023domain}, where both \emph{domain shift} and \emph{label shift} frequently co-occur.
While semi/self-supervised learning~\cite{sem-survey, ssl-survey} reduces labeling demands, it still operates under closed-world constraints. Domain Adaptation (DA) and Domain Generalization (DG)~\cite{da-survey, dg-survey1} address distribution shift but assume a fixed set of categories. Open-set DG~\cite{saito2018open, ODG3} allows test-time novelty but collapses all unknowns into a single rejection class, erasing semantic granularity.

Generalized Category Discovery (GCD)~\cite{gcd, gcd1, gcd2} seeks to identify both known and novel classes from unlabeled test data but requires joint access to labeled and unlabeled samples from the same domain. Cross-Domain GCD (CD-GCD)~\cite{rongali2024cdadnetbridgingdomaingaps, wen2024cross} introduces domain shift but still assumes concurrent access to source--target domains during training. In contrast, \textbf{Domain Generalization with GCD (DG-GCD)}~\cite{dg2net} represents a more realistic setting: \textit{the model is trained solely on labeled source data and must generalize to an unseen target domain containing both seen and novel categories.}

Addressing DG-GCD requires (i) learning domain-invariant features and (ii) discovering novel semantic structures without supervision. The only existing solution tailored for DG-GCD, DG$^2$CD-Net~\cite{dg2net} approaches this via episodic training with synthetic domains and task aggregation, but suffers from high computational cost and cumulative approximation errors that limit generalization.

From a different perspective, existing GCD and DG-GCD methods typically rely on Euclidean or hyperspherical geometry~\cite{dg2net, rongali2024cdadnetbridgingdomaingaps}, which struggle to capture semantic hierarchies. Hyperbolic geometry~\cite{nickel2017poincare, chami2019hyperbolic}, with its negative curvature and exponential volume growth, offers a natural alternative for modeling inter-class structure (Fig. \ref{fig:teaser}). While hyperbolic embeddings have shown benefits in GCD (HypCD)~\cite{Liu2025HypCD} and DG~\cite{ghadimi2021hyperbolic} recently, their use in DG-GCD, where both domain and label shifts co-occur, remains unexplored. This raises our central question:

\begin{wrapfigure}{r}{0.48\textwidth}
\vspace{-12pt}
\centering
\includegraphics[width=\linewidth]{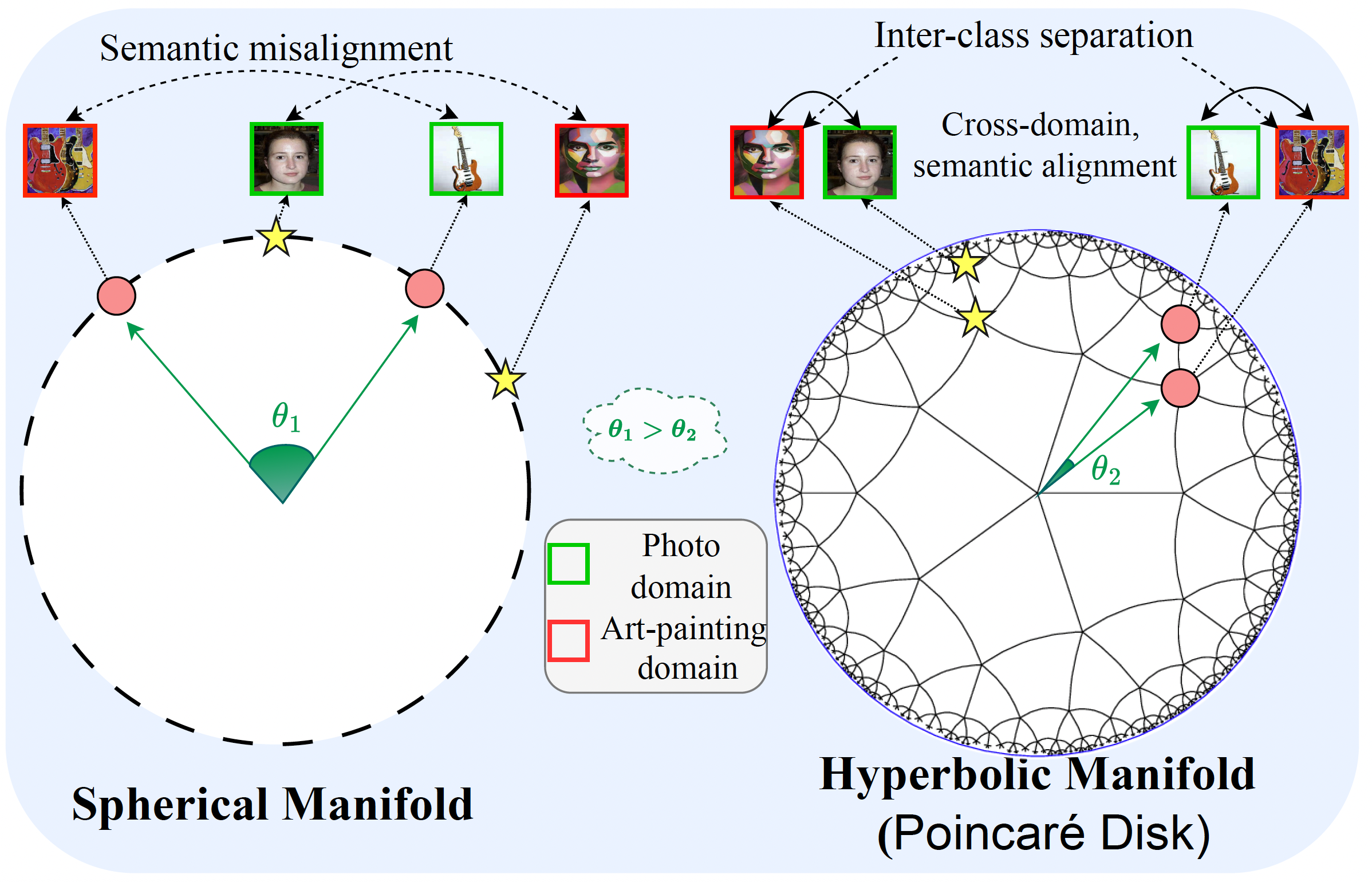}
\caption{
\textbf{Spherical}  vs. \textbf{hyperbolic} (Poincaré) embeddings on PACS. Same-class samples from different domains (green/red) cluster more tightly in hyperbolic space, demonstrating improved class separation. Refer to \textbf{Sup. Mat.} for quantitative analysis.}

\label{fig:teaser}
\vspace{-12pt}
\end{wrapfigure}

\begin{quote}
\emph{Can hyperbolic geometry provide a unified foundation for solving DG-GCD,\\ addressing both distribution shift and novel-class discovery?}
\end{quote}

\noindent\textbf{Our approach.} We introduce \textsc{HiDISC}, the first hyperbolic geometry-aware framework for DG-GCD that learns semantically structured and domain-invariant embeddings in the Poincaré ball~\cite{nickel2017poincare} without requiring target supervision. Unlike HypCD~\cite{Liu2025HypCD}, which addresses standard GCD in single-domain settings, and DG$^2$CD-Net~\cite{dg2net}, which operates in Euclidean space and relies on episodic simulation, \textsc{HiDISC} provides a unified, non-episodic solution using the representational advantages of hyperbolic space.

Since the hyperbolic space offers substantial domain invariance and focuses on shared semantics \cite{ghadimi2021hyperbolic} (Fig. \ref{fig:teaser}, \textbf{Sup. Mat.}), we still chose to augment the training source domain with controlled stylistic variations and generate only 1–2 synthetic domains per image via a GPT-4o~\cite{openai2024chatgpt}-guided diffusion model, avoiding the computational overhead of task-based simulation in DG$^2$CD-Net. A novel \textit{domain-diversity score} ranks these augmentations by measuring source divergence and intra-pair variability, enabling a principled and scalable domain diversification strategy.

To ensure the model does not overfit the known classes and encourage semantic diversity, we propose \emph{Tangent CutMix}, a curvature-aware interpolation method that mixes labeled features in the tangent space of the Poincaré ball to create pseudo-open samples. Unlike Euclidean mixing used in SimGCD~\cite{gcd3} or CMS~\cite{gcd6}, our method preserves hyperbolic consistency and generates geometrically valid pseudo-novel embeddings that support open-set regularization.

To structure the latent space, we introduce a unified loss that combines three novel components not jointly explored in prior (DG)-GCD literature: (i) a penalized Busemann loss that aligns seen-class features to fixed prototypes at the hyperbolic boundary while reserving interior space for unknowns; (ii) a hybrid hyperbolic contrastive loss balancing angular and geodesic similarities to enable fine-grained clustering across known and novel categories; and (iii) an adaptive outlier rejection loss that pushes synthetic cut-mix samples away from known-class regions, encouraging open-space generalization without relying on adversarial or domain-specific objectives.
 A learnable curvature parameter further adapts the geometry to dataset-specific complexity. 
Major contributions include:

\begin{itemize}[leftmargin=1.5em,label=--]
\item \textsc{HiDISC}, the first hyperbolic DG-GCD framework, jointly handles domain and category shift without target supervision or episodic simulation.
\item A unified loss formulation integrating Busemann alignment, hybrid contrastive regularization, and an adaptive outlier repulsion.
\item Tangent CutMix, the first open-set augmentation designed specifically for hyperbolic geometry.
\item State-of-the-art results on PACS~\cite{pacs}, Office-Home~\cite{officehome}, and DomainNet~\cite{domainnet}, outperforming all the baselines consistently and reducing training FLOPs by over $96\times$ vs.\ DG$^2$CD-Net.
\end{itemize}

\section{Related Works}

\textbf{Domain Generalization.}
DG aims to train models on labeled data from one or more source domains to generalize effectively to previously unseen target domains~\cite{dg-survey1, dg-survey2}. DG variants include closed-set, open-set, single-source, and multi-source settings~\cite{wang2024hilolearningframeworkgeneralized}. Methods like MixStyle~\cite{closedDG1} and StyleHallucination~\cite{zhao2022stylehallucinateddualconsistencylearning} enhance robustness via feature-level style perturbations, while meta-learning techniques~\cite{closedDG2, closedDG4} simulate domain shifts episodically to improve adaptability. \textit{Open-set DG methods address novel test-time classes~\cite{ODG1, ODG2, ODG3, ODG4}, but typically collapse all unknowns into a single “outlier” class, hindering fine-grained discovery needed in DG-GCD.}

\textbf{Category Discovery.}
Category Discovery seeks to partition unlabeled data into known and novel categories. While Novel Category Discovery (NCD)~\cite{ncd} assumes complete disjointness between training and test classes, GCD allows overlap and requires identifying both seen and unseen categories during inference~\cite{gcd, gcd1, gcd2, gcd3, gcd4, gcd5, gcd6}. Most GCD approaches assume joint access to source and target domains during training, limiting real-world applicability. CD-GCD methods like CDAD-Net~\cite{rongali2024cdadnetbridgingdomaingaps} and HiLo~\cite{wang2024hilolearningframeworkgeneralized} reduce domain gaps via adversarial alignment or style normalization but still depend on concurrent domain access. To remove this constraint, DG-GCD~\cite{dg2net} simulates domain shifts through text-driven image manipulation (e.g., InstructPix2Pix~\cite{brooks2023instructpix2pix}) and aggregates task-specific knowledge via task vectors~\cite{task_arithmetic}. \textit{However, these methods operate in Euclidean spaces, which struggle to encode the hierarchical and shared semantic structures crucial for robust domain and category generalization.}

\textbf{Hyperbolic Embedding Spaces.}
Hyperbolic geometry, defined by negative curvature and exponential volume growth, is well-suited for modeling hierarchical and part-whole semantic structures~\cite{nickel2017poincare, chami2019hyperbolic}. Hyperbolic embeddings have improved performance in classification~\cite{ermolov2022hyperbolic, guo2022clipped, khrulkov2020hyperbolic}, few-shot learning~\cite{gao2021curvature}, segmentation~\cite{weng2021unsupervised}, and action recognition~\cite{franco2023hyperbolic}, supported by hyperbolic variants of standard network components~\cite{shimizu2020hyperbolic, bdeir2023fully, chami2019hyperbolic, gulcehre2018hyperbolic}. Recent Busemann-based techniques~\cite{ghadimi2021hyperbolic, keller2020theory} anchor ideal prototypes on the Poincaré boundary for directional alignment. HypCD~\cite{Liu2025HypCD} successfully applies this to GCD, but assumes joint source–target access. Beyond these, hyperbolic methods have been applied across diverse tasks: Ge et al.~\cite{ge2023hyperbolic} explore contrastive learning for hierarchical scene–object representation, Yue et al.~\cite{yue2024understanding} study metric learning with hard negatives, Liu et al.~\cite{chang2025multi} extend contrastive learning to EEG, Sun \& Ma~\cite{sun2024hyperbolic} investigate recommendation, while others address hashing~\cite{wei2024exploring} and face anti-spoofing with hierarchical prototypes~\cite{hu2024rethinking}. \textit{To date, no work has explored hyperbolic representations for DG-GCD, which combines open-set discovery and domain shift without target supervision.}

\section{Methodology}
\subsection{The DG-GCD Problem Definition}

In DG-GCD, we are given a labeled source-domain dataset:
\[
D_{S} = \{(x_i^s, y_i^s)\}_{i=1}^{n_s}, \quad x_i^s \in \mathcal{X}_s,\; y_i^s \in \mathcal{Y}_s,
\]
where \(x_i^s\) represents source-domain inputs, and \(y_i^s\) denotes labels drawn from a set of \textit{known} categories \(\mathcal{Y}_s\). At test time, we encounter an unlabeled target-domain dataset:
\[
D_{T} = \{x_j^t\}_{j=1}^{n_t}, \quad x_j^t \in \mathcal{X}_t,
\]
where samples belong either to known categories (\(\mathcal{Y}_t^{\text{old}} = \mathcal{Y}_s\)) or previously unseen, \textit{novel} categories (\(\mathcal{Y}_t^{\text{new}}\)), such that \(\mathcal{Y}_t^{\text{new}} \cap \mathcal{Y}_s = \emptyset\). Crucially, the data distributions across domains differ significantly, i.e., \(P(\mathcal{X}_s) \neq P(\mathcal{X}_t)\), and the target dataset \(D_{T}\) is inaccessible during training.

Our objective is to construct an embedding space using only \(D_{S}\) that generalizes across domains and categories, effectively clustering both known and novel-class samples from the unseen dataset \(D_{T}\).

\subsection{Rationale Behind Using Hyperbolic Space for DG-GCD}
\label{sec:hyperbolic_space}

Semantic structures in visual data—such as hierarchies, taxonomies, and part–whole relationships—are inherently suited to spaces with exponential capacity. Hyperbolic space, characterized by negative curvature and exponential volume growth, naturally encodes such structures, making it particularly beneficial for DG-GCD, where labeled classes typically reflect coarse semantic strata, while novel categories often reside in finer or more abstract regions.

In contrast to Euclidean or spherical embeddings~\cite{gcd1,gcd9}, which are constrained by polynomial growth, hyperbolic embeddings support both local compactness and global semantic separation. 
Moreover, hyperbolic geometry improves domain generalization by amplifying higher-level semantic distances and attenuating domain-specific low-level variations, thereby fostering robust, domain-invariant representations under substantial distributional shifts (more details provided in \textbf{Sup. Mat}).

\paragraph{Poincaré Ball Geometry.}
We adopt the Poincaré ball~\cite{nickel2017poincare} as our hyperbolic model. For curvature \( -c^2 \), the \( n \)-dimensional ball is defined as:
\[
    \mathbb{D}_c^n = \left\{ \mathbf{a} \in \mathbb{R}^n \;\middle|\; c \|\mathbf{a}\|^2 < 1 \right\},
\]
where \( \|\cdot\| \) denotes the Euclidean norm. Additional geometric details are provided in the \textbf{Sup. Mat.}


\begin{figure}
    \centering
    \includegraphics[width=\linewidth]{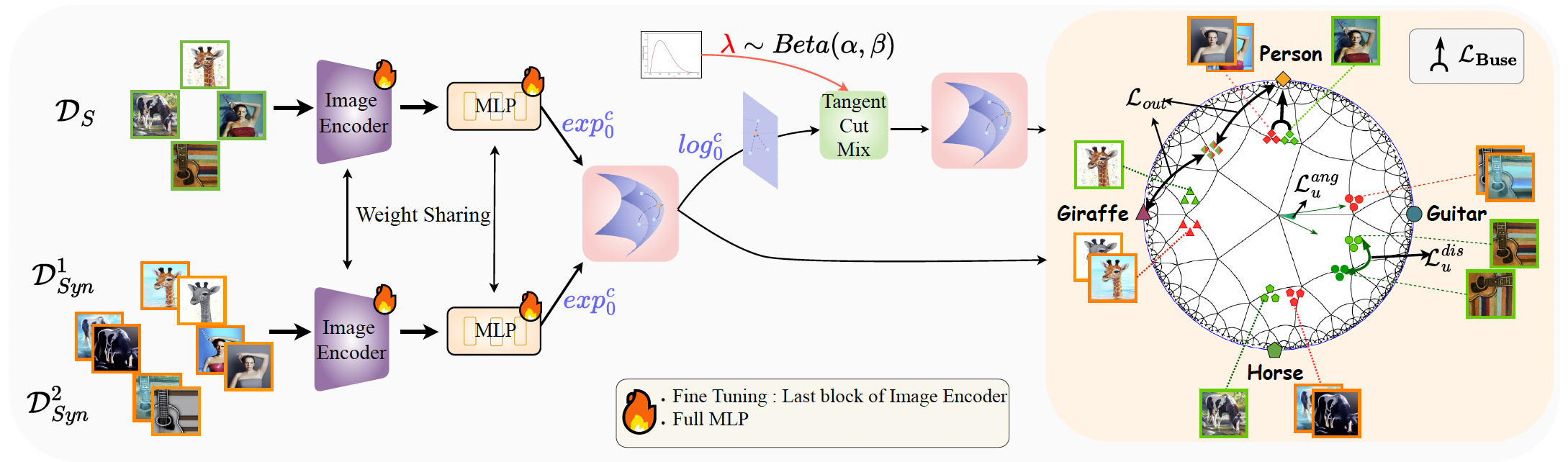}
    \caption{
\textbf{Illustration of the \textsc{HiDISC} pipeline for DG-GCD in hyperbolic space}. The model is trained using labeled source data $\mathcal{D}_S$ (green borders) and 1–2 GPT-guided synthetic domains $\mathcal{D}_{\text{Syn}}^1$, $\mathcal{D}_{\text{Syn}}^2$ (orange borders) to simulate domain shift. Features from the shared encoder are projected to the Poincaré ball via $\exp_0^c$. To mimic novel categories, Tangent CutMix performs interpolation in the tangent space and maps the result $z_{\text{mix}}$ back to hyperbolic space. The embedding space is structured via: (i) penalized Busemann loss $\mathcal{L}_{\text{Buse}}$ for aligning seen classes to boundary-fixed prototypes; (ii) hybrid contrastive loss $\mathcal{L}_u$ for clustering and separability; and (iii) adaptive outlier loss $\mathcal{L}_{\text{out}}$ to repel pseudo-novel points. Together, these shape a curvature-aware space for generalization and discovery.
}
\label{fig:model_arch}
\vspace{-15px}
\end{figure}

\subsection{Navigating through \textsc{HiDISC} for DG-GCD}
\label{sec:methodology}

We propose \textsc{HiDISC} (Fig. \ref{fig:model_arch}), a hyperbolic framework that jointly addresses the dual challenges of DG-GCD: domain-invariant representation learning and unsupervised semantic disentanglement. The synthesis-driven components of our model include: (i) \textit{Synthetic Domain Augmentation}, which introduces a compact set of diverse, diffusion-generated domains to simulate realistic distribution shifts without relying on target access; and (ii) \textit{Tangent CutMix}, a curvature-aware interpolation mechanism operating in the tangent space of the Poincaré ball, generating pseudo-novel samples while preserving manifold fidelity. Complementing these are three loss-driven modules: (iii) \textit{Prototype Anchoring}, which aligns seen-class embeddings to fixed ideal prototypes on the Poincaré boundary, reserving central space for novel classes; (iv) \textit{Adaptive Outlier Loss}, which ensures synthetic samples are repelled from known-class clusters, promoting open-space regularization; and (v) \textit{Hybrid Hyperbolic Contrastive Loss}, which combines geodesic and angular similarity to improve local cohesion and global separability. Each component is described in detail in the subsequent sections.

\subsubsection{Lightweight Synthetic Domain Augmentation}

To simulate domain variability without relying on expensive episodic training (as in DG$^2$CD-Net~\cite{dg2net}), we generate only one or two synthetic domains per experiment using a diffusion model guided by GPT-4o~\cite{openai2024chatgpt}-curated prompts (e.g., \texttt{underwater}, \texttt{night-time} variants of \emph{class} instances) (see \textbf{Sup. Mat.} for qualitative visualizations). These synthetic domains serve as proxy distributions that expose the model to varied visual shifts and support generalization to unseen domains.

Unlike DG$^2$CD-Net, which depends on numerous episodic tasks and synthetic domain permutations, our strategy is lightweight and avoids both computational overhead and error propagation across episodes. Crucially, in hyperbolic space, even a small number of diverse augmentations can induce expansive representational changes due to the geometry’s exponential capacity---effectively stretching the semantic space and encouraging separation between seen and unseen regions.

First, we introduce a \textit{domain-diversity score} that ranks a given synthetic domain $\mathcal{D}_{\text{syn}}^{(s)}$ with respect to other synthetic domains $\{\mathcal{D}_{\text{syn}}^{(l)}\}_{l=1}^{\mathcal{M}}$ and the source domain $\mathcal{D}_S$ based on the notion of mutual diversity calculated using the Fréchet Inception Distance (FID)~\cite{eiter1994computing} for $\mathcal{M}$ synthesized domains:


\begin{equation}
\text{Score}(\mathcal{D}_{syn}^{(s)}) = \frac{1}{\mathcal{M}-1} \sum_{\substack{l=1, \\ l \neq s}}^{\mathcal{M}} \bigg[
\text{FID}(\mathcal{D}_S, \mathcal{D}_{\text{syn}}^{(s)}) + \text{FID}(\mathcal{D}_{\text{syn}}^{(s)}, \mathcal{D}_{\text{syn}}^{(l)})
\bigg].
\end{equation}



This scoring promotes both source-domain divergence and intra-pair complementarity. We select the top-scoring $1-2$ domains to augment $\mathcal{D}_S$ to obtain $\mathcal{D}_{\text{train}}$.

As shown in Fig.~\ref{fig:ablation_dual}, excessive augmentation leads to overfitting on seen classes and degrades novel class discovery. \textit{Notably, training solely on \( \mathcal{D}_S \) yields competitive results, highlighting the inherent domain robustness of hyperbolic space, and using these augmentation provides marginal boosts.} Effects of redundant augmentations are mentioned in \textbf{Sup. Mat.}

\subsubsection{Mapping Visual Features into Curvature-Aware Hyperbolic Geometry}
With the augmented training set, we learn representations using a frozen DINO~\cite{dino}-pretrained ViT~\cite{vit} followed by a 3-layer MLP. The resulting Euclidean feature \( \mathbf{z}^{\mathbb{E}} \in \mathbb{R}^d \) is projected into the Poincar\'e ball \( \mathbb{D}_c^d \) via the exponential map:
\begin{equation}
\mathbf{z}_i^s = \exp_0^c(\mathbf{z}^{\mathbb{E}}) = \tanh(\sqrt{c} \|\mathbf{z}^{\mathbb{E}}\|) \cdot \frac{\mathbf{z}^{\mathbb{E}}}{\sqrt{c} \|\mathbf{z}^{\mathbb{E}}\|},
\end{equation}
where $c$ is a learnable curvature parameter in our case to approximate the data complexity more effectively. Let $z_i \coloneqq \mathbf{z}_i^s$ denote the hyperbolic feature. This projection facilitates the encoding of hierarchical semantics and ensures geometric consistency in the downstream tasks.

\subsubsection{Tangent CutMix and Adaptive Outlier Loss}
\label{sec:tangent_cutmix}

To hallucinate novel-category samples and regularize the open space in hyperbolic geometry, we introduce \textbf{Tangent CutMix}~\cite{yun2019cutmix}—a curvature-aware variant of CutMix tailored for the Poincaré ball. Traditional CutMix interpolates feature representations in Euclidean space to synthesize outliers, which can violate the geometric constraints of hyperbolic space. In contrast, Tangent CutMix performs mixing in the tangent space at the origin, ensuring consistency with the underlying manifold structure.

Given two embeddings \( z_i, z_j \in \mathbb{D}_c^d \) with different class labels, we:

\begin{enumerate}[leftmargin=1.5em, label=(\arabic*)]
  \item \textbf{Project to tangent space:} \( v_i = \log_0^c(z_i),\; v_j = \log_0^c(z_j) \)
    \item \textbf{Linear mix:} Compute \( v_{\text{mix}}^{i,j} = \lambda v_i + (1 - \lambda) v_j \), where \( \lambda \sim \mathrm{Beta}(1,1)= \operatorname{Uniform}(0,1) \)
  \item \textbf{Map back:} \( z_{\text{mix}}^{i,j} = \exp_0^c(v_{\text{mix}}) \)
\end{enumerate}

The resulting embedding \( z_{\text{mix}} \) represents a curvature-preserving interpolation of features with incompatible semantics, mimicking out-of-distribution behavior while remaining valid in the hyperbolic space.
Furthermore, to prevent these synthetic features from collapsing into known class regions, we apply an adaptive outlier loss:
\begin{equation}
\mathcal{L}_{\text{out}} = \underset{(x,y) \sim \mathcal{P}( \mathcal{D}_{\text{train}})}{\mathbb{E}} \sum_{i,j, y_i \neq y_j} \max(0, \gamma - \min_{k \in \mathcal{Y}_s} \mathbb{D}_{\mathbb{H}}(z_{\text{mix}}^{i,j}, \mathbf{p}_k)),
\end{equation}
where \( \mathbb{D}_{\mathbb{H}} \) is the hyperbolic distance, and \( \gamma \) is a quantile-based adaptive margin over the distances from all class prototypes in \( \{\mathbf{p}_k\}_{k=1}^{|\mathcal{Y}_s|} \). This encourages pseudo-novel embeddings to remain outside the regions occupied by seen classes, effectively reserving space for novel category discovery. For further analysis regarding adaptive margin and the generated CutMix samples, see \textbf{Sup. Mat.}


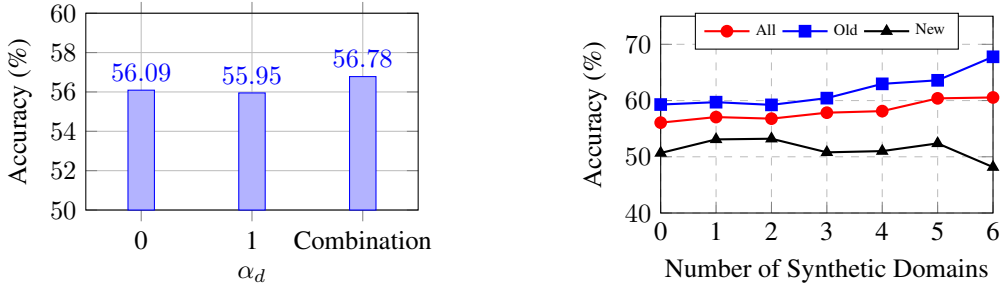
\begin{figure}[ht]
\centering
\begin{minipage}[t]{0.45\textwidth}
\centering
\begin{tikzpicture}
\begin{axis}[
    ybar,
    symbolic x coords={0, 1, Combination},
    xtick=data,
    ylabel={Accuracy (\%)},
    xlabel={$\alpha_d$},
    ymin=50, ymax=60,
    bar width=10pt,
    nodes near coords,
    enlarge x limits=0.25,
    height=4.2cm,
    width=6cm,
    grid=major,
]
\addplot coordinates {(0, 56.09) (1, 55.95) (Combination, 56.78)};
\end{axis}
\end{tikzpicture}
\end{minipage}%
\hfill
\begin{minipage}[t]{0.45\textwidth}
\centering
\begin{tikzpicture}
\begin{axis}[
    xlabel=Number of Synthetic Domains,
    ylabel=Accuracy (\%),
    xmin=0, xmax=6,
    ymin=40, ymax=75,
    xtick={0,1,2,3,4,5,6},
    xticklabels={0,1,2,3,4,5,6},
    grid=both,
    grid style=dashed,
    height=4.2cm,
    width=6cm,
    legend style={
        draw=black,
        fill=white,
        at={(0.5,0.83)},
        anchor=south,
        font=\tiny,
        legend columns=3,
    },
    cycle list name=color list,
]

\addplot+[mark=*, thick] coordinates {
    (0,56.07) (1,57.06) (2,56.78) (3,57.84) (4,58.12) (5,60.39) (6,60.55)
};
\addplot+[mark=square*, thick] coordinates {
    (0,59.29) (1,59.71) (2,59.23) (3,60.43) (4,62.97) (5,63.6) (6,67.78)
};
\addplot+[mark=triangle*, thick] coordinates {
    (0,50.67) (1,53.09) (2,53.21) (3,50.79) (4,51.01) (5,52.38) (6,48.16)
};

\legend{All, Old, New}

\end{axis}
\end{tikzpicture}
\end{minipage}
\vspace{-3pt}
\caption{
\textbf{(Left)} Effect of $\alpha_d$ in hybrid contrastive loss. A balanced combination of angular and geodesic components achieves the highest accuracy. 
\textbf{(Right)} Impact of synthetic domains on old and new category performance. While old-class accuracy increases due to augmented seen data, new-class performance slowly degrades with more synthetic domains, as they cause seen-class bias.
}
\label{fig:ablation_dual}
\vspace{-12px}
\end{figure}

\subsubsection{Prototype Anchoring with Penalized Busemann Loss}

To enforce semantic structure among the known categories, we associate each class \( k \in \mathcal{Y}_s \) with a fixed \textit{ideal prototype} \( \mathbf{p}_k \in \partial \mathbb{D}_c^d \)~\cite{ghadimi2021hyperbolic}, placed uniformly on the boundary of the Poincaré ball. These prototypes serve as directional anchors and remain fixed throughout training, enabling compact clustering of seen-class features while leaving the interior volume of the ball available for unknown category discovery.

To align the features \( z_i \) with their respective class prototypes, we adopt a \textbf{penalized Busemann loss}:
\begin{equation}
\mathcal{L}_{\text{Buse}} = \log\left( \frac{\|z_i - \mathbf{p}_{y_i}\|^2}{1 - c \|z_i\|^2} \right) + \phi \log(1 - \|z_i\|^2),
\end{equation}
where \( \mathbf{p}_{y_i} \) is the prototype corresponding to the class label of $z_i$, and \( \phi \) is a regularization coefficient.
The first term guides directional alignment between features and their prototypes, preserving semantic proximity in the hyperbolic geometry. The second term penalizes embeddings that approach the boundary too aggressively, thereby maintaining stability during optimization and avoiding overconfidence.

\subsubsection{Hybrid Hyperbolic Contrastive Loss}

While the Busemann loss anchors known classes via directional alignment, it does not explicitly enforce local structure among unlabeled or novel samples. To address this, we incorporate a \textbf{hybrid hyperbolic contrastive loss}~\cite{Liu2025HypCD}, designed to refine the latent space by encouraging consistency between augmented views and separating unrelated instances—even in the absence of explicit labels.

For each positive pair of embeddings \(  z_i',z_i^{''} \), corresponding to different augmentations of the same input, we define the contrastive objective as:
\begin{equation}
\mathcal{L}_u = \frac{1}{|B|} \sum_{i \in B} -\log \frac{\exp(\delta(z_i^{''}, z_i') / \tau)}{\sum_{j \neq i} \exp(\delta(z_i', z_j) / \tau)},
\end{equation}
where \( \tau \) is a temperature hyperparameter and \( B \) is the batch of samples.
We use a \textbf{hybrid similarity function} \( \delta(., .) \), which linearly combines distance-based and angle-based measures:
\begin{equation}
\delta(., .) = \alpha_d \cdot \underbrace{[-\mathbb{D}_{\mathbb{H}}(., .)]}_{L_{u}^{dis}} + (1 - \alpha_d) \cdot \underbrace{\cos(., .)}_{L_{u}^{ang}},
\end{equation}
\( \cos(\cdot, \cdot) \) computes cosine similarity in the tangent space, thanks to the co-conformality of the Euclidean and Hyperbolic spaces. $\alpha_d$ is the balancing factor (for more details see \textbf{Sup. Mat.}).

This hybrid formulation leverages the metric structure of hyperbolic space to promote global semantic separation via geodesic distances, while retaining angular consistency within local neighborhoods. Fig. \ref{fig:ablation_dual} shows the importance of the full $\delta$ over the individual distance metrics.

\subsubsection{Training Objective}
Our final \textit{minimization} objective integrates all components:
\begin{equation}
\mathcal{L}_{\text{total}} = 
\lambda_1 \underbrace{\mathcal{L}_{\text{Buse}}}_{\text{Semantic alignment}} + 
\lambda_2 \underbrace{\mathcal{L}_{u}}_{\text{Contrastive regularization}} + 
\lambda_3 \underbrace{\mathcal{L}_{\text{out}}}_{\text{Outlier repulsion}}, \quad \text{where } \lambda_1 + \lambda_2 + \lambda_3 = 1.
\end{equation}

\paragraph{Test-time Protocol.} After training, we extract hyperbolic features from target-domain samples and perform clustering using K-Means as in ~\cite{dg2net, gcd}. {See \textbf{Sup. Mat.} for detailed algorithm.}

\subsection{Theoretical Justification: Generalization in Euclidean vs. Hyperbolic Spaces}

\label{sec:theoretical_proof}

We analyze \textsc{HiDISC} under the lens of generalization theory in hyperbolic space. Given \( \mathcal{D}_S \), \( \mathcal{D}_T \), and synthetic augmentations \( \{\mathcal{D}_{\text{syn}}^{(l)}\}_{l=1}^{\mathcal{M}} \), the goal is to minimize the expected target risk:
\begin{equation}
\mathcal{L}_{T}(f) = \mathbb{E}_{x \sim T}[\ell(f(x))],
\end{equation}
Extending Rademacher-based analysis to the Poincaré ball \( \mathbb{D}_c^d \)~\cite{chami2019hyperbolic}, we obtain:
\begin{equation}
\mathcal{L}_{T}(f) \leq \mathcal{L}_{S'}(f) + \Delta_{\mathbb{H}}(S', T) + \mathcal{R}_{\mathbb{H}}(\mathcal{H}) + \epsilon,
\end{equation}
where:
\( \mathcal{L}_{S'}(f) \): empirical loss over the augmented training set \( S' \);
\( \Delta_{\mathbb{H}}(S', T) \): hyperbolic discrepancy between augmented source and target;
\( \mathcal{R}_{\mathbb{H}}(\mathcal{H}) \): Rademacher complexity of the hypothesis class \( \mathcal{H} \);
\( \epsilon \): a residual optimization error.

Compared to the Euclidean bound \( \mathcal{L}_{T}^{\mathbb{E}}(f) \leq \mathcal{L}_{S'}(f) + \Delta_{\mathbb{E}}(S', T) + \mathcal{R}_{\mathbb{E}}(\mathcal{H}) + \epsilon \), the hyperbolic version benefits from the exponential volume and hierarchical structure of \( \mathbb{D}_c^d \). This allows semantically distant concepts to be placed further apart with less distortion and curvature-driven compression around known classes—thereby making fewer, well-chosen augmentations sufficient to span the generalization space. As such, \( \Delta_{\mathbb{H}}(S', T) < \Delta_{\mathbb{E}}(S', T) \) holds under the same augmentation budget, yielding a tighter bound (see \textbf{Sup. Mat.} for a formal proof).

Each loss in \textsc{HiDISC} contributes to improving specific terms:
\textbf{(i)} The Busemann loss \( \mathcal{L}_{\text{Buse}} \) aligns seen-class features to ideal prototypes at the boundary, stabilizing \( \mathcal{L}_{S'}(f) \) via directional compactness;
\textbf{(ii)} The hybrid contrastive loss \( \mathcal{L}_u \) integrates angular and geodesic similarity to encourage semantically meaningful clusters and reduce model complexity \( \mathcal{R}_{\mathbb{H}}(\mathcal{H}) \);
\textbf{(iii)} The outlier loss \( \mathcal{L}_{\text{out}} \), applied on Tangent CutMix samples, helps partition the open space, reducing false positives on novel categories without explicit domain alignment;
\textbf{(iv)} The curated synthetic domains \( \{\mathcal{D}_{\text{syn}}^{(l)}\} \) enrich \( S' \), approximating \( T \)'s support and reducing \( \Delta_{\mathbb{H}}(S', T) \) in a geometry-consistent manner.

In \textbf{Sup. Mat.}, we show that FID-based estimates of \( \Delta_{\mathbb{H}} \) yield minimal improvement over the inherent domain-independence of hyperbolic geometry. We also compare our loss terms in Euclidean and hyperbolic spaces, demonstrating that hyperbolic geometry better reduces the generalization gap.




\begin{figure*}[t]
    \centering
    \includegraphics[width=0.95\linewidth]{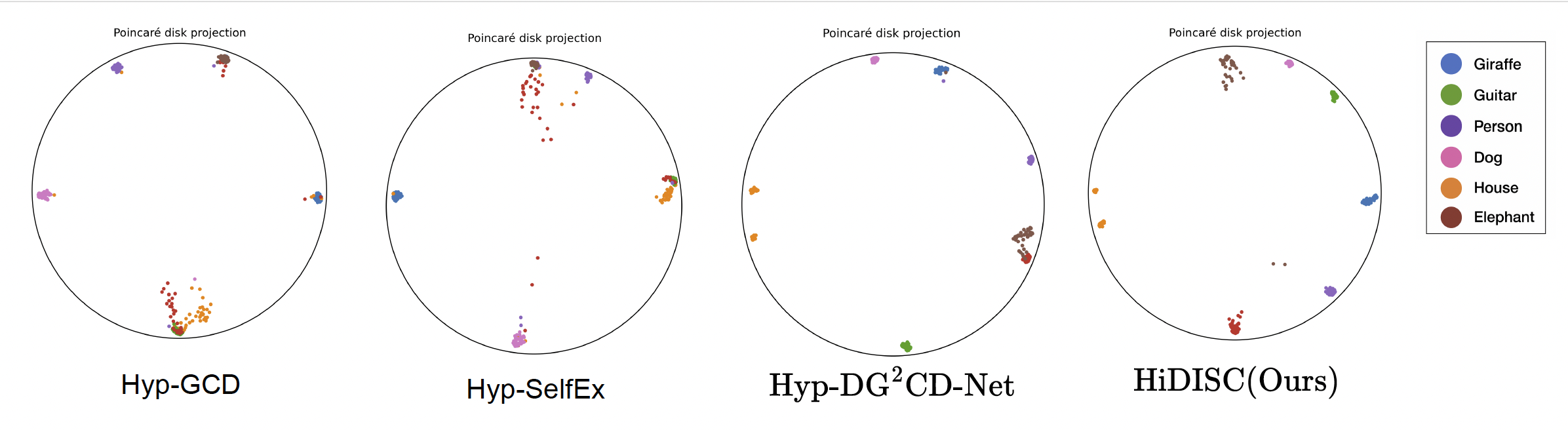}
    \vspace{-2px}
    \caption{\textbf{Poincaré‐disk UMAP \cite{mcinnes2018umap} embeddings} of the target domain (``Photo") clusters, as produced by Hyp-GCD \cite{Liu2025HypCD}, Hyp-SelfEx \cite{Liu2025HypCD}, Hyp-\prevmodel, and \ourmodel  (Ours) for the PACS dataset, with ``Sketch" as the source. \ourmodel\ produces a visually clean and compact embedding space, supported by \textbf{silhouette scores}~\cite{rousseeuw1987silhouettes} ($\in [-1.1]$, $\uparrow$), indicating improved cluster compactness and separation: \hl{(Hyp-GCD: \textbf{-0.52}, Hyp-SelfEx: \textbf{-0.42}, Hyp-DG$^2$CD-Net: \textbf{-0.29},  \textsc{HiDisc}: \textbf{-0.14})}}
    \label{fig:umap}
    \vspace{-15px}
\end{figure*}

\vspace{-2mm}
\section{Experimental Evaluations}
\label{sec:experimental_settings}
\noindent \textbf{Dataset Details.}
We evaluate our method on three standard DG-GCD benchmarks: PACS~\cite{pacs}, Office-Home\cite{officehome}, and Domain-Net\cite{domainnet}. We follow the protocol of \cite{dg2net} for constructing known/novel class-splits and source-target domain pairs. The dataset details are provided in the \textbf{Sup. Mat}.

\noindent \textbf{Evaluation Metrics.}
Following~\cite{gcd,rongali2024cdadnetbridgingdomaingaps,dg2net}, we evaluate clustering using three metrics: \textbf{Old} (accuracy on known classes \( \mathcal{Y}_t^{\text{old}} \)), \textbf{New} (accuracy on novel classes \( \mathcal{Y}_t^{\text{new}} \)), and \textbf{All} (overall accuracy on \( \mathcal{D}_T \)). Hungarian matching is used to align predicted clusters with ground-truth labels. Scores are averaged over three runs and all source-target combinations.
Further experimental details and hyper-parameter choices are mentioned in \textbf{Sup. Mat.}



\begin{table*}[t]
\centering
\vspace{-5px}
\caption{
\textbf{Comparison of clustering accuracy} (\%) for known (Old), novel (New), and overall (All) categories across PACS, Office-Home, and DomainNet. It can be seen that \textsc{HiDisc} beats other synthetic domain augmentation based baselines using significantly less number of synthetic domains (from 6/9 to 2). (\textbf{Bold} : best , \underline{underline} : second best).
}
\label{tab:dg-gcd-main}
\scriptsize
\setlength{\tabcolsep}{3pt}
\renewcommand{\arraystretch}{1.05}
\rowcolors{3}{gray!5}{white}
\begin{tabular}{lccccccccccccc}
\toprule
\textbf{Method} & \textbf{Venue} & \multicolumn{3}{c}{\textbf{PACS}} & \multicolumn{3}{c}{\textbf{Office-Home}} & \multicolumn{3}{c}{\textbf{DomainNet}} & \multicolumn{3}{c}{\textbf{Avg.}} \\
\cmidrule(lr){3-5} \cmidrule(lr){6-8} \cmidrule(lr){9-11} \cmidrule(lr){12-14}
 & & All & Old & New & All & Old & New & All & Old & New & All & Old & New \\
\midrule
\textcolor{black}{ViT} \cite{vit} & ICLR'21 & 41.98 & 50.91 & 33.16 & 26.17 & 29.13 & 21.62 & 25.35 & 26.48 & 22.41 & 31.17 & 35.51 & 25.73 \\
\textcolor{black}{GCD} \cite{gcd} & CVPR'22 & 52.28 & 62.20 & 38.39 & 52.71 & 54.19 & 50.29 & 27.41 & 27.88 & 26.13 & 44.13 & 48.09 & 38.27 \\
\textcolor{black}{SimGCD} \cite{gcd3} & ICCV'23 & 34.55 & 38.64 & 30.51 & 36.32 & 49.48 & 13.55 & 2.84  & 2.16  & 3.75  & 24.57 & 30.09 & 15.94 \\
\textcolor{black}{CMS} \cite{gcd6} & CVPR'24 & 28.95 & 28.13 & 36.80 & 10.02 & 9.66  & 10.53 & 2.33  & 2.40  & 2.17  & 13.77 & 13.40 & 16.50 \\
\textcolor{black}{SelfEx} \cite{gcd8} & ECCV'24 & 71.82 & 73.37 & 71.55 & 50.18 & 48.59 & 52.16 & 24.78 & 24.99 & 24.21 & 48.93 & 48.98 & 49.31 \\
\textcolor{black}{CDAD-Net} \cite{rongali2024cdadnetbridgingdomaingaps} & CVPR-W'24 & 69.15 & 69.40 & 68.83 & 53.69 & \underline{57.07} & 47.32 & 24.12 & 23.99 & 24.35 & 48.99 & 50.15 & 46.83 \\
\midrule
\textcolor{blue!60!black}{GCD+ 6 Synth} & CVPR'22 & 65.33 & 67.10 & 64.42 & 50.50 & 51.48 & 48.96 & 24.71 & 24.80 & 21.94 & 46.85 & 47.78 & 45.11 \\
\textcolor{blue!60!black}{SimGCD+ 6 Synth} & ICCV'23 & 39.76 & 43.76 & 35.97 & 35.57 & 48.58 & 12.89 & 2.71  & 1.99  & 4.14  & 26.01 & 31.44 & 17.67 \\
\textcolor{blue!60!black}{CMS+ 6 Synth} & CVPR'24 & 28.01 & 26.71 & 29.04 & 12.09 & 12.66 & 11.13 & 3.22  & 3.28  & 3.03  & 14.44 & 14.22 & 14.40 \\
\textcolor{blue!60!black}{CDAD+ 6 Synth} & CVPR-W'24 & 60.76 & 61.67 & 59.49 & 53.49 & 56.90 & 47.76 & 23.85 & 23.88 & 24.26 & 46.03 & 47.47 & 43.84 \\
\midrule
\textcolor{purple!60!black}{Hyp-GCD} \cite{Liu2025HypCD} & CVPR'25 & 65.33 & 67.11 & 64.42 & 50.13 & 49.36 & 48.08 & 22.88  & 23.74  & 25.89  & 46.12 & 46.74 & 46.13 \\
\textcolor{purple!60!black}{Hyp-SelfEx} \cite{gcd8} & ECCV'24 & 72.44 & 74.70 & 71.20 & 52.91 & 52.65 & 52.96 & \underline{29.30} & \underline{30.45} & \underline{26.37} & 51.55 & 52.60 & 50.18 \\
\midrule
\textcolor{red!70!black}{\textbf{DG$^2$CD-Net}} \cite{dg2net} (9 Synth) & CVPR'25 & 73.30 & \underline{75.28} & 72.56 & \underline{53.86} & 53.37 & \textbf{54.33} & 29.01 & 30.38 & 25.46 & \underline{52.06} & \underline{53.01} & \underline{50.78} \\
\textcolor{purple!60!black}{Hyp-DG$^2$CD-Net \textsuperscript{\textdagger}(9 Synth)} & CVPR'25 & \underline{74.07} & 74.40 & \underline{73.95} & 49.40 & 50.29 & 48.03 & 22.31 & 21.52 & 24.29 & 48.59 & 48.74 & 48.76 \\
\midrule
\textcolor{red!80!black}{\textbf{\textsc{HiDISC} (Ours)}} (2 Synth) & -- & \textbf{75.07} & \textbf{75.54} & \textbf{74.52} & \textbf{56.78} & \textbf{59.23} & \underline{53.21} & \textbf{30.51} & \textbf{31.40} & \textbf{28.41} & \textbf{54.12} & \textbf{55.39} & \textbf{52.05} \\
\textbf{$\Delta$} & -- 
& \textcolor{green!60!black}{+1.00} & \textcolor{green!60!black}{+0.26} & \textcolor{green!60!black}{+0.57}
& \textcolor{green!60!black}{+2.92} & \textcolor{green!60!black}{+2.16} & \textcolor{red!70!black}{–1.12}
& \textcolor{green!60!black}{+1.21} & \textcolor{green!60!black}{+0.95} & \textcolor{green!60!black}{+2.04}
& \textcolor{green!60!black}{+2.06} & \textcolor{green!60!black}{+2.38} & \textcolor{green!60!black}{+1.27} \\
\midrule
\textcolor{gray!70!black}{CDAD-Net (DA) [UB]} & CVPR-W'24 & 83.25 & 87.58 & 77.35 & 67.55 & 72.42 & 63.44 & 70.28 & 76.46 & 65.19 & 73.69 & 78.82 & 68.66 \\
\bottomrule
\end{tabular}
\vspace{-20px}
\end{table*}

\vspace{-5px}
\subsection{Comparisons to the Literature}
\vspace{-5px}
Table~\ref{tab:dg-gcd-main} compares our proposed \textsc{HiDISC} against state-of-the-art methods on the said datasets. Baselines are categorized into four groups: (i) \textbf{Euclidean source-only GCD methods}, including GCD~\cite{gcd}, SimGCD~\cite{gcd3}, CMS~\cite{gcd6}, and SelfEx~\cite{gcd8}; (ii) \textbf{Synthetic augmentation-based GCD methods}, such as SimGCD+Synthetic, CMS+Synthetic, and CDAD-Net+Synthetic~\cite{rongali2024cdadnetbridgingdomaingaps}, which incorporate domain-shifted images via diffusion-based generation; (iii) \textbf{Hyperbolic GCD methods}, including Hyp-GCD~\cite{Liu2025HypCD} and Hyp-SelfEx, which project features into hyperbolic space to improve clustering but do not generalize across domains. To ensure consistency with the DG-GCD setting, we retain only the components of these methods that rely on labeled data during training and omit terms involving unlabeled samples in all the above baselines, as recommended in~\cite{dg2net}; 
and (iv) the \textbf{DG-GCD baseline} DG$^2$CD-Net~\cite{dg2net}, which simulates multiple domains using diffusion models and aggregates task-level knowledge via episodic training and task vectors. For a fairer comparison, we also implement a hyperbolic variant, Hyp-DG$^2$CD-Net, by replacing its embedding space with a Poincaré ball. As in~\cite{dg2net}, we report results for CDAD-Net~\cite{rongali2024cdadnetbridgingdomaingaps} under joint access to source and target domains as an upper bound of our results.


\begin{wrapfigure}{r}{0.48\textwidth} 
\vspace{-20pt} 
\begin{minipage}{0.48\textwidth}
\centering
\captionof{table}{\textbf{Estimated number of clusters}. Correct estimates are in \textcolor{green!60!black}{green}, small errors in \textcolor{orange!80!black}{orange}, and large deviations in \textcolor{red!80!black}{red}.}
\label{tab:cluster_estimation}
\renewcommand{\arraystretch}{1.2}
\rowcolors{2}{gray!5}{white}
\resizebox{\textwidth}{!}{%
\begin{tabular}{lccc}
\toprule
\textbf{Method}               & \textbf{PACS} & \textbf{Office-Home} & \textbf{DomainNet} \\
\midrule
Ground Truth                  & 7             & 65                   & 345                \\
\textbf{DG$^2$CD-Net}         & \textcolor{green!60!black}{7} & \textcolor{orange!80!black}{67} & \textcolor{orange!80!black}{355} \\
CDAD-Net (DG)                & \textcolor{red!80!black}{12}  & \textcolor{orange!80!black}{60} & \textcolor{red!80!black}{362}  \\
CDAD-Net (DA)                & \textcolor{green!60!black}{7} & \textcolor{green!60!black}{66}  & \textcolor{orange!80!black}{349} \\
\textbf{\ourmodel~(Ours)}     & \textcolor{green!60!black}{7} & \textcolor{green!60!black}{66}  & \textcolor{orange!80!black}{351} \\
\bottomrule
\end{tabular}
}
\end{minipage}
\vspace{-10pt}
\end{wrapfigure}
\vspace{-5pt}
Quantitatively, \textsc{HiDISC} achieves state-of-the-art performance across all metrics and datasets. It improves upon DG$^2$CD-Net by +2.06\% in average overall clustering accuracy and by +1.27\% on novel class discovery. On DomainNet—the most diverse and challenging benchmark—\textsc{HiDISC} outperforms the best previous method by +1.21\%. UMAP visualizations (Fig. \ref{fig:umap}) show {\ourmodel} forms a compact embedding space. These gains are achieved without target access and with over \textbf{96$\times$ lower training FLOPs} than \cite{dg2net} while using the same number of synthetic domains (see \textbf{Sup. Mat}). On the other hand, the performance of DG$^2$CD-Net degrades drastically as the number of synthetic domains is reduced (see \textbf{Sup. Mat.})        

Furthermore, \label{sec:cluster_estimation}
Table~\ref{tab:cluster_estimation} compares the \textbf{estimated number of clusters} inferred by each method against the ground truth on PACS, Office-Home, and DomainNet. \textsc{HiDisc} is found to approximate the cluster counts more precisely than the counterparts.

The \textbf{learnable curvature} converges to dataset-specific values: 0.041 for Office-Home, 0.059 for PACS, and 0.38 for DomainNet. The curvature evolution plots are mentioned in \textbf{Sup. Mat.}

\vspace{-5px}
\subsection{Ablation Analysis}

\noindent \textbf{Impact of Loss Components and Key Model Components.}
Table~\ref{tab:loss_components} evaluates the contribution of each loss component in {\ourmodel} on Office-Home. The vanilla model, trained without any loss terms, yields only 26.17\% overall accuracy. Introducing the Busemann loss alone improves performance substantially to 56.32\%, while the hybrid hyperbolic contrastive loss independently achieves 50.95\%. Combining both leads to further gains, particularly in old-class accuracy. Incorporating the outlier repulsion term yields the best overall result, with 56.78\% total accuracy, 59.23\% on known classes, and 53.21\% on novel classes.

Table~\ref{tab:ablation} presents ablations on key architectural components. Manual augmentations achieve 50.80\% accuracy. Replacing the learnable curvature with static values ($c=0.01$ and $0.03$) reduces accuracy considerably, as it fails to manage the data manifold effectively. Substituting Tangent CutMix with Euclidean CutMix lowers performance by 3.96\%, confirming the benefits of curvature-consistent mixing. The complete {\ourmodel} configuration consistently outperforms all variants, confirming the complementary benefits of its geometric and loss-driven design.

\begin{table}[t]
    \centering
    \begin{minipage}[t]{0.48\textwidth}
        \centering
        \caption{Impact of \textbf{loss components} of {\ourmodel} on Office-Home}
        \resizebox{\linewidth}{!}{%
        \begin{tabular}{ccccccc}
            \toprule
            \multirow{2}{*}{\textbf{Config.}} & \multirow{2}{*}{$\mathcal{L}_{\text{Buse}}$} & \multirow{2}{*}{$\mathcal{L}_{\text{hrep}}^{u}$} & \multirow{2}{*}{$\mathcal{L}_{\text{out}}$} & \multicolumn{3}{c}{\textbf{Office-Home}} \\
            \cline{5-7}
             &  &  &  & \textbf{All} & \textbf{Old} & \textbf{New} \\
            \midrule
            \cellcolor{pink!10}\textbf{Vanilla} & \ding{55} & \ding{55} & \ding{55} & 26.17 & 29.13 & 21.62 \\
            \cellcolor{red!10}\textbf{$\mathcal{L}_{\text{Buse}}$} & \ding{51} & \ding{55} & \ding{55} & 56.32 & 59.74 & 50.32 \\
            \cellcolor{red!10}\textbf{$\mathcal{L}_{\text{hrep}}^{u}$} & \ding{55} & \ding{51} & \ding{55} & 50.95 & 49.33 & 53.06 \\
            \cellcolor{gray!10}\textbf{$\mathcal{L}_{\text{Buse}}$+$\mathcal{L}_{\text{hrep}}^{u}$} & \ding{51} & \ding{51} & \ding{55} & 56.29 & 60.36 & 50.41 \\
            \cellcolor{gray!10}\textbf{$\mathcal{L}_{\text{Buse}}$+$\mathcal{L}_{\text{out}}$} & \ding{51} & \ding{55} & \ding{51} & 51.04 & 51.51 & 50.29 \\
            \rowcolor{blue!10}
            \textbf{Full \ourmodel} & \ding{51} & \ding{51} & \ding{51} & \textbf{56.78} & \textbf{59.23} & \textbf{53.21} \\
            \bottomrule
        \end{tabular}%
        }
        \label{tab:loss_components}
        \vspace{-10px}
    \end{minipage}\hfill
    \begin{minipage}[t]{0.48\textwidth}
        \centering
        \caption{Performance metrics demonstrating the \textbf{influence of key model components} of {\ourmodel} for Office-Home.}
        \resizebox{\linewidth}{!}{%
        \begin{tabular}{lccc}
            \toprule
            \multirow{2}{*}{\textbf{Model Variant}} & \multicolumn{3}{c}{\textbf{Office-Home}} \\
            \cline{2-4}
            & \textbf{All} & \textbf{Old} & \textbf{New} \\
            \midrule
             \textbf{- With manual augmentations based $\mathcal{D}_{\text{syn}}$} & 50.80 & 51.75 & 49.15 \\
             \textbf{- Without synthetic domain} & 56.07 & 59.29 & 50.67 \\
            \textbf{- Fixed curvature (c=0.01, close to Euclidean)} & 56.23 & 58.65 & 52.68 \\
             \textbf{- Fixed curvature (c=0.03)} & 55.67 & 57.39 & 52.69 \\
             \textbf{- Cut-Mix (In Euclidean Space)} & 53.46 & 54.86 & 51.06 \\
            \hline
            \rowcolor{blue!10}
            \textbf{- Full {\ourmodel}} & \textbf{56.78} & \textbf{59.23} & \textbf{53.21} \\
            \bottomrule
        \end{tabular}}
        \label{tab:ablation}
    \vspace{-15px}
    \end{minipage}
\end{table}

\noindent \textbf{Ablation of Norm Radius and Slope in Hyperbolic Embedding.} 
\label{sec:ablation_radius_phi} We study two key hyperparameters in our hyperbolic embedding setup: the $\ell_2$ norm radius before exponential mapping and the slope $\phi$ in the penalized Busemann loss, both controlling embedding compactness and placement. As per Table~\ref{tab:slope_ablation}, lower slopes (e.g., $\phi = 0.10$) favor seen-class accuracy but hurt generalization, while higher slopes (e.g., $\phi = 0.90$) improve novel-class performance by restricting dispersion. We choose $\phi = 0.75$ for balance. For the norm radius, Table~\ref{tab:radius_ablation} shows that $1.5$ best balances alignment to boundary-anchored prototypes and generalization, whereas smaller values (e.g., $1.0$) overfit $\mathcal{Y}_S$.

\begin{table}[!htbp]
    \centering
    \vspace{-10pt}
    \caption{
        \textbf{Ablation on hyperbolic embedding parameters on Office-Home.} (\textbf{Left}) Effect of slope coefficient $\phi$ in the penalized Busemann loss. Lower $\phi$ concentrates embeddings near the boundary, improving seen-class accuracy but reducing generalization. (\textbf{Right}) Effect of $\ell_2$ radius constraint before exponential mapping. Radius = 1.5 yields the best trade-off between known and novel categories.
    }
    \vspace{0.5em}
    \begin{minipage}[t]{0.45\textwidth}
        \centering
        \renewcommand{\arraystretch}{1.1}
        \rowcolors{2}{gray!5}{white}
        \begin{tabular}{c|c|c|c}
            \toprule
            \textbf{Slope $\boldsymbol{\phi}$} & \textbf{All} & \textbf{Old} & \textbf{New} \\
            \midrule
            0.10 & 58.84 & 65.77 & 47.07 \\
            0.75 & \textbf{56.78} & \textbf{59.23} & \textbf{53.21} \\
            0.90 & 57.76 & 62.82 & 49.18 \\
            \bottomrule
        \end{tabular}
        
        \label{tab:slope_ablation}
    \end{minipage}
    \hfill
    \begin{minipage}[t]{0.45\textwidth}
        \centering
        \renewcommand{\arraystretch}{1.1}
        \rowcolors{2}{gray!5}{white}
        \begin{tabular}{c|c|c|c}
            \toprule
            \textbf{Radius} & \textbf{All} & \textbf{Old} & \textbf{New} \\
            \midrule
            1.5 & \textbf{56.78} & \textbf{59.23} & \textbf{53.21} \\
            1.0 & 57.33 & 61.14 & 51.76 \\
            2.3 & 57.31 & 60.96 & 52.04 \\
            \bottomrule
        \end{tabular}
        
        \label{tab:radius_ablation}
    \end{minipage}
    \vspace{-6pt}
\end{table}

\noindent \textbf{Choice of Hyperbolic Model: Poincaré vs. Lorentz Model}

We adopt the Poincaré ball model, which we find performs more favorably than the Lorentz model~\cite{nickel2018learning} for DG-GCD on Office-Home in Table \ref{tab:poincare}. Full theoretical details are in the \textbf{Sup. Mat.} The empirical comparison is below:

\noindent\textbf{Hyperparameter Sensitivity for Loss Weights.}
We conducted an ablation study on the loss weights ($\lambda_1, \lambda_2, \lambda_3$) and found our chosen configuration achieves near-optimal performance, demonstrating robustness Table \ref{tab:ablation_loss}. More details are in the \textbf{Sup. Mat.}

\begin{table*}[!ht]
\centering 

\begin{minipage}{0.48\textwidth}
    \centering
    \captionof{table}{Comparison of Poincaré ball and Lorentz models on Office-Home.}
    \label{tab:poincare}
    \resizebox{\linewidth}{!}{%
        \begin{tabular}{lccc}
            \toprule
            \textbf{Model} & \textbf{All} & \textbf{Old} & \textbf{New} \\
            \midrule
            Poincaré Ball~\cite{nickel2017poincare} & \textbf{56.78} & \textbf{59.23} & \textbf{53.21} \\
            Lorentz Model~\cite{nickel2018learning} & 54.28 & 56.01 & 51.41 \\
            \bottomrule
        \end{tabular}%
    }
\end{minipage}
\hfill 
\begin{minipage}{0.48\textwidth}
    \centering
    \captionof{table}{Ablation study on loss term weights ($\lambda_1, \lambda_2, \lambda_3$) on OfficeHome.}
    \label{tab:ablation_loss}
    \resizebox{\linewidth}{!}{%
        \begin{tabular}{l ccc ccc}
            \toprule
            & \multicolumn{3}{c}{\textbf{Loss Weights}} & \multicolumn{3}{c}{\textbf{Acc. (\%)}} \\
            \cmidrule(lr){2-4} \cmidrule(lr){5-7}
            \textbf{Config.} & $\lambda_1$ & $\lambda_2$ & $\lambda_3$ & All & Old & New \\
            \midrule
            \textbf{Config. 1} & 0.60 & 0.25 & 0.15 & \textbf{56.78} & \textbf{59.23} & \textbf{53.21} \\
            \textbf{Config. 2} & 0.15 & 0.60 & 0.25 & 52.12 & 53.33 & 50.07 \\
            \textbf{Config. 3} & 0.25 & 0.15 & 0.60 & 51.37 & 52.17 & 50.01 \\
            \bottomrule
        \end{tabular}%
    }
\end{minipage}

\end{table*}

\begin{figure}[ht]
  \centering
  \begin{minipage}[t]{0.48\textwidth}
    \vspace{0pt}  
    \centering
    \includegraphics[width=\linewidth]{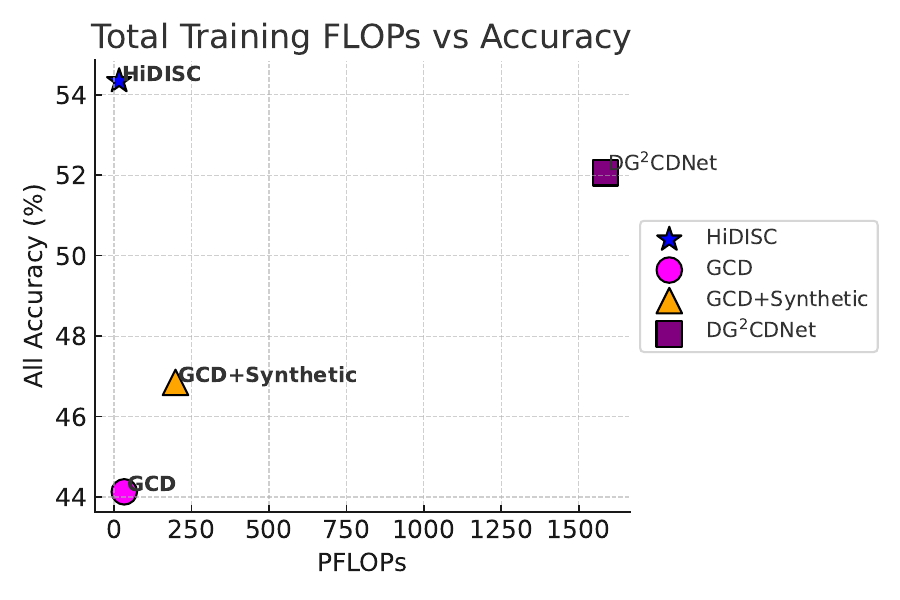}
  \end{minipage}%
  \hfill
  \begin{minipage}[t]{0.48\textwidth}
    \vspace{0pt}  
    \captionof{figure}{
      \textbf{Computational efficiency of {\ourmodel}.} Hyp‐Busemann requires only \textbf{16.53 PFLOPs} over 50 epochs with a batch size of $128\times2$, representing a $\sim2\times$ reduction compared to GCD (33.06 PFLOPs), $\sim12\times$ vs.\ GCD+Synthetic (198.36 PFLOPs), and nearly $\sim96\times$ vs.\ \prevmodel{} (1,586 PFLOPs). Despite this efficiency, Hyp‐Busemann maintains superior accuracy without relying on episodic-training loops, simplifying the overall training pipeline.
    }
    \label{fig:flops}
  \end{minipage}
\end{figure}


\vspace{-3mm}

\section{Takeaways }
\label{sec:takeaways}
\vspace{-2mm}
We addressed the problem of DG-GCD, where novel categories emerge in unseen domains without target supervision. To this end, we proposed \textsc{HiDISC}, a hyperbolic representation learning framework that leverages penalized Busemann alignment, Tangent CutMix-based augmentation, and hybrid contrastive regularization to enable domain- and category-level generalization.
Extensive experiments across PACS, Office-Home, and DomainNet show that \textsc{HiDISC} achieves state-of-the-art performance, particularly improving novel class discovery under domain shift. Our findings underscore the utility of hyperbolic geometry for scalable open-world recognition. \textbf{Future directions} include extending \textsc{HiDISC} to continual DG-GCD and integrating it with large-scale vision-language models.

\textbf{Broader Impact and Limitations:} While \textsc{HiDISC} advances open-world recognition under domain shift using geometry-aware learning, which is extremely practical, its reliance on synthetic augmentations guided by diffusion models may limit applicability in resource-constrained or safety-critical environments where generative artifacts could propagate bias.

\newpage
\begin{ack}
We thank our colleague Shubhranil B. for his assistance with the figures in this paper. 
We are also grateful to Adobe Research and the CMInDS department for providing the necessary resources and support. 
\end{ack}

\bibliographystyle{unsrtnat}
\bibliography{main}

\newpage


\newpage

\section*{NeurIPS Paper Checklist}

The checklist is designed to encourage best practices for responsible machine learning research, addressing issues of reproducibility, transparency, research ethics, and societal impact. Do not remove the checklist: {\bf The papers not including the checklist will be desk rejected.} The checklist should follow the references and follow the (optional) supplemental material.  The checklist does NOT count towards the page
limit. 

Please read the checklist guidelines carefully for information on how to answer these questions. For each question in the checklist:
\begin{itemize}
    \item You should answer \answerYes{}, \answerNo{}, or \answerNA{}.
    \item \answerNA{} means either that the question is Not Applicable for that particular paper or the relevant information is Not Available.
    \item Please provide a short (1–2 sentence) justification right after your answer (even for NA). 
\end{itemize}

{\bf The checklist answers are an integral part of your paper submission.} They are visible to the reviewers, area chairs, senior area chairs, and ethics reviewers. You will be asked to also include it (after eventual revisions) with the final version of your paper, and its final version will be published with the paper.

The reviewers of your paper will be asked to use the checklist as one of the factors in their evaluation. While "\answerYes{}" is generally preferable to "\answerNo{}", it is perfectly acceptable to answer "\answerNo{}" provided a proper justification is given (e.g., "error bars are not reported because it would be too computationally expensive" or "we were unable to find the license for the dataset we used"). In general, answering "\answerNo{}" or "\answerNA{}" is not grounds for rejection. While the questions are phrased in a binary way, we acknowledge that the true answer is often more nuanced, so please just use your best judgment and write a justification to elaborate. All supporting evidence can appear either in the main paper or the supplemental material, provided in appendix. If you answer \answerYes{} to a question, in the justification please point to the section(s) where related material for the question can be found.

IMPORTANT, please:
\begin{itemize}
    \item {\bf Delete this instruction block, but keep the section heading ``NeurIPS Paper Checklist"},
    \item  {\bf Keep the checklist subsection headings, questions/answers and guidelines below.}
    \item {\bf Do not modify the questions and only use the provided macros for your answers}.
\end{itemize}


\begin{enumerate}

\item {\bf Claims}
    \item[] Question: Do the main claims made in the abstract and introduction accurately reflect the paper's contributions and scope?
    \item[] Answer: \answerYes{} 
    \item[] Justification: We state the contributions in the abstract and introduction sections.
    \item[] Guidelines:
    \begin{itemize}
        \item The answer NA means that the abstract and introduction do not include the claims made in the paper.
        \item The abstract and/or introduction should clearly state the claims made, including the contributions made in the paper and important assumptions and limitations. A No or NA answer to this question will not be perceived well by the reviewers. 
        \item The claims made should match theoretical and experimental results, and reflect how much the results can be expected to generalize to other settings. 
        \item It is fine to include aspirational goals as motivation as long as it is clear that these goals are not attained by the paper. 
    \end{itemize}

\item {\bf Limitations}
    \item[] Question: Does the paper discuss the limitations of the work performed by the authors?
    \item[] Answer: \answerYes{} 
    \item[] Justification: We discuss the Limitations in \ref{sec:takeaways}
    \item[] Guidelines:
    \begin{itemize}
        \item The answer NA means that the paper has no limitation while the answer No means that the paper has limitations, but those are not discussed in the paper. 
        \item The authors are encouraged to create a separate "Limitations" section in their paper.
        \item The paper should point out any strong assumptions and how robust the results are to violations of these assumptions (e.g., independence assumptions, noiseless settings, model well-specification, asymptotic approximations only holding locally). The authors should reflect on how these assumptions might be violated in practice and what the implications would be.
        \item The authors should reflect on the scope of the claims made, e.g., if the approach was only tested on a few datasets or with a few runs. In general, empirical results often depend on implicit assumptions, which should be articulated.
        \item The authors should reflect on the factors that influence the performance of the approach. For example, a facial recognition algorithm may perform poorly when image resolution is low or images are taken in low lighting. Or a speech-to-text system might not be used reliably to provide closed captions for online lectures because it fails to handle technical jargon.
        \item The authors should discuss the computational efficiency of the proposed algorithms and how they scale with dataset size.
        \item If applicable, the authors should discuss possible limitations of their approach to address problems of privacy and fairness.
        \item While the authors might fear that complete honesty about limitations might be used by reviewers as grounds for rejection, a worse outcome might be that reviewers discover limitations that aren't acknowledged in the paper. The authors should use their best judgment and recognize that individual actions in favor of transparency play an important role in developing norms that preserve the integrity of the community. Reviewers will be specifically instructed to not penalize honesty concerning limitations.
    \end{itemize}

\item {\bf Theory assumptions and proofs}
    \item[] Question: For each theoretical result, does the paper provide the full set of assumptions and a complete (and correct) proof?
    \item[] Answer: \answerYes{} 
    \item[] Justification: The paper includes the full set of assumptions and a complete (and correct) proof in Section \ref{sec:theoretical_proof}, with detailed extensions provided in the \textbf{Sup. Mat.}
    \item[] Guidelines:
    \begin{itemize}
        \item The answer NA means that the paper does not include theoretical results. 
        \item All the theorems, formulas, and proofs in the paper should be numbered and cross-referenced.
        \item All assumptions should be clearly stated or referenced in the statement of any theorems.
        \item The proofs can either appear in the main paper or the supplemental material, but if they appear in the supplemental material, the authors are encouraged to provide a short proof sketch to provide intuition. 
        \item Inversely, any informal proof provided in the core of the paper should be complemented by formal proofs provided in appendix or supplemental material.
        \item Theorems and Lemmas that the proof relies upon should be properly referenced. 
    \end{itemize}

    \item {\bf Experimental result reproducibility}
    \item[] Question: Does the paper fully disclose all the information needed to reproduce the main experimental results of the paper to the extent that it affects the main claims and/or conclusions of the paper (regardless of whether the code and data are provided or not)?
    \item[] Answer: \answerYes{} 
    \item[] Justification: For details on reproducing the experimental results, please refer to section \ref{sec:experimental_settings} and for further details refer \textbf{Sup. Mat.}
    \item[] Guidelines:
    \begin{itemize}
        \item The answer NA means that the paper does not include experiments.
        \item If the paper includes experiments, a No answer to this question will not be perceived well by the reviewers: Making the paper reproducible is important, regardless of whether the code and data are provided or not.
        \item If the contribution is a dataset and/or model, the authors should describe the steps taken to make their results reproducible or verifiable. 
        \item Depending on the contribution, reproducibility can be accomplished in various ways. For example, if the contribution is a novel architecture, describing the architecture fully might suffice, or if the contribution is a specific model and empirical evaluation, it may be necessary to either make it possible for others to replicate the model with the same dataset, or provide access to the model. In general. releasing code and data is often one good way to accomplish this, but reproducibility can also be provided via detailed instructions for how to replicate the results, access to a hosted model (e.g., in the case of a large language model), releasing of a model checkpoint, or other means that are appropriate to the research performed.
        \item While NeurIPS does not require releasing code, the conference does require all submissions to provide some reasonable avenue for reproducibility, which may depend on the nature of the contribution. For example
        \begin{enumerate}
            \item If the contribution is primarily a new algorithm, the paper should make it clear how to reproduce that algorithm.
            \item If the contribution is primarily a new model architecture, the paper should describe the architecture clearly and fully.
            \item If the contribution is a new model (e.g., a large language model), then there should either be a way to access this model for reproducing the results or a way to reproduce the model (e.g., with an open-source dataset or instructions for how to construct the dataset).
            \item We recognize that reproducibility may be tricky in some cases, in which case authors are welcome to describe the particular way they provide for reproducibility. In the case of closed-source models, it may be that access to the model is limited in some way (e.g., to registered users), but it should be possible for other researchers to have some path to reproducing or verifying the results.
        \end{enumerate}
    \end{itemize}

\item {\bf Open access to data and code}
    \item[] Question: Does the paper provide open access to the data and code, with sufficient instructions to faithfully reproduce the main experimental results, as described in supplemental material?
    \item[] Answer: \answerYes{} 
    \item[] Justification: Please refer to Section ~\ref{sec:experimental_settings} for the datasets used , all are publicly available. Further code will be available on the project website.
    \item[] Guidelines:
    \begin{itemize}
        \item The answer NA means that paper does not include experiments requiring code.
        \item Please see the NeurIPS code and data submission guidelines (\url{https://nips.cc/public/guides/CodeSubmissionPolicy}) for more details.
        \item While we encourage the release of code and data, we understand that this might not be possible, so “No” is an acceptable answer. Papers cannot be rejected simply for not including code, unless this is central to the contribution (e.g., for a new open-source benchmark).
        \item The instructions should contain the exact command and environment needed to run to reproduce the results. See the NeurIPS code and data submission guidelines (\url{https://nips.cc/public/guides/CodeSubmissionPolicy}) for more details.
        \item The authors should provide instructions on data access and preparation, including how to access the raw data, preprocessed data, intermediate data, and generated data, etc.
        \item The authors should provide scripts to reproduce all experimental results for the new proposed method and baselines. If only a subset of experiments are reproducible, they should state which ones are omitted from the script and why.
        \item At submission time, to preserve anonymity, the authors should release anonymized versions (if applicable).
        \item Providing as much information as possible in supplemental material (appended to the paper) is recommended, but including URLs to data and code is permitted.
    \end{itemize}

\item {\bf Experimental setting/details}
    \item[] Question: Does the paper specify all the training and test details (e.g., data splits, hyperparameters, how they were chosen, type of optimizer, etc.) necessary to understand the results?
    \item[] Answer: \answerYes{}
    \item[] Justification: Please refer to Section \ref{sec:experimental_settings} for experimental settings about the data splits , test details , training details(in Sup Mat) .
    \item[] Guidelines:
    \begin{itemize}
        \item The answer NA means that the paper does not include experiments.
        \item The experimental setting should be presented in the core of the paper to a level of detail that is necessary to appreciate the results and make sense of them.
        \item The full details can be provided either with the code, in appendix, or as supplemental material.
    \end{itemize}

\item {\bf Experiment statistical significance}
    \item[] Question: Does the paper report error bars suitably and correctly defined or other appropriate information about the statistical significance of the experiments?
    \item[] Answer: \answerYes{} 
    \item[] Justification: The paper carefully reports the statistical significance details in Section \ref{sec:experimental_settings}, following standard practices in the literature.
    \item[] Guidelines:
    \begin{itemize}
        \item The answer NA means that the paper does not include experiments.
        \item The authors should answer "Yes" if the results are accompanied by error bars, confidence intervals, or statistical significance tests, at least for the experiments that support the main claims of the paper.
        \item The factors of variability that the error bars are capturing should be clearly stated (for example, train/test split, initialization, random drawing of some parameter, or overall run with given experimental conditions).
        \item The method for calculating the error bars should be explained (closed form formula, call to a library function, bootstrap, etc.)
        \item The assumptions made should be given (e.g., Normally distributed errors).
        \item It should be clear whether the error bar is the standard deviation or the standard error of the mean.
        \item It is OK to report 1-sigma error bars, but one should state it. The authors should preferably report a 2-sigma error bar than state that they have a 96\% CI, if the hypothesis of Normality of errors is not verified.
        \item For asymmetric distributions, the authors should be careful not to show in tables or figures symmetric error bars that would yield results that are out of range (e.g. negative error rates).
        \item If error bars are reported in tables or plots, The authors should explain in the text how they were calculated and reference the corresponding figures or tables in the text.
    \end{itemize}

\item {\bf Experiments compute resources}
    \item[] Question: For each experiment, does the paper provide sufficient information on the computer resources (type of compute workers, memory, time of execution) needed to reproduce the experiments?
    \item[] Answer: \answerYes{} 
    \item[] Justification: All experiments are done on NVIDIA A100-SMX-80GB GPU's.
    \item[] Guidelines:
    \begin{itemize}
        \item The answer NA means that the paper does not include experiments.
        \item The paper should indicate the type of compute workers CPU or GPU, internal cluster, or cloud provider, including relevant memory and storage.
        \item The paper should provide the amount of compute required for each of the individual experimental runs as well as estimate the total compute. 
        \item The paper should disclose whether the full research project required more compute than the experiments reported in the paper (e.g., preliminary or failed experiments that didn't make it into the paper). 
    \end{itemize}
    
\item {\bf Code of ethics}
    \item[] Question: Does the research conducted in the paper conform, in every respect, with the NeurIPS Code of Ethics \url{https://neurips.cc/public/EthicsGuidelines}?
    \item[] Answer: \answerYes{} 
    \item[] Justification: We confirm the research conducted in the paper conform, in every respect,
with the NeurIPS Code of Ethics.
    \item[] Guidelines:
    \begin{itemize}
        \item The answer NA means that the authors have not reviewed the NeurIPS Code of Ethics.
        \item If the authors answer No, they should explain the special circumstances that require a deviation from the Code of Ethics.
        \item The authors should make sure to preserve anonymity (e.g., if there is a special consideration due to laws or regulations in their jurisdiction).
    \end{itemize}

\item {\bf Broader impacts}
    \item[] Question: Does the paper discuss both potential positive societal impacts and negative societal impacts of the work performed?
    \item[] Answer: \answerYes{} 
    \item[] Justification:  We discuss the broader impacts of our work in Section \ref{sec:takeaways}
    \item[] Guidelines:
    \begin{itemize}
        \item The answer NA means that there is no societal impact of the work performed.
        \item If the authors answer NA or No, they should explain why their work has no societal impact or why the paper does not address societal impact.
        \item Examples of negative societal impacts include potential malicious or unintended uses (e.g., disinformation, generating fake profiles, surveillance), fairness considerations (e.g., deployment of technologies that could make decisions that unfairly impact specific groups), privacy considerations, and security considerations.
        \item The conference expects that many papers will be foundational research and not tied to particular applications, let alone deployments. However, if there is a direct path to any negative applications, the authors should point it out. For example, it is legitimate to point out that an improvement in the quality of generative models could be used to generate deepfakes for disinformation. On the other hand, it is not needed to point out that a generic algorithm for optimizing neural networks could enable people to train models that generate Deepfakes faster.
        \item The authors should consider possible harms that could arise when the technology is being used as intended and functioning correctly, harms that could arise when the technology is being used as intended but gives incorrect results, and harms following from (intentional or unintentional) misuse of the technology.
        \item If there are negative societal impacts, the authors could also discuss possible mitigation strategies (e.g., gated release of models, providing defenses in addition to attacks, mechanisms for monitoring misuse, mechanisms to monitor how a system learns from feedback over time, improving the efficiency and accessibility of ML).
    \end{itemize}
    
\item {\bf Safeguards}
    \item[] Question: Does the paper describe safeguards that have been put in place for responsible release of data or models that have a high risk for misuse (e.g., pretrained language models, image generators, or scraped datasets)?
    \item[] Answer: \answerNA{} 
    \item[] Justification: This work poses no such risks.
    \item[] Guidelines:
    \begin{itemize}
        \item The answer NA means that the paper poses no such risks.
        \item Released models that have a high risk for misuse or dual-use should be released with necessary safeguards to allow for controlled use of the model, for example by requiring that users adhere to usage guidelines or restrictions to access the model or implementing safety filters. 
        \item Datasets that have been scraped from the Internet could pose safety risks. The authors should describe how they avoided releasing unsafe images.
        \item We recognize that providing effective safeguards is challenging, and many papers do not require this, but we encourage authors to take this into account and make a best faith effort.
    \end{itemize}

\item {\bf Licenses for existing assets}
    \item[] Question: Are the creators or original owners of assets (e.g., code, data, models), used in the paper, properly credited and are the license and terms of use explicitly mentioned and properly respected?
    \item[] Answer: \answerYes{}
    \item[] Justification:  We cite all datasets and models utilized in our experiments.
    \item[] Guidelines:
    \begin{itemize}
        \item The answer NA means that the paper does not use existing assets.
        \item The authors should cite the original paper that produced the code package or dataset.
        \item The authors should state which version of the asset is used and, if possible, include a URL.
        \item The name of the license (e.g., CC-BY 4.0) should be included for each asset.
        \item For scraped data from a particular source (e.g., website), the copyright and terms of service of that source should be provided.
        \item If assets are released, the license, copyright information, and terms of use in the package should be provided. For popular datasets, \url{paperswithcode.com/datasets} has curated licenses for some datasets. Their licensing guide can help determine the license of a dataset.
        \item For existing datasets that are re-packaged, both the original license and the license of the derived asset (if it has changed) should be provided.
        \item If this information is not available online, the authors are encouraged to reach out to the asset's creators.
    \end{itemize}

\item {\bf New assets}
    \item[] Question: Are new assets introduced in the paper well documented and is the documentation provided alongside the assets?
    \item[] Answer: \answerNA{} 
    \item[] Justification:  This work does not release new assets.
    \item[] Guidelines:
    \begin{itemize}
        \item The answer NA means that the paper does not release new assets.
        \item Researchers should communicate the details of the dataset/code/model as part of their submissions via structured templates. This includes details about training, license, limitations, etc. 
        \item The paper should discuss whether and how consent was obtained from people whose asset is used.
        \item At submission time, remember to anonymize your assets (if applicable). You can either create an anonymized URL or include an anonymized zip file.
    \end{itemize}

\item {\bf Crowdsourcing and research with human subjects}
    \item[] Question: For crowdsourcing experiments and research with human subjects, does the paper include the full text of instructions given to participants and screenshots, if applicable, as well as details about compensation (if any)? 
    \item[] Answer: \answerNA{} 
    \item[] Justification:  This work does not involve crowdsourcing nor research with human subjects.
    \item[] Guidelines:
    \begin{itemize}
        \item The answer NA means that the paper does not involve crowdsourcing nor research with human subjects.
        \item Including this information in the supplemental material is fine, but if the main contribution of the paper involves human subjects, then as much detail as possible should be included in the main paper. 
        \item According to the NeurIPS Code of Ethics, workers involved in data collection, curation, or other labor should be paid at least the minimum wage in the country of the data collector. 
    \end{itemize}

\item {\bf Institutional review board (IRB) approvals or equivalent for research with human subjects}
    \item[] Question: Does the paper describe potential risks incurred by study participants, whether such risks were disclosed to the subjects, and whether Institutional Review Board (IRB) approvals (or an equivalent approval/review based on the requirements of your country or institution) were obtained?
    \item[] Answer: \answerNA{} 
    \item[] Justification: This work does not involve crowdsourcing nor research with human subjects.
    \item[] Guidelines:
    \begin{itemize}
        \item The answer NA means that the paper does not involve crowdsourcing nor research with human subjects.
        \item Depending on the country in which research is conducted, IRB approval (or equivalent) may be required for any human subjects research. If you obtained IRB approval, you should clearly state this in the paper. 
        \item We recognize that the procedures for this may vary significantly between institutions and locations, and we expect authors to adhere to the NeurIPS Code of Ethics and the guidelines for their institution. 
        \item For initial submissions, do not include any information that would break anonymity (if applicable), such as the institution conducting the review.
    \end{itemize}

\item {\bf Declaration of LLM usage}
    \item[] Question: Does the paper describe the usage of LLMs if it is an important, original, or non-standard component of the core methods in this research? Note that if the LLM is used only for writing, editing, or formatting purposes and does not impact the core methodology, scientific rigorousness, or originality of the research, declaration is not required.
    \item[] Answer: \answerYes{} 
    \item[] Justification: The Gpt 4-o usage is described in Section \ref{sec:methodology}
    \item[] Guidelines:
    \begin{itemize}
        \item The answer NA means that the core method development in this research does not involve LLMs as any important, original, or non-standard components.
        \item Please refer to our LLM policy (\url{https://neurips.cc/Conferences/2025/LLM}) for what should or should not be described.
    \end{itemize}

\end{enumerate}

\newpage


\centerline{\textbf{\LARGE \ourmodel: Supplementary Material}}

This supplementary document provides additional technical details, experimental results, and theoretical insights to complement the main paper. Specifically, it includes:

\begin{itemize}
  \item \textbf{Quantitative and qualitative analyses} comparing Euclidean (spherical) and hyperbolic (Poincaré) embedding spaces, including inter-class separation and intra-class variability metrics (Sec.~\ref{sec:cosine_similarity}, Sec.~\ref{sec:separation_variability}).
  \item \textbf{Mathematical background} on Poincaré ball geometry and key hyperbolic operations used in our model (Sec.~\ref{sup:poincare_geometry}).
  \item \textbf{Details on synthetic data generation}, illustrating the lightweight augmentation pipeline and example images (Sec.~\ref{sec:synthetic_data}, Sec.~\ref{sec:pacs_images}).
  \item \textbf{Comprehensive ablations} exploring the effects of synthetic domain count, adaptive outlier margin, embedding dimension, curvature learning, and domain combinations on model performance (Sec.~\ref{sup:ablation_synthetic_domains}–\ref{sec:ablation_dimension}, Sec.~\ref{sup:curvature_evolution}).
  \item \textbf{Theoretical justification} outlining generalization bounds and domain discrepancy advantages of hyperbolic embeddings over Euclidean embeddings (Sec.~\ref{sup:theory_generalization}).
  \item \textbf{Extended comparisons} against state-of-the-art methods on three major benchmarks (PACS, Office-Home, DomainNet), with detailed accuracy tables and discussions (Sec.~\ref{sec:comparative_analysis}).
  \item \textbf{Additional visualization results} demonstrating embedding geometry and augmentation effects in hyperbolic space.
\end{itemize}

Together, these materials reinforce the core claims of our paper and provide the necessary context and evidence for reproducibility and deeper understanding.

\newpage
\section{Quantitative Comparison of Spherical vs. Hyperbolic Embeddings}
\label{sec:cosine_similarity}

To quantitatively support the qualitative illustration presented in Figure~1 of the main paper, we compare feature alignment in \textbf{Euclidean} (spherical) and \textbf{Hyperbolic} (Poincaré) embedding spaces across domains. Specifically, we evaluate the semantic consistency of same-class embeddings between two distinct domains—\textit{Art Painting} and \textit{Sketch}—from the \textbf{PACS} dataset.

We extract features using a frozen \textbf{DINO ViT-B16} \cite{dino} backbone and compute cross-domain similarity scores for each class as follows:

\begin{itemize}
    \item \textbf{Euclidean Embedding}: Features are L2-normalized, and cosine similarity is computed via dot product between class-averaged embeddings from each domain.
    \item \textbf{Hyperbolic Embedding}: The same features are projected to the Poincaré ball via exponential mapping, and cosine similarity is computed in the tangent space, respecting hyperbolic geometry.
\end{itemize}

As shown in Table~\ref{tab:cosine_similarity}, same-class embeddings exhibit consistently high alignment in hyperbolic space (cosine similarity $\approx$ 0.98–0.99) across all classes, whereas their Euclidean counterparts show substantially lower similarity. This demonstrates that hyperbolic space facilitates superior cross-domain semantic consistency, even in the presence of strong domain shifts.

\begin{table}[htbp]
    \centering
    \begin{tabular}{ccc}
        \toprule
        \textbf{Class} & \textbf{Euclidean Similarity} & \textbf{Hyperbolic Similarity} \\
        \midrule
        Horse & 0.1409 & 0.9812 \\
        Elephant & 0.3350 & 0.9904 \\
        House & 0.2418 & 0.9857 \\
        Dog & 0.3091 & 0.9869 \\
        Guitar & 0.3079 & 0.9833 \\
        Person & 0.3459 & 0.9891 \\
        Giraffe & 0.1439 & 0.9805 \\
        \bottomrule
    \end{tabular}
    \caption{Cosine similarity between class-wise average embeddings across \textit{Art Painting} and \textit{Sketch} domains in Euclidean and Hyperbolic spaces. Hyperbolic embeddings exhibit significantly improved semantic alignment.}
    \label{tab:cosine_similarity}
\end{table}

These results validate our claim that the hyperbolic space enables improved inter-class separation and cross-domain semantic alignment, thereby providing an ideal geometric foundation for domain generalization and generalized category discovery (DG-GCD).


\section{Inter-Class Separation and Intra-Class Variability}
\label{sec:separation_variability}

To further validate our theoretical motivation in lines 134–138 of the main paper, we evaluate both global semantic separation and local compactness in Euclidean vs. hyperbolic spaces.

We compute:

\begin{itemize}
    \item \textbf{Inter-Class Distance (ICD)}: Mean cosine distance between class centroids.
    \item \textbf{Intra-Class Variability (ICV)}: Standard deviation of cosine similarity within each class across domains.
\end{itemize}

\vspace{-0.75em}
\begin{table}[htbp]
    \centering
    \begin{tabular}{lcc}
        \toprule
        \textbf{Metric} & \textbf{Euclidean} & \textbf{Hyperbolic} \\
        \midrule
        Inter-Class Distance (↑) & 0.45 & \textbf{2.03} \\
        Intra-Class Variability (↓) & 0.2964 & \textbf{0.10} \\
        \bottomrule
    \end{tabular}
    \caption{Comparison of inter-class separation and intra-class variability across embedding spaces. Hyperbolic space provides superior class separation and compactness.}
    \label{tab:icd_icv}
\end{table}

These results show that hyperbolic embeddings support both global semantic separation and local compactness, as hypothesized. Higher inter-class distance amplifies semantic dissimilarity, while reduced intra-class variability reflects robustness to domain-specific low-level perturbations.


\section{Poincaré Ball Geometry}
\label{sup:poincare_geometry}

In the main text (Sec.~2.1) we define the $n$-dimensional Poincaré ball of curvature $-c^2$ as
\[
  \mathbb{D}^n_c \;=\;\bigl\{\,x\in R^n : c\,\|x\|^2 < 1\bigr\},
  \qquad
  \|x\|=\sqrt{x^\top x}.
\]
Below we collect the key operations used in our implementation.

\paragraph{Möbius Addition.}
For any $a,b\in\mathbb{D}^n_c$ the Möbius sum is
\[
  a \oplus_c b 
  = \frac{(1 + 2\,c\,\langle a,b\rangle + c\,\|b\|^2)\,a \;+\;(1 - c\,\|a\|^2)\,b}
         {1 + 2\,c\,\langle a,b\rangle + c^2\,\|a\|^2\,\|b\|^2}\,.
\]

\paragraph{Geodesic Distance.}
The Riemannian distance on $\mathbb{D}^n_c$ is
\[
  d_{\mathbb{D}_c}(a,b)
  = \frac{2}{\sqrt{c}}\,
    \tanh\!\bigl(\sqrt{c}\,\|{-}a\oplus_c b\|\bigr).
\]

\paragraph{Exponential and Logarithmic Maps at the Origin.}
To move between Euclidean tangent vectors and the manifold:
\[
  \exp^c_0(u)
  = \tanh\!\bigl(\sqrt{c}\,\|u\|\bigr)\,
    \frac{u}{\sqrt{c}\,\|u\|},
  \quad
  \log^c_0(x)
  = \tanh\!\bigl(\sqrt{c}\,\|x\|\bigr)\,
    \frac{x}{\sqrt{c}\,\|x\|},
\]
for all $u\in R^n$ and $x\in\mathbb{D}^n_c$.  These correspond to our `expmap0(u,c)` and its inverse in code.

\paragraph{Euclidean Limit.}
As $c\to0$, hyperbolic operations recover their Euclidean counterparts:
\[
  \lim_{c\to0} d_{\mathbb{D}_c}(a,b) = 2\,\|a-b\|,
  \quad
  \lim_{c\to0} \exp^c_0(u) = u,
  \quad
  \lim_{c\to0} (a\oplus_c b) = a + b.
\]
This ensures full compatibility with our Euclidean baseline.

\section{Technical details about Synthetic Data Generation} 
\label{sec:synthetic_data}
To implement the lightweight synthetic domain augmentation described in Sec.~3.3.1 of the main paper,following \cite{dg2net} for the image transformation tasks, we utilized the InstructPix2Pix pipeline from Hugging Face's Diffusers library, leveraging the \texttt{timbrooks/instruct-} \texttt{pix2pix} model, for image transformation tasks.To balance processing time with output quality, we set the  \texttt{num\_inference\_steps} parameter to 10. The \texttt{image\_guidance\_scale} parameter was set to 1.0, e, ensuring the model preserved the essential structure of the input image while applying the desired transformations. Additionally, the \texttt{guidance\_scale} was set to 7.5 to enhance the alignment with the transformation prompt. These settings enable the easy replication of our process while ensuring high-quality results.

\subsection{Synthetic Domain Images}
\label{sec:pacs_images}
\begin{figure}
    \centering
    \includegraphics[width=0.95\linewidth]{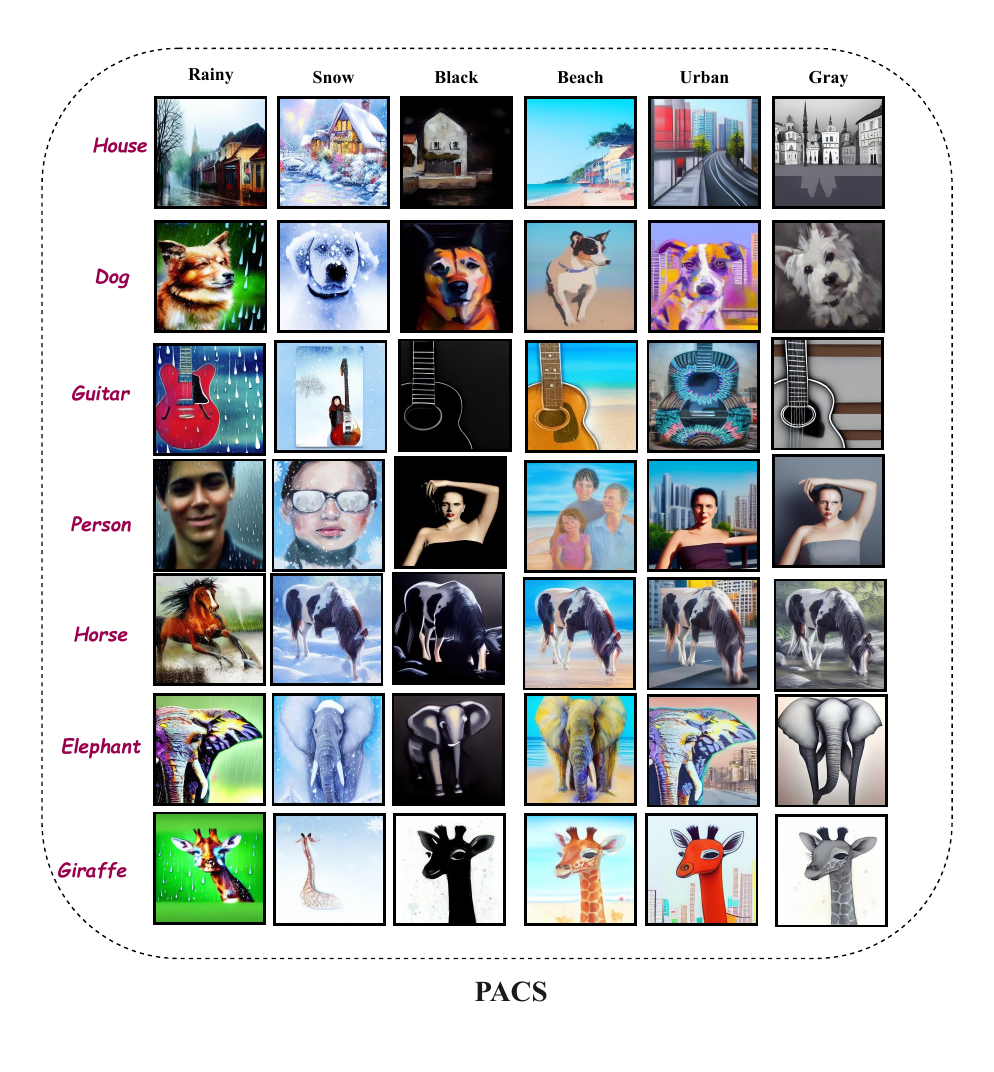}
    \caption{Examples of synthetic images generated for the PACS dataset, demonstrating various styles and domains across seven different categories.}
    \label{fig:pacs_synthetic}
\end{figure}

In Figure \ref{fig:pacs_synthetic}, we present a few examples of the synthetic images generated using the strategy outlined above for synthetic data generation.

\subsection*{On the Role of GPT-Guided Diffusion for Domain Augmentation}

Although GPT-guided diffusion does not sample uniformly over the domain manifold, uniform coverage is neither tractable nor necessary for DG-GCD. Instead, HiDISC leverages task-aware, semantically structured augmentations that improve generalization:

\begin{itemize}[leftmargin=*,noitemsep]
    \item \textbf{Task-aware domain coverage.} Let $\mathcal{D} \subset \mathcal{X}$ denote the latent domain manifold. Uniform sampling $P(x) \sim \mathcal{U}(\mathcal{D})$ is intractable in high-dimensional image spaces. We instead construct a conditional distribution $P_G(x \mid t)$ using prompts $t$ aligned with meaningful domain shifts, yielding diverse and relevant perturbations.
    
    \item \textbf{Empirical diminishing returns.} As shown in Fig.~3 (right), adding synthetic domains beyond 1–2 does not yield monotonic gains. Formally, the curvature of the loss $\mathcal{L}_{\text{gen}}$ satisfies $\frac{\partial^2 \mathcal{L}_{\text{gen}}}{\partial N^2} > 0$ for $N > 2$, indicating diminishing returns and potential overfitting.
    
    \item \textbf{Hyperbolic geometry enhances generalization.} Even without synthetic domains, HiDISC achieves 56.07\% accuracy on Office-Home, surpassing Euclidean and augmented baselines. Theoretical analysis (Sec.~3.4) further shows that the generalization bound in hyperbolic space satisfies $\Delta_{\mathbb{H}} \leq \Delta_{\mathbb{E}}$, implying fewer samples are required to cover semantic variability.
\end{itemize}

In summary, GPT-guided augmentation supplies diverse but semantically structured shifts, while hyperbolic geometry ensures robust generalization without requiring exhaustive domain sampling.

\section{Ablations}
\subsection{Number of Synthetic Domains}
\label{sup:ablation_synthetic_domains}

We quantify the effect of our lightweight synthetic domain augmentation (Sec.~3.3.1 in the main paper) by varying the number of GPT-4o–guided synthetic domains added per training image.  Using the Office-Home benchmark, we train with $k\in\{0,\dots,6\}$ synthetic domains in addition to the original source data and report three metrics.

\begin{figure}[!ht]
    \centering
    \begin{tikzpicture}
    \begin{axis}[
        width=0.85\linewidth,
        height=6cm,
        xlabel={\# Synthetic Domains ($k$)},
        ylabel={Accuracy (\%)},
        xmin=0, xmax=6,
        ymin=45, ymax=75,
        xtick={0,1,2,3,4,5,6},
        grid=major,
        legend pos=south east,
        cycle list name=color list,
    ]
    \addplot+[mark=*, thick]   coordinates {(0,56.07) (1,57.06) (2,56.78) (3,57.84) (4,58.12) (5,60.39) (6,60.55)};
    \addplot+[mark=square*, thick] coordinates {(0,59.29) (1,59.71) (2,59.23) (3,60.43) (4,62.97) (5,63.60) (6,67.78)};
    \addplot+[mark=triangle*, thick] coordinates {(0,50.67) (1,53.09) (2,53.21) (3,50.79) (4,51.01) (5,52.38) (6,48.16)};
    \legend{All, Old, New}
    \end{axis}
    \end{tikzpicture}
    \caption{Ablation on the number of synthetic domains per image ($k$) for the Office-Home dataset.  Adding up to six synthetic styles steadily improves overall (\emph{All}) and seen-class (\emph{Old}) accuracies, while novel-class (\emph{New}) performance peaks at $k=2$ before declining.}
    \label{fig:synthetic_domains_ablation_line_sup}
\end{figure}
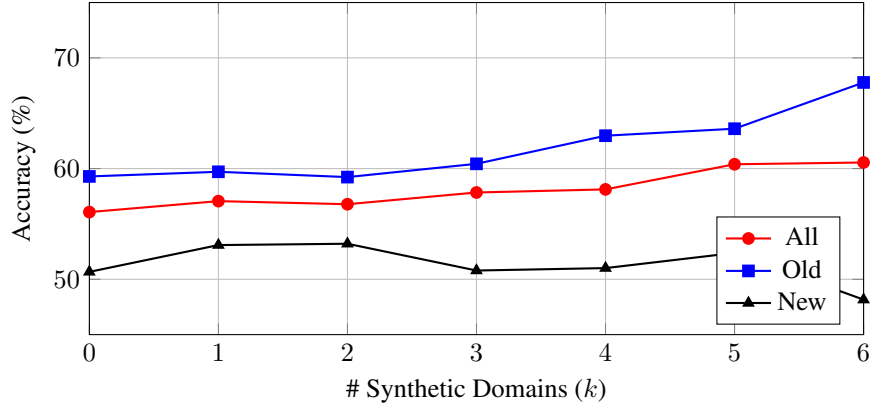

Figure~\ref{fig:synthetic_domains_ablation_line_sup} plots these metrics as a function of $k$.  We observe the following.
\begin{enumerate}
  \item \emph{Monotonic gains} in \textbf{All} and \textbf{Old} accuracies up to $k=6$, demonstrating that more diverse styles in hyperbolic space reinforce domain-invariant feature learning.
  \item \textbf{New}–class accuracy peaks at $k=2$ (53.21\%) before declining, indicating that excessive augmentation may oversaturate the model and hinder novel category discrimination.
\end{enumerate}
Overall, $k=2$ supplies the best trade-off between known-class retention and novel-class discovery, and is used in our main experiments.

\subsection{Adaptive Outlier Margin.}
\label{sec:outlier_margin}
To set a robust margin for the adaptive outlier loss (Eq.~3 in the main paper) we compute the margin \(\gamma\) dynamically on the first training mini‐batch as follows.  Let
\[
  d_i \;=\;\min_{k\in\mathcal{Y}_s}\;d_{\mathbb{D}_c}(z_i,\,p_k)
  \quad\text{for each embedding }z_i
\]
where \(p_k\) are the seen‐class prototypes.  We then define
\[
  \gamma \;=\;\mathrm{Quantile}_{0.8}\bigl(\{d_i\}\bigr),
\]
i.e.\ the 80th percentile of the per‐sample minimum distances.  Using this batch‐wise adaptive margin ensures that (1) \(\gamma\) reflects the actual spread of the feature distribution, (2) outlier penalties remain neither too weak nor excessively strict, and (3) the model stays discriminative and resilient to noisy embeddings.  In practice, \(\gamma\) is recomputed once at the start of training and held fixed thereafter.```

We study how the choice of the quantile \(\alpha\) for setting the adaptive outlier margin \(\gamma\) (see Sec.~3.3.3) affects novel‐class discovery on the PACS dataset.  For each \(\alpha\in\{0.05,0.10,0.20,0.80,0.90\}\), we compute \(\gamma\) as the \(\alpha\)-quantile of per‐sample minimum prototype distances on the first training batch, then hold \(\gamma\) fixed throughout training.

\begin{table}[ht]
  \centering
  \caption{New (\%) on PACS for different quantile settings \(\alpha\).}
  \label{tab:margin_quantile_ablation}
  \begin{tabular}{c|ccccc}
    \toprule
    Quantile \(\alpha\) & 0.05 & 0.10 & 0.20 & 0.80 & 0.90 \\
    \midrule
    New (\%)         & 72.1 & 73.4 & 74.1 & \textbf{74.52 } & 74.31  \\
    \bottomrule
  \end{tabular}
\end{table}

As Table~\ref{tab:margin_quantile_ablation} shows, moderate quantiles (\(\alpha=0.80\)) yield the best novel‐class discrimination .

\subsection{Poincaré‐Disk Visualizations of Tangent CutMix on PACS}
\label{sec:cutmix-pacs-visualization}

\begin{figure}[!ht]
  \centering
  \includegraphics[width=\textwidth]{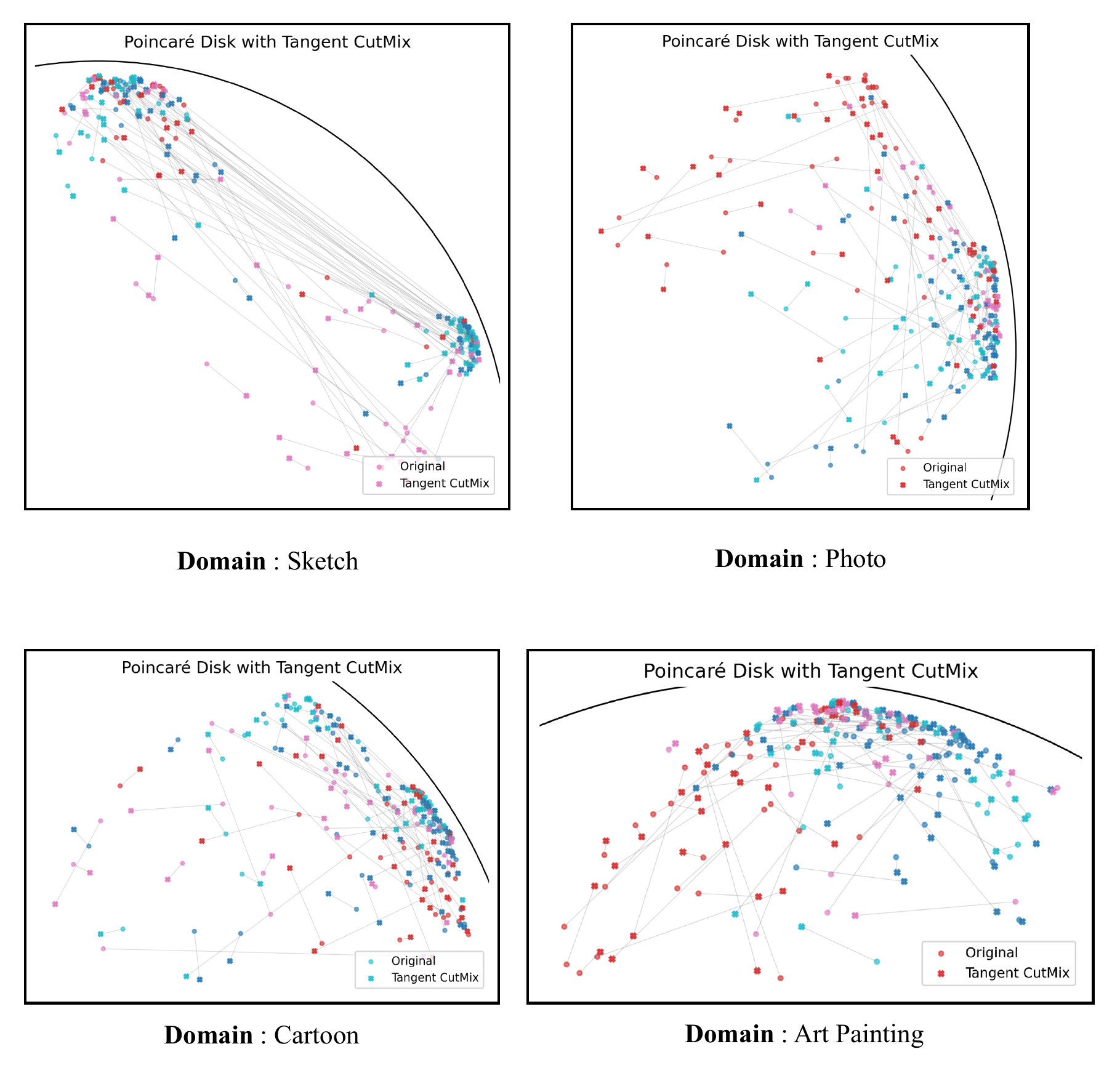}
  \caption{
    \textbf{UMAP‐projected Poincaré‐disk embeddings of original and Tangent CutMix samples on PACS.}
    We first embed features into the Poincaré ball and then apply UMAP to project onto 2D for visualization. Each subplot corresponds to one PACS domain (Sketch, Photo, Cartoon, Art Painting), showing four seen‐class prototypes (solid circles) and their Tangent CutMix augmentations (cross markers). Grey lines link each synthetic point back to its original partners, highlighting curvature‐aware mixing in the tangent space at the origin.
  }
  \label{fig:cutmix-pacs-umpapoincare}
\end{figure}

For each domain, we plot the normalized feature prototypes of four seen classes (solid circles) alongside their Tangent CutMix augmentations (cross markers), all projected via UMAP from the hyperbolic Poincaré ball. Grey geodesic‐consistent connectors illustrate the tangent‐space interpolation at the origin. Across Sketch and Photo (top row), mixed samples fan out toward the periphery—regions of higher curvature—thereby expanding open space between clusters. Similarly, in Cartoon and Art Painting (bottom row), synthetic embeddings occupy under‐populated areas while respecting manifold geometry. These UMAP‐projected Poincaré visualizations demonstrate how Tangent CutMix generates out‐of‐distribution samples and reserves hyperbolic “reserve space” for novel‐class discovery without collapsing into known regions.

\subsection{Redundant Augmentations and Diminishing Returns}
\label{sec:supp-redundant}

We call an augmentation “redundant” once it fails to introduce any novel variation in hyperbolic feature space beyond what our existing synthetic domains already cover.  In practice we see that after adding one or two well‐chosen domains , further domains lie in the same regions of the Poincaré ball (high pairwise FID Table~\ref{tab:fid_pairs_reduced}) and thus bring negligible gains in novel‐class discovery.  

\begin{table}[!ht]
\centering
\caption{Pairwise FID between synthetic domain pairs.}
\label{tab:fid_pairs_reduced}
\scriptsize
\setlength{\tabcolsep}{10pt}
\renewcommand{\arraystretch}{1.1}
\begin{tabular}{r ll c}
\toprule
\textbf{Idx} & \textbf{Domain D1} & \textbf{Domain D2} & \textbf{FID(D1,D2)} \\
\midrule
1  & Rainy   & Beach  & 123.33 \\
2  & Rainy   & Black  & 133.74 \\
3  & Rainy   & Gray   & 138.74 \\
4  & \textbf{Rainy}   & \textbf{Urban} & \textbf{158.24} \\
5  & Rainy   & Snow   &  71.47 \\
6  & Beach   & Black  & 118.41 \\
7  & Beach   & Gray   &  93.15 \\
8  & Beach   & Urban  & 120.73 \\
9  & Beach   & Snow   &  89.21 \\
10 & Black   & Gray   &  57.07 \\
11 & Black   & Urban  & 123.01 \\
12 & Black   & Snow   & 107.16 \\
13 & Gray    & Urban  & 107.78 \\
14 & Gray    & Snow   &  94.53 \\
15 & Urban   & Snow   & 127.33 \\
\bottomrule
\end{tabular}
\end{table}

\subsubsection{Effect of other domain combinations on performance}

\begin{table}[!ht]
  \centering
  \caption{Effect of domain‐combination paths on Office‐Home}
  \label{tab:paths}
  \scriptsize
  \setlength{\tabcolsep}{10pt}
  \renewcommand{\arraystretch}{1.1}
  \begin{tabular}{r ccc}
    \toprule
    \textbf{Idx} & \textbf{All (\%)} & \textbf{Old (\%)} & \textbf{New (\%)} \\
    \midrule
    1  & 57.34    &  60.96   & 51.20   \\
    2  & 58.34    & 61.99   & 52.04   \\
    3  & 57.22    & 61.25    & 50.38   \\
    4  & \textbf{56.78} & \textbf{59.23} & \textbf{53.21} \\
    5  &  57.71   &  62.21   & 50.06   \\
    6  &   57.55  &  62.12   & 49.79   \\
    7  &   58.93  &  63.14   &  51.85  \\
    8  &   58.08  &  63.93   &  48.37  \\
    9  &   56.42  &  59.41   &  51.31  \\
    10  &   58.62  & 63.27   &  50.79  \\
    
    \bottomrule
  \end{tabular}
\end{table}

\paragraph{Adaptive Balancing of Angular and Distance-based Losses.} 
The parameter \(\alpha_d\) to balance the contributions of distance-based and angle-based loss components during training. This parameter is scheduled to increase linearly with training progress and is defined as:
\[
\alpha_d = \frac{e \cdot \alpha_d^{\text{max}}}{50},
\]
where \(e\) denotes the current training epoch, and \(\alpha_d^{\text{max}}\) is the maximum weight assigned to the angular loss. We set \(\alpha_d^{\text{max}} = 1.0\) for all three datasets. This scheduling encourages the model to initially emphasize structural separation through distance-based learning and gradually shift focus toward angular alignment, improving cluster compactness and generalization over time.

\begin{table}[!ht]
  \centering
  \caption{Ablation on the maximum angular weight \(\alpha_d^{\mathrm{max}}\) for the Office-Home dataset. We report overall accuracy (\emph{All}), seen-class accuracy (\emph{Old}), and novel-class accuracy (\emph{New}).}
  \label{tab:alpha_d_max_ablation}
  \begin{tabular}{c|ccc}
    \toprule
    \(\alpha_d^{\mathrm{max}}\) & All (\%) & Old (\%) & New (\%) \\
    \midrule
    0.25  & 56.43 & 58.10 & 53.95 \\
    0.50  & 55.23 & 56.31 & 53.61 \\
    0.75  & 56.05 & 58.81 & 52.02 \\
    1.00  & \textbf{56.78} & \textbf{59.23} & \textbf{53.21} \\
    \bottomrule
  \end{tabular}
\end{table}

\subsection{Observation on Old-New class splits:}
Table \ref{tab:splits} illustrates the impact of varying base (old) and novel (new) class splits on the performance of {\ourmodel} on the Office-Home dataset.

\begin{table}[!htbp]
    \centering
    \begin{tabular}{cccc}
        \toprule
        \multirow{2}{*}{\textbf{Splits (Old-New)}} & \multicolumn{3}{c}{\textbf{Office-Home}} \\
        \cline{2-4}
        & \textbf{All} & \textbf{Old} & \textbf{New} \\
        \midrule
        \textbf{40 - 25} & 56.78 & 59.23 & 53.21 \\
        \textbf{30 - 35} & 55.31 & 58.87 & 52.47 \\
        \textbf{25 - 40} & 54.76 & 57.95 & 52.84 \\
        \textbf{55 - 10} & 57.85 & 59.63 & 48.46 \\
        \textbf{50 - 15} & 57.13 & 58.23 & 53.59 \\
        \bottomrule
    \end{tabular}
    \caption{Sensitivity on different \texttt{Old}-\texttt{New} class splits.}
    \label{tab:splits}
\end{table}

In Table \ref{tab:splits}, we observe that increasing the number of known classes (50–15 split vs. 40–25 split) leads to a small drop in Old-class accuracy ($\approx$ 1\%). This can be explained by two interacting effects:

\begin{itemize}[leftmargin=*,noitemsep]
  \item \textbf{Boundary crowding.} Adding more \textbf{Old} classes places additional prototypes closer to the Poincaré boundary, which increases overlap and slightly raises confusion among old categories. Interestingly, this higher density benefits discovery of \textbf{New} classes by pushing unfamiliar samples toward the interior, yielding the observed tradeoff (Old $\downarrow$, New $\uparrow$).
  
  \item \textbf{Per-class sample dilution.} With a fixed training budget, more classes imply fewer samples per class. This increases per-class variance and mildly reduces Old-class accuracy.
\end{itemize}

Overall, the effect is minor and consistent with our geometric intuition: higher boundary density aids novel-class separation at the cost of a small accuracy reduction on known classes.





\clearpage
\section{Theoretical Justification: Generalization Bounds in Hyperbolic Space}
\label{sup:theory_generalization}

\subsection{Formal Justification: Why \texorpdfstring{$\Delta_{\{\mathbb{H}\}}$, $\Delta_{\{\mathbb{H}\}}(\mathbb{E})\text{s}$, ($\Delta_{\mathbb{H}} < \Delta_{\mathbb{E}}$) Can Hold}{Formal Justification: Why Delta H, Delta H(E)s, (Delta H < Delta E) Can Hold}}

Let \( \mathcal{D}_S' \) denote the augmented source domain and \( \mathcal{D}_T \) the target domain. For a hypothesis class \( \mathcal{H}_{\mathcal{G}} \) defined over geometry \( \mathcal{G} \in \{\mathbb{E}, \mathbb{H}\} \), the domain discrepancy is:

\[
\Delta_{\mathcal{G}}(\mathcal{D}_S', \mathcal{D}_T) = \sup_{f \in \mathcal{H}_{\mathcal{G}}} \left| \mathbb{E}_{x \sim \mathcal{D}_S'} [f(x)] - \mathbb{E}_{x \sim \mathcal{D}_T} [f(x)] \right|.
\]

Assume:
\begin{itemize}
    \item[(A1)] The latent structure of \( \mathcal{D}_S' \cup \mathcal{D}_T \) is approximately hierarchical (e.g., tree-like);
    \item[(A2)] The encoders \( \phi_{\mathcal{E}} \), \( \phi_{\mathcal{H}} \) are 1-Lipschitz under Euclidean and hyperbolic metrics respectively;
    \item[(A3)] \( \mathcal{H}_{\mathcal{G}} \) is 1-Lipschitz with respect to \( d_{\mathcal{G}} \).
\end{itemize}

Using Kantorovich–Rubinstein duality~\cite{villani2009optimal}, we upper-bound the discrepancy via the Wasserstein-1 distance:

\[
\Delta_{\mathcal{G}}(\mathcal{D}_S', \mathcal{D}_T) \leq W_1^{\mathcal{G}}(\phi_{\mathcal{G}\#} \mathcal{D}_S', \phi_{\mathcal{G}\#} \mathcal{D}_T),
\]
where \( \phi_{\mathcal{G}\#} \mathcal{D} \) is the pushforward distribution in \( \mathcal{M}_{\mathcal{G}} \).

\paragraph{Why Hyperbolic Distance Tightens Discrepancy.}
In tree-like or hierarchical spaces, embeddings into hyperbolic geometry suffer significantly less distortion than Euclidean ones. Formally, Bourgain’s theorem and subsequent results in geometric embedding theory show that the distortion for embedding \( n \)-node trees into:
\begin{itemize}
    \item Euclidean space is \( \Omega(\log n) \)~\cite{linial1995geometry};
    \item Hyperbolic space is \( \mathcal{O}(1) \)~\cite{verbeek2011metric}.
\end{itemize}

Thus, for semantically structured data (as in class hierarchies or taxonomies), the Wasserstein-1 distance between distributions is tighter in hyperbolic space:
\[
W_1^{\mathbb{H}}(\phi_{\mathbb{H}\#} \mathcal{D}_S', \phi_{\mathbb{H}\#} \mathcal{D}_T) \leq W_1^{\mathbb{E}}(\phi_{\mathbb{E}\#} \mathcal{D}_S', \phi_{\mathbb{E}\#} \mathcal{D}_T) + \mathcal{O}(\log n),
\]
where \( n \) is the number of semantic entities.

\paragraph{Implication for DG-GCD.} Since domain discrepancy \( \Delta_{\mathbb{G}} \) directly impacts generalization bounds (cf. \cite{mohri2018foundations, chami2019hyperbolic}), this structural preservation implies:
\[
\Delta_{\mathbb{H}}(\mathcal{D}_S', \mathcal{D}_T) < \Delta_{\mathbb{E}}(\mathcal{D}_S', \mathcal{D}_T),
\]
leading to a tighter bound in the hyperbolic setting—provided semantic hierarchy and low distortion hold.

 Hyperbolic geometry more naturally encodes semantic divergence with exponential volume growth, allowing fewer augmentations to approximate complex target distributions. This justifies why the same loss functions, when deployed in \( \mathbb{D}_c^d \), yield a tighter generalization guarantee compared to their Euclidean counterparts.

\subsection*{On the Geometric Validity of Tangent CutMix}

We provide a mathematical justification that Tangent CutMix preserves hyperbolic consistency under the Poincaré ball model. Let $\mathbb{D}^d_c = \{ z \in \mathbb{R}^d : \|z\| < 1/\sqrt{c} \}$ denote the $d$-dimensional Poincaré ball with curvature $-c$. Given two embeddings $z_i, z_j \in \mathbb{D}^d_c$, we map them to the tangent space at the origin via the logarithmic map:
\[
v_i = \log_0^c(z_i) = \frac{2}{\lambda_{z_i}} \tanh^{-1}(\sqrt{c}\|z_i\|) \cdot \frac{z_i}{\|z_i\|}, \quad \lambda_{z_i} = 1 - c\|z_i\|^2,
\]
yielding $v_i, v_j \in \mathbb{R}^d$. Tangent CutMix then interpolates in this Euclidean tangent space:
\[
v_{\text{mix}} = \lambda v_i + (1 - \lambda) v_j, \quad \lambda \sim \text{Uniform}(0,1),
\]
and projects back to the manifold using the exponential map:
\[
z_{\text{mix}} = \exp_0^c(v_{\text{mix}}) = \tanh(\sqrt{c} \|v_{\text{mix}}\|) \cdot \frac{v_{\text{mix}}}{\sqrt{c} \|v_{\text{mix}}\|}.
\]
Since $\tanh(x) < 1$ for all $x>0$, we have $\|z_{\text{mix}}\| < 1/\sqrt{c}$, ensuring $z_{\text{mix}} \in \mathbb{D}^d_c$. Thus, Tangent CutMix guarantees valid hyperbolic embeddings.

To further ensure stability, we employ the penalized Busemann loss (Eq.~4), whose $\log(1 - \|z\|^2)$ term discourages embeddings from approaching the boundary. Empirically, no violations of geometric constraints were observed. Replacing Tangent CutMix with Euclidean CutMix caused a significant performance drop ($-3.96\%$ on Office-Home; Table 4), underscoring the necessity of hyperbolic consistency for robust generalization.

\subsection*{On the Role of Hyperbolic Capacity in Data Augmentation}

Hyperbolic geometry provides a natural explanation for why fewer augmentations are required compared to Euclidean spaces. In a $d$-dimensional Poincaré ball with curvature $-c$, the volume of a ball grows exponentially with radius, i.e.,
\[
V_{\text{Hyp}}(r) \sim \exp(\sqrt{c}\,r),
\]
in contrast to the polynomial growth $V_{\text{Euc}}(r) \sim r^d$ in Euclidean space. This exponential expansion increases representational capacity, enabling semantic clusters to be separated with less overlap even when data coverage is sparse. Consequently, a small number of augmentations suffices to populate the embedding space without significant distortion.

Empirically, we find that {\ourmodel} achieves 56.07\% accuracy on Office-Home even without any synthetic domains, outperforming several Euclidean and hyperbolic baselines that rely on 6–9 augmentations. Introducing only 1–2 GPT-guided augmentations improves performance further (56.78\%), whereas excessive augmentation degrades novel-class accuracy due to overfitting (Fig.~3, right). 

Finally, our generalization analysis (Sec.~3.4, Eq.~9) shows that the hyperbolic discrepancy term $\Delta_{\mathbb{H}}(S',T)$ yields a tighter Rademacher-based bound than its Euclidean counterpart $\Delta_{\mathbb{E}}(S',T)$ under the same augmentation budget. Together, these theoretical and empirical findings support the claim that hyperbolic geometry’s exponential capacity reduces the reliance on heavy augmentation for domain generalization.

\subsection{Euclidean vs.\ Hyperbolic Loss Ablation}
\label{sec:supp-euc-hyp-ablation}


Below we list the direct Euclidean analogues of our three hyperbolic loss components, obtained by replacing hyperbolic distances in the Poincaré ball with standard $\ell_{2}$ norms in $\mathbb{R}^d$.

\paragraph{1. Euclidean Prototype‐Softmax (Analogue of Busemann Loss)}  
\[
\mathcal{L}_{\mathrm{Euc\text{-}Buse}}(z, y)
= -\log
  \frac{\exp\bigl(-\|z - \mu_{y}\|_{2}^{2}\bigr)}
       {\sum_{j=1}^{C}\exp\bigl(-\|z - \mu_{j}\|_{2}^{2}\bigr)} 
  \quad,\quad
\mu_{j}\;=\;\frac{1}{N_{j}}\sum_{i:y_{i}=j}z_{i}
\]
where $\{\mu_{j}\}$ are the class prototypes in $\mathbb{R}^{d}$.

\paragraph{2. Euclidean Contrastive Loss (Analogue of Hyperbolic Contrastive Term)}  
\[
\mathcal{L}_{\mathrm{Euc\text{-}Con}}
= -\frac{1}{|B|}
  \sum_{i\in B}
  \log
  \frac{\exp\!\bigl(\langle\hat z_{i},\hat z_{i}^{+}\rangle/\tau\bigr)}
       {\displaystyle\sum_{a\in B\setminus\{i\}}
             \exp\!\bigl(\langle\hat z_{i},\hat z_{a}\rangle/\tau\bigr)}
\]
where $\hat z=\tfrac{z}{\|z\|_{2}}$ denotes $\ell_{2}$ normalization, $B$ is the batch of views, and $\tau$ is the temperature.

\paragraph{3. Euclidean Outlier Repulsion (Analogue of Adaptive Outlier Loss)}  
\[
\mathcal{L}_{\mathrm{Euc\text{-}Out}}
= \frac{1}{|B|}
  \sum_{i\in B}
  \max\Bigl(0,\;\gamma
                 - \min_{k}\|\,z_{i}^{\mathrm{mix}} - \mu_{k}\|_{2}\Bigr)
\]
which pushes each synthetic mix‐up embedding $z_{i}^{\mathrm{mix}}$ at least a margin $\gamma$ away from all class prototypes in Euclidean space.

 Table~\ref{tab:supp-euc-hyp} shows that replacing hyperbolic distance with Euclidean norm widens the generalization gap: although both variants improve over a vanilla ViT baseline, only the hyperbolic formulation consistently boosts novel‐class discovery (New) while raising overall (All) and old‐class (Old) accuracy.

\begin{table}[!ht]
  \centering
  \caption{Ablation of prototype‐softmax loss in Euclidean vs.\ Hyperbolic space on Office-home (2 synthetic domains).}
  \label{tab:supp-euc-hyp}
  \scriptsize
  \setlength{\tabcolsep}{12pt}
  \renewcommand{\arraystretch}{1.1}
  \begin{tabular}{lccc}
    \toprule
    \textbf{Manifold}  & \textbf{All (\%)} & \textbf{Old (\%)} & \textbf{New (\%)} \\
    \midrule
    Euclidean Manifold 
      & 49.46 & 52.08 & 45.66 \\
    Hyperbolic Manifold
      & \textbf{56.78} & \textbf{59.23} & \textbf{53.21} \\
    \bottomrule
  \end{tabular}
\end{table}



\section{Dataset details} \label{sec:dataset_details}
Our experiments were carried out using three benchmark datasets: (i) \textbf{PACS} \cite{pacs}, (ii) \textbf{Office-Home} \cite{officehome}, and (iii) \textbf{DomainNet} \cite{domainnet}.

For both the PACS and Office-Home datasets, we used each domain as the source, while treating all other domains as target domains. In the case of DomainNet, we selected a subset of source-target pairs, as outlined in Table \ref{tab:domainnet_config}, to ensure the model was evaluated across a variety of domain combinations.

\begin{table}[!h!t!p]
\centering
\footnotesize
\resizebox{0.75\columnwidth}{!}{%
\begin{tabular}{@{}c>{\centering\arraybackslash}p{8cm}@{}}
\toprule
\textbf{Source} & \textbf{Targets} \\ 
\midrule
Sketch & Clipart, Painting, Infograph, Quickdraw, Real World \\ 
Painting & Clipart, Sketch, Infograph, Quickdraw, Real World \\ 
Clipart & Painting, Sketch, Infograph, Quickdraw, Real World \\ 
\bottomrule
\end{tabular}}
\caption{Source-target configurations for DomainNet}
\label{tab:domainnet_config}
\end{table}

\section{Further Comparison with Literature}

\subsection{DG-GCD-Specific Adaptations for Baselines}
As described in \cite{dg2net}, we adapted several state-of-the-art methods for generalized category discovery (GCD) and domain generalization (DG) to the DG-GCD framework in a similar manner. This framework completely excludes access to target domain data during training, and for some methods, synthetic domain data is introduced to simulate domain shifts.

For the ViT-B/16 model pre-trained with DINO, we fine-tuned only the final block using source domain data, adhering to standard GCD protocols. We then evaluated the model on target domains without incorporating any target domain data during training. Similarly, for GCD, we fine-tuned the last block of the backbone using only source domain data and introduced a synthetic domain variant to better account for domain shifts.

We applied the same adaptation procedure to CMS (Contrastive Mean Shift) and SimGCD. For these methods, we fine-tuned the last block using only source domain data and created synthetic variants by incorporating synthetic domain data to evaluate their performance in handling domain shifts effectively.

For CDAD-Net, initially designed for cross-domain adaptation, we adapted it for the DG-GCD setting by training solely on source domain data, excluding target domain information. Additionally, we created synthetic variants to assess its performance on unseen domains.

For Hyp-GCD and Hyp-SelfEx, we followed the procedure outlined in \cite{Liu2025HypCD}. As for the Hyp-{\prevmodel}, we adapted the code directly from the authors, modifying the two loss functions in hyperbolic space following the strategy described in \cite{Liu2025HypCD}.

\subsection{Implementation Details.}
\label{sec:implementation}
We build {\ourmodel} in PyTorch and run all experiments on a single NVIDIA A100-SXM4-80GB GPU. Following~\cite{gcd,dg2net}, we adopt a DINO-pretrained ViT-B/16~\cite{vit,dino}, fine-tuning only its final transformer block and using the [CLS] token as our image representation.These features pass through a 3-layer MLP projection head \( g_\phi \) to yield 32-dimensional embeddings.
To simulate domain shift, we generate class-preserving synthetic domains using diffusion models guided by GPT-4o-curated prompts. These are merged with the labeled source domain \( D_S \) into a unified training set. We rank candidate domains using a composite FID-based score that balances source divergence and inter-domain diversity (Eq.1 of main paper).

During training, features are mapped into the Poincaré ball \( \mathbb{D}_c^d \) using the exponential map, with a learnable curvature parameter \( c \), initialized to 0.05. We clip embeddings to ensure numerical stability and normalize them using \(\tanh\). Class prototypes are initialized randomly on the boundary and refined via a 1000-step SGD loop (learning rate 0.1, momentum 0.9). We employ a penalized Busemann loss for aligning features to fixed ideal prototypes and a hybrid hyperbolic contrastive loss~\cite{Liu2025HypCD} for instance-level alignment. The hybrid loss combines geodesic distance and cosine similarity in tangent space, with a weighting factor \( \alpha_d \) that varies linearly over epochs.

Training is performed for 50 epochs using SGD (learning rate 0.01, momentum 0.9, weight decay \(5 \times 10^{-5}\)), with cosine annealing. The curvature \( c \), projection head, final ViT block, and prototype embeddings are optimized jointly. For Office-Home, PACS, and Domain-Net, we set the Poincaré radius to 1.5, 1.0, and 2.3 respectively. We follow \cite{dg2net} for known/novel splits and use the K-estimation method from the same work to estimate the number of clusters \(K\) via Brent’s algorithm, constrained to \([|\mathcal{Y}_s|, 1000]\).


\noindent \textbf{Training Objective and Loss Function Combination:} 
The final training objective is a weighted sum of three key loss components: the \textit{Penalized Busemann Loss}, the \textit{Hybrid contrastive Loss}, and the \textit{Outlier Loss}. The total loss is formulated as follows.

\begin{equation}
\mathcal{L}_{\text{total}} = 
\lambda_1 \underbrace{\mathcal{L}_{\text{Buse}}}_{\text{Semantic alignment}} + 
\lambda_2 \underbrace{\mathcal{L}_{u}}_{\text{Contrastive regularization}} + 
\lambda_3 \underbrace{\mathcal{L}_{\text{out}}}_{\text{Outlier repulsion}}, \quad \text{where } \lambda_1 + \lambda_2 + \lambda_3 = 1.
\end{equation}

In our implementation, we set $\lambda_1 = 0.60$, $\lambda_2 = 0.25$, and $\lambda_3 = 0.15$, which ensures a balanced contribution from each term.  These hyperparameters are influenced by prior works~\cite{gcd,dg2net}. The \textit{Penalized Busemann Loss} aligns the feature embeddings with ideal class prototypes in hyperbolic space, while the \textit{Hybrid contrastive Loss} encourages discriminative features between positive and negative pairs across augmented views. The \textit{Outlier Loss} with the adaptive margin penalizes embeddings that deviate significantly from the prototypes, improving generalization by limiting the influence of outliers. This combined loss function guides the model to learn robust, domain-invariant feature representations for both seen and unseen classes during training.


\begin{algorithm}[!ht]
\caption{Training Procedure for DG-GCD with {\ourmodel}}
\label{alg:training}

\textbf{Input:} 
\begin{itemize}[leftmargin=*]
    \item Labeled source dataset $\mathcal{D}_S = \{(x_i^s, y_i^s)\}_{i=1}^{n_s}$ with $N$ known classes
    \item Synthetic domain augmentations $\{\mathcal{D}_{\mathrm{syn}}^{(k)}\}_{k=1}^M$ to simulate domain shift
    \item Pretrained ViT backbone $F_\theta$ (frozen except last transformer block)
    \item Projection head $G_\phi$ mapping features to hyperbolic embeddings
    \item Fixed ideal prototypes $\{\mathbf{p}_j\}_{j=1}^N$ placed on the boundary $\partial \mathbb{D}_c^d$ of Poincaré ball
    \item Learnable curvature parameter $c$ controlling hyperbolic geometry
    \item Hyperparameters: batch size $B$, epochs $E$, learning rates, penalization coefficient $\phi$, number of views $V$, and loss weights $\lambda_1, \lambda_2, \lambda_3$ with $\lambda_1 + \lambda_2 + \lambda_3 = 1$
\end{itemize}

\textbf{Output:} Optimized parameters $\theta, \phi$, curvature $c$, and fixed prototypes $\{\mathbf{p}_j\}$

\begin{algorithmic}[1]

\For{$e = 1$ \textbf{to} $E$} \Comment{Epoch loop}

    \State Set $F_\theta$, $G_\phi$ to training mode

    \For{each batch $(X, Y)$ sampled from combined datasets $\mathcal{D}_S \cup \bigcup_k \mathcal{D}_{\mathrm{syn}}^{(k)}$} \Comment{Batch loop}

        \State Generate $V$ augmented views per input in $X$

        \State Extract features: $Z^\mathbb{E} = F_\theta(X)$

        \State Project embeddings: $Z = G_\phi(Z^\mathbb{E})$

        \State Clip embeddings to radius $r$ (dependent on dataset and curvature $c$)

        \State Generate mixed embeddings $Z_{\mathrm{mix}}$ by convex combination in tangent space at origin:
        \[
            Z_{\mathrm{mix}} = \exp_0^c\big( \lambda \log_0^c(Z_i) + (1-\lambda) \log_0^c(Z_j) \big), \quad \lambda \sim \mathrm{Uniform}(0,1)
        \]

        \State Apply element-wise $\tanh$ to ensure embeddings lie strictly within Poincaré ball $\mathbb{D}_c^d$

        \State Construct batch prototypes $P_Y = \{\mathbf{p}_{y_i}\}_{i=1}^{B}$ repeated $V$ times corresponding to labels $Y$

        \State Compute \textbf{penalized Busemann loss}~\cite{ghadimi2021hyperbolic}:
        \[
            \mathcal{L}_{\mathrm{Buse}} = \frac{1}{BV} \sum_{i=1}^{BV} \left[ \log \left( \frac{\|z_i - \mathbf{p}_{y_i}\|^2}{1 - c \|z_i\|^2} \right) + \phi \log (1 - \|z_i\|^2) \right]
        \]

        \State Compute \textbf{hybrid hyperbolic contrastive loss}~\cite{Liu2025HypCD} on $V$ views:
        \[
            \mathcal{L}_u = \mathrm{HybridContrastiveLoss}(Z, V, \alpha(e))
        \]
        \Comment{$\alpha(e)$ is epoch-dependent weighting between geodesic and angular similarity}

        \If{$e=1$ and first batch}
            \State Compute adaptive margin $\delta$ as quantile over distances $d_{\mathbb{H}}(Z_{\mathrm{mix}}, \mathbf{p}_j)$ to prototypes
        \EndIf

        \State Compute \textbf{adaptive outlier loss}:
        \[
            \mathcal{L}_{\mathrm{out}} = \frac{1}{|Z_{\mathrm{mix}}|} \sum_{z \in Z_{\mathrm{mix}}} \mathrm{ReLU}\big( \delta - \min_j d_{\mathbb{H}}(z, \mathbf{p}_j) \big)
        \]

        \State Aggregate total loss:
        \[
            \mathcal{L} = \lambda_1 \mathcal{L}_{\mathrm{Buse}} + \lambda_2 \mathcal{L}_u + \lambda_3 \mathcal{L}_{\mathrm{out}}
        \]

        \State Backpropagate $\mathcal{L}$ and update parameters $\theta, \phi, c$ via optimizer

    \EndFor

    \State Update learning rate scheduler

\EndFor

\end{algorithmic}
\end{algorithm}


\subsection{Effect of Reduced Synthetic Domains on Model Performance}

\begin{table}[!ht]
\centering
\caption{Performance comparison on the PACS dataset between {\ourmodel} and {\prevmodel} with 2 and 6 synthetic domains.}
\label{tab:synthetic_domain_comparison}
\begin{tabular}{lccc}
    \toprule
    \textbf{Model Variant} & \textbf{All} & \textbf{Old} & \textbf{New} \\
    \midrule
    {\ourmodel} (2 Synthetic Domains) & 75.07 & 75.54 & 74.52 \\
    {\prevmodel} (2 Synthetic Domains) & 66.86 & 69.11 & 63.75 \\
    {\ourmodel} (6 Synthetic Domains) & 74.00 & 75.64 & 71.95 \\
    {\prevmodel} (6 Synthetic Domains) & 73.30 & 75.28 & 72.56 \\
    \bottomrule
\end{tabular}
\end{table}

Table~\ref{tab:synthetic_domain_comparison} presents a detailed comparison between {\ourmodel} and {\prevmodel} on the PACS dataset when using 2 versus 6 synthetic domains during training. While {\ourmodel} maintains robust performance with fewer synthetic domains, {\prevmodel} experiences a marked degradation, confirming the claim in the main paper regarding {\prevmodel}'s sensitivity to synthetic domain count.

This underscores the efficiency and robustness of our lightweight synthetic domain augmentation strategy, which achieves strong generalization with fewer synthetic domain samples.

\subsection*{Why Naïve Hyperbolic Adaptations for DG$^2$CD-Net Fail}

To better understand the design choices in {\ourmodel}, we also implemented a hyperbolic variant of {\prevmodel} (denoted Hyp-{\prevmodel}). Interestingly, Hyp-{\prevmodel} underperforms both the original {\prevmodel} and our proposed {\ourmodel}. We attribute this gap to several factors:

\begin{itemize}[leftmargin=*,noitemsep]
    \item \textbf{Task Vector Arithmetic Mismatch.} {\prevmodel} aggregates task vectors via Euclidean averaging during episodic training. In hyperbolic space, vector addition is not associative or commutative, so retaining Euclidean arithmetic introduces geometric distortion.
    
    \item \textbf{Episodic Prototype Drift.} Injecting synthetic novel classes episodically without curvature-aware constraints causes embeddings to drift toward the boundary, reducing separability and increasing overconfidence.
    
    \item \textbf{Loss Function Misalignment.} Hyp-{\prevmodel} reuses Euclidean-designed losses, which are not well-defined on manifolds. In contrast, {\ourmodel} employs Busemann-aligned supervision, geodesic-aware contrastive regularization, and adaptive outlier repulsion to maintain consistency with curved geometry.
    
    \item \textbf{Empirical Results.} As shown in Table~1, Hyp-{\prevmodel} performs consistently worse than both {\prevmodel} and {\ourmodel} across all benchmarks, underscoring the importance of end-to-end geometry-aware design rather than simple embedding substitution.
\end{itemize}

These findings emphasize that effective hyperbolic adaptation requires full-stack integration of geometry-aware components—embedding, loss design, and training protocol—rather than replacing only the representation space.

\subsection{Learned Curvature Evolution}
\label{sup:curvature_evolution}

\begin{figure}[ht]
    \centering
    \includegraphics[width=\linewidth]{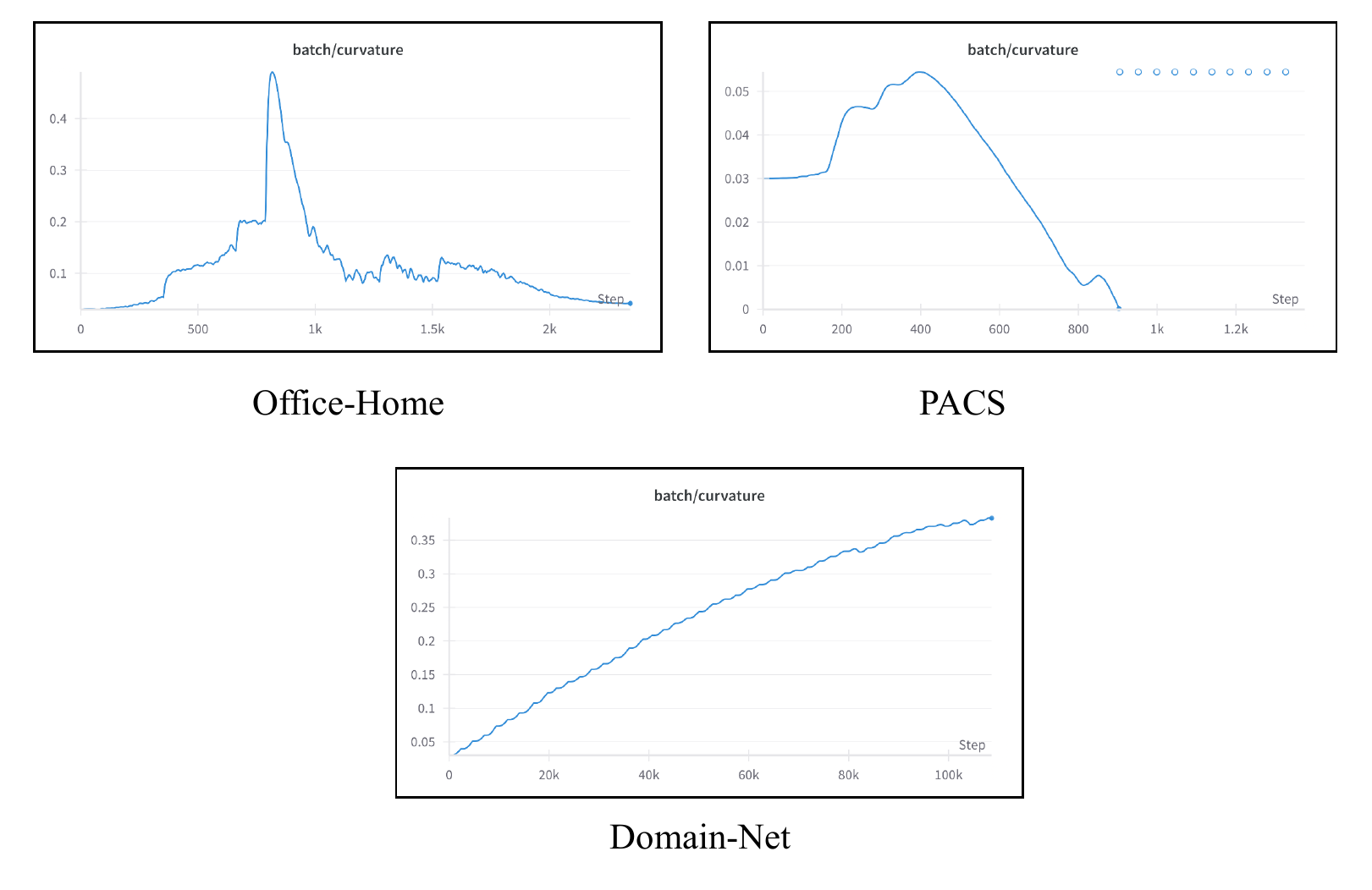}
    \caption{Learned curvature \(c\) evolution across training batches for Office-Home (left), PACS (center), and DomainNet (right).These plots were generated from training logs visualized using Weights \& Biases (WandB)~\cite{wandb}.}
    \label{fig:curvature_evolution}
\end{figure}

Our model learns the hyperbolic curvature \(c\) jointly during training, adapting it to dataset-specific geometry. Figure~\ref{fig:curvature_evolution} illustrates the curvature evolution for three benchmark datasets: Office-Home, PACS, and DomainNet.
These learned curvatures are consistent with the intuition that more complex or diverse datasets like DomainNet require a higher curvature, while relatively simpler domains such as Office-Home stabilize at lower curvature values.

\subsection{Single‐Source Domain Generalization}
\label{sec:ssdg_combined}

To isolate the effect of embedding geometry, we train two shallow MLPs from scratch—\emph{Euclidean‐Net} and \emph{Hyperbolic‐Net}—on a single source domain, and evaluate them on the three held‐out domains of both the PACS and Office Home benchmarks.  Both models share identical capacity (1.12 M trainable parameters), a unified training schedule (100 epochs, batch size 128, learning rate $1\times10^{-3}$), and are optimized using the cross‐entropy loss.

\begin{table}[!htbp]
  \centering
  \caption{Average test accuracy (\%) when training on one source domain and testing on three held‐out domains.}
  \label{tab:ssdg_combined}
  \begin{tabular}{lccc}
    \toprule
    \textbf{Dataset}       & \textbf{EuclideanNet} & \textbf{HyperbolicNet} & \textbf{Relative Gain} \\
    \midrule
    PACS                   & 12.9                  & 22.3                   & +72.86\%                  \\
    Office Home            &  1.96                 &  6.17                  & +214.79\%                \\
    Domain Net             &  0.28                 & 0.96                   & + 242.85 \%
    \\
    \bottomrule
  \end{tabular}
\end{table}

\noindent\textbf{Discussion.}  
Across both benchmarks, Hyperbolic-Net substantially outperforms Euclidean-Net under the single‐source regime. On PACS, hyperbolic embeddings yield a 73 \% relative improvement (22.3 \% vs.\ 12.9 \%), while on Office Home they achieve a 215 \% gain (6.17 \% vs.\ 1.96 \%). These consistent gains demonstrate that hyperbolic geometry inherently captures the relational structure necessary for generalization when only one domain is available. 

\subsection{Embedding Dimension}
\label{sec:ablation_dimension}
\begin{figure}[!ht]
    \centering
\begin{tikzpicture}
\begin{axis}[
    ybar,
    bar width=15pt,
    ylabel=Accuracy (\%),
    ymin=0, ymax=70,
    enlarge x limits=0.15,
    symbolic x coords={32,64,128,256,512},
    xtick=data,
    xtick style={draw=none},
    xlabel=Dimension,
    axis x line*=bottom,
    axis y line*=left,
    ymajorgrids=true,
    grid style=dashed,
    legend style={
        draw=black,
        fill=white,
        at={(1.05,0.5)},
        anchor=west,
        legend columns=1,
        font=\small,
    },
    nodes near coords,
    every node near coord/.append style={font=\tiny,anchor=south},
    width=12cm,
    height=5cm,
    cycle list name=color list,
]
\legend{All,Old,New}
\addplot+[draw=black, fill=blue!50] 
    coordinates {(32,56.78) (64,57.87) (128,57.57) (256,45.09) (512,39.01)};

\addplot+[draw=black, fill=red!50] 
    coordinates {(32,59.23) (64,64) (128,62.75) (256,45.84) (512,39.9)};

\addplot+[draw=black, fill=green!50] 
    coordinates {(32,53.21) (64,47.46) (128,48.76) (256,43.83) (512,37.5)};

\end{axis}
\end{tikzpicture}
\caption{Ablation study on embedding dimension on the Office-Home dataset. Lower dimensions (32, 64, 128) perform better than larger ones (256, 512), suggesting that compact embeddings are beneficial for generalization.}
\label{fig:dim_ablation}
\end{figure}
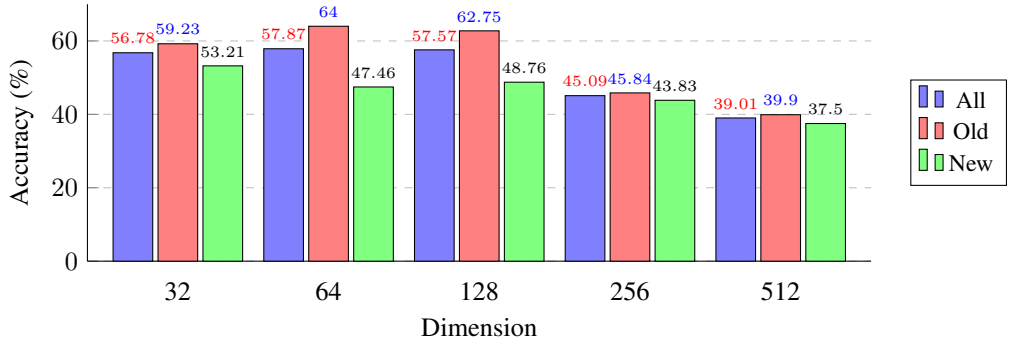

In order to investigate the effect of embedding dimensionality on model performance, we perform experiments on the Office-Home dataset by varying the embedding dimension between $\{32, 64, 128, 256, 512\}$. Figure~\ref{fig:dim_ablation} summarizes the results in three evaluation settings: overall accuracy (\textit{All}), accuracy on seen classes (\textit{Old}), and accuracy on novel classes (\textit{New}).

We observe that lower-dimensional embeddings (\textit{32, 64, 128}) consistently outperform larger dimensions (\textit{256, 512}) across all metrics. In particular, a dimension of $64$ achieves the highest precision in the seen class ($64.00\%$), while a dimension of $32$ provides the best performance in novel classes ($53.23\%$). Increasing the dimension beyond $128$ leads to a sharp degradation in performance, indicating that excessively large embeddings may overfit the seen classes while impairing generalization to novel categories. Based on these findings, we select a dimension of $32$ as the default setting for subsequent experiments, achieving a favorable balance between the performance seen and the novel class performance.

\subsection{For Non-learnable curvature}

\begin{table}[!ht]
    \centering
    \caption{Impact of hyperbolic curvature $c$ on performance.}
    \vspace{0.5em}
    \begin{tabular}{c|c|c|c}
        \toprule
        \textbf{Curvature $c$} & \textbf{All (\%)} & \textbf{Old (\%)} & \textbf{New (\%)} \\
        \midrule
        0.01 & 56.23 & 58.65 & 52.68 \\
        0.03  & 55.67 & 57.39 & 52.69 \\
        0.05 & 56.01 & 58.19 & 52.78 \\
        0.10 & 56.27 & 58.52 & 52.97 \\
        \bottomrule
    \end{tabular}
    \label{tab:curvature_ablation}
\end{table}

\subsection{Prototype count analysis}  
We ablated the number of prototypes per class (\(P \in \{1, 2, 4\}\)) on Office-Home. Results are summarized below:

\vspace{0.5em}
\begin{table}[!ht]
\centering
\begin{tabular}{lccc}
\toprule
\textbf{Prototypes / Class} & \textbf{All} & \textbf{Old} & \textbf{New} \\
\midrule
1 & 56.78 & 59.23 & 53.21 \\
2 & 54.40 & 57.85 & 48.68 \\
4 & 56.23 & 59.75 & 50.27 \\
\bottomrule
\end{tabular}
\caption{Prototype count analysis}
\label{tab:prototypes_ablation}
\end{table}
\vspace{0.5em}
    
We find that performance peaks at \(P = 1\), suggesting that one discriminative prototype per class is sufficient in the hyperbolic space. Larger values of \(P\) did not improve generalization and sometimes hurt it, possibly due to overfitting or prototype redundancy.

\subsection*{Ablation on Overconfidence Regularizers}

To complement the main results, we evaluate alternative strategies for mitigating overconfident embeddings in hyperbolic space. Specifically, we compare our proposed \textbf{Penalized Busemann Loss} against entropy regularization, confidence penalty (KL to uniform), and logit margin penalty. Each variant replaces only the penalty term in the Busemann component while keeping all other settings identical (backbone, optimizer, loss weights, etc.), ensuring a controlled one-factor analysis.

\begin{table}[!ht]
    \centering
    \caption{\small Ablation of overconfidence regularizers on Office-Home. Our Penalized Busemann Loss achieves the best overall trade-off between known and novel class performance.}
    \label{tab:supp_reg_ablation}
    \vspace{0.5em}
    \begin{tabular}{lccc}
    \toprule
    \textbf{Regularizer} & \textbf{All} & \textbf{Old} & \textbf{New} \\
    \midrule
    Entropy & 51.63 & 57.31 & 41.86 \\
    Confidence Penalty & 56.69 & \textbf{61.15} & 49.10 \\
    Logit Margin & 54.49 & 56.54 & 51.02 \\
    \textbf{Penalized Busemann (Ours)} & \textbf{56.78} & 59.23 & \textbf{53.21} \\
    \bottomrule
    \end{tabular}
\end{table}

Overall, confidence-based penalties yield stronger performance on ``Old'' categories but degrade novel discovery, while entropy maximization is least effective under domain shift. Our Penalized Busemann Loss achieves the most balanced results, underscoring the importance of geometry-aware regularization for stable and generalizable discovery in curved spaces.

\subsection*{Geometric Motivation and Theoretical Properties}

\noindent\textbf{1. Busemann‐based Penalty Aligns with Hyperbolic Geometry.}  
In the Poincaré ball, the \emph{Busemann function} centered at prototype $p$ is
\[
  B_p(z) \;=\;
  \log\!\frac{\|z - p\|^2}{1 - c\,\|z\|^2}\,,
\]
which measures signed distance to the horosphere through $p$.  Penalizing $B_p(z)$
\begin{itemize}[leftmargin=*,noitemsep]
  \item aligns directly with the true geodesic bisector between classes,
  \item adapts automatically to curvature $c$ via the $(1 - c\|z\|^2)$ term,
  \item yields \emph{uniform} radial repulsion—no directional bias in hyperbolic space.
\end{itemize}

\noindent\textbf{2. Theoretical Properties.}
\begin{itemize}[leftmargin=*,noitemsep]
  \item \emph{Isometry‐invariance:}  Horosphere distances are invariant under hyperbolic isometries, preserving model symmetry.
  \item \emph{Convexity in Busemann coordinates:}  Under the change of variables $u = B_p(z)$, the penalty is convex, improving optimization stability.
  \item \emph{Unified geometric form:}
    \[
      \underbrace{\log\!\frac{\|z - p\|^2}{1 - c\,\|z\|^2}}_{\text{Busemann}}
      \;+\;
      \underbrace{\phi\,\log\!(1 - \|z\|^2)}_{\text{radial repulsion}}
      \quad\longrightarrow\quad
      \text{single, curvature‐aware loss.}
    \]
\end{itemize}

Overall, unlike off‐the‐shelf entropy or KL penalties that operate in Euclidean logit‐space, our Penalized Busemann Loss is \emph{intrinsically hyperbolic}, principled by geometry, and thus yields superior control over overconfidence in the Poincaré embedding.



\subsection*{Choice of Hyperbolic Model: Poincaré vs. Lorentz}

We adopt the Poincaré ball model rather than the Lorentz model due to both theoretical and empirical considerations. First, our Tangent CutMix strategy (Sec. \ref{sec:tangent_cutmix}) relies on linear interpolation via exponential/logarithmic maps, which have closed-form and numerically stable expressions in the Poincaré model. In contrast, Lorentz formulations require projection onto the hyperboloid with normalization constraints, complicating implementation. Second, the conformal property of the Poincaré ball (angle-preserving) ensures compatibility with cosine similarity in the tangent space, which is essential for the angular term in our hybrid contrastive loss (Eq.~6). Third, while we experimented with a Lorentzian variant, we observed unstable gradients and weaker convergence when optimizing Busemann-aligned and contrastive losses. Finally, the bounded geometry of the Poincaré ball allows for intuitive visualization and interpretability (e.g., Fig.~4), which is useful for cluster analysis in DG-GCD.

For completeness, we compare the two models empirically on Office-Home under the DG-GCD setting:

\begin{table}[h!]
\centering
\begin{tabular}{lccc}
\hline
\textbf{Model} & \textbf{All} & \textbf{Old} & \textbf{New} \\
\hline
Poincaré Ball & \textbf{56.78} & \textbf{59.23} & \textbf{53.21} \\
Lorentz       & 54.28 & 56.01 & 51.41 \\
\hline
\end{tabular}
\caption{Comparison of Poincaré ball and Lorentz models in the DG-GCD setting on Office-Home.}
\label{tab:poincarevsLorentz}
\end{table}

The Poincaré ball consistently outperforms the Lorentz model across all splits, supporting our choice as the more effective and stable geometry for {\ourmodel}.

\subsection*{Varying the Number of Source and Target Domains}

To test robustness under different domain splits, we conducted an ablation where the number of source and target domains was systematically varied while keeping the total image budget fixed. For each configuration, the remaining domains were treated as target domains, and results were averaged across all possible target combinations.

\begin{table}[!htbp]
\centering
\begin{tabular}{lcccc}
\hline
\textbf{Training Domains} & \textbf{Method} & \textbf{All $\uparrow$} & \textbf{Old $\uparrow$} & \textbf{New $\uparrow$} \\
\hline
Art                     & {\prevmodel} & 54.47 & 53.65 & 55.54 \\
                        & {\ourmodel}   & 54.33 & 54.75 & 53.53 \\
\hline
Art + Clipart           & {\prevmodel} & 66.67 & 65.78 & 67.99 \\
                        & {\ourmodel}   & 67.34 & 70.78 & 62.02 \\
\hline
Art + Clipart + Product & {\prevmodel} & 66.83 & 66.89 & 66.75 \\
                        & {\ourmodel}   & \textbf{69.86} & \textbf{79.17} & 54.24 \\
\hline
\end{tabular}
\vspace{0.5em}
\caption{Ablation varying the number of source and target domains (Office-Home). {\ourmodel} consistently matches or exceeds {\prevmodel} across all scenarios.}
\label{tab:source_target_ablation}
\end{table}

Across all settings, {\ourmodel} maintains competitive or superior performance relative to {\prevmodel}. Notably, with three source domains, {\ourmodel} achieves the largest gain on Old classes, while still retaining strong performance on New categories. These results highlight that {\ourmodel} generalizes robustly even when the number of target domains decreases, consistent with its source-only, domain-agnostic training paradigm.

\subsection{Hyperparameter Sensitivity Analysis}

We appreciate the concern about hyperparameter complexity. In fact, we conducted a comprehensive ablation over all key knobs—curvature~$c$, slope~$\phi$, radius before mapping, contrastive weight~$\alpha_d$, and loss-term weights ($\lambda_1, \lambda_2, \lambda_3$) on the OfficeHome dataset. We found that only the penalized-Busemann slope and mapping radius materially affect results, with robust defaults ($\phi = 0.75$, radius = $1.5$) generalizing across benchmarks. For the remaining loss weights, setting $(\lambda_1, \lambda_2, \lambda_3) = (0.60, 0.25, 0.15)$ balances our alignment, contrastive, and outlier terms and yields overall accuracy $56.78\%$ ($59.23\%$ old, $53.21\%$ new). Swapping to $(0.15, 0.60, 0.25)$ or $(0.25, 0.15, 0.60)$ only reduces overall accuracy to $52.12\%$ and $51.37\%$, respectively—drops of ~4–5 points—demonstrating that our defaults achieve near-optimal performance without exhaustive tuning. Table~\ref{tab:hyperparam_ablation} shows these results.

\begin{table}[!htbp]
\centering
\caption{Ablation study on the loss term weights ($\lambda_1, \lambda_2, \lambda_3$) on the OfficeHome dataset. Our default configuration (Config 1) provides the best balance and overall accuracy.}
\label{tab:hyperparam_ablation}
\begin{tabular}{l ccc ccc}
\toprule
& \multicolumn{3}{c}{\textbf{Loss Weights}} & \multicolumn{3}{c}{\textbf{Accuracy (\%)}} \\
\cmidrule(lr){2-4} \cmidrule(lr){5-7}
\textbf{Configuration} & $\lambda_1$ & $\lambda_2$ & $\lambda_3$ & Avg. (All) & Avg. (Old) & Avg. (New) \\
\midrule
Config 1 (ours) & 0.60 & 0.25 & 0.15 & \textbf{56.78} & \textbf{59.23} & \textbf{53.21} \\
Config 2        & 0.15 & 0.60 & 0.25 & 52.12 & 53.33 & 50.07 \\
Config 3        & 0.25 & 0.15 & 0.60 & 51.37 & 52.17 & 50.01 \\
\bottomrule
\end{tabular}
\end{table}

\subsection{Comprehensive comparative analyses of {\ourmodel}  across multiple datasets} \label{sec:comparative_analysis}

This section presents a concise comparative analysis of {\ourmodel} on PACS, Office-Home, and DomainNet in Table-\ref{tab:PACS_comparison2}, \ref{tab:office_home_comparison2} and \ref{tab:domain_net_comparison3} respectively. Each dataset challenges {\ourmodel} with unique domain shifts, showcasing its adaptability and robustness. This evaluation aims to validate {\ourmodel}'s performance against established benchmarks, highlighting it's strengths and identifying opportunities for advancement in domain generalization.

\begin{table}[!htbp]
\centering
\resizebox{0.98\textwidth}{!}{%
\begin{tabular}{>{\bfseries}lccc|ccc|ccc|ccc|ccc|ccc}
\toprule
\multicolumn{19}{c}{\textbf{PACS}} \\ 
\midrule
 \multirow{2}{*}{\textbf{Methods}} & \multicolumn{3}{c|}{\textbf{Art Painting $\rightarrow$ Sketch}} & \multicolumn{3}{c|}{\textbf{Art Painting $\rightarrow$ Cartoon}} & \multicolumn{3}{c|}{\textbf{Art Painting $\rightarrow$ Photo}} & \multicolumn{3}{c|}{\textbf{Photo $\rightarrow$ Art Painting}} & \multicolumn{3}{c|}{\textbf{Photo $\rightarrow$ Cartoon}} & \multicolumn{3}{c}{\textbf{Photo $\rightarrow$ Sketch}} \\
 \cline{2-4} \cline{5-7} \cline{8-10} \cline{11-13} \cline{14-16} \cline{17-19}
 & \texttt{All} & \texttt{Old} & \texttt{New} & \texttt{All} & \texttt{Old} & \texttt{New} & \texttt{All} & \texttt{Old} & \texttt{New} & \texttt{All} & \texttt{Old} & \texttt{New} & \texttt{All} & \texttt{Old} & \texttt{New} & \texttt{All} & \texttt{Old} & \texttt{New} \\
\midrule
ViT \cite{vit}                  & 37.44                            & \textbf{50.73 }       & 19.5         & 47.4                              & 61.3         & 35.25        & 76.05                           & 87.13        & 64.64        & 53.17                           & 77.31        & 31.67        & 47.01                     & 55.54        & 39.57        & 31.87                    & 37.57        & 24.16        \\
GCD \cite{gcd}                  & 32.02                            & 41.53        & 19.12        & 46.78                             & 60.35        & 28.57        & 79.16                           & 99.45        & 48.73        & 74.73                           & 80.26        & 67.31        & 57.53                     & 60.46        & 53.6         & 46.23                    & \textbf{48.56 }       & 43.08        \\
SimGCD \cite{gcd3}               & 29.35                            & 17.3         & \underline{62.12 }      & 23.08                             & 28.26        & 16.32        & 51.98                           & 74.44        & 33.26        & 46.29                           & 48.96        & 43.17        & 34.26                     & 44.91        & 20.35        & 24.84                    & 31.88        & 5.68         \\
CDAD-Net \cite{rongali2024cdadnetbridgingdomaingaps}                 & 46.02                            & \underline{45.95}        & 46.21        & 51.71                             & 53.43        & 49.46        & 99.04                           & 99.21        & 98.9         & 76.61                           & 76.97        & 76.19        & 56.78                     & 56.67        & \underline{56.93}        & \textbf{46.65}                    & 46.15        & 48.01        \\
GCD With Synthetic   & 45.78                            & 36.71        & 58.01        & 54.84                             & 73.47        & 38.57        & 82.6                            & 66.29        & 99.39        & 79                              & 86.84        & 72.02        & 53.56                     & 67.93        & 41.01        & 44.18                    & 47.78        & 39.32        \\
CDAD-Net with Synthetic  & 43.09                            & 42.53        & 44.6         & 49.45                             & 59.31        & 36.58        & 99.16                           & 99.21        & 99.12        & 65.38                           & 62.83        & 68.36        & 42.92                     & 41.97        & 44.15        & 41.51                    & 43.79        & 35.32        \\
Hyp-GCD \cite{Liu2025HypCD} & 45.78 & 36.71 & 58.01 & 54.84 & \textbf{73.47} & 38.57 & 82.6 & 66.29 & 99.39 & 79 & 86.84 & 72.02 & 53.56 & 67.93 & 41.01 & 44.18 & \underline{47.78} & 39.32 \\

Hyp-SelfEx \cite{gcd8} & 45.03 & 36.57 & 56.45 & \underline{62.33} & 71.28 & 54.52 & 98.37 & 98.17 & 98.57 & 88.21 & 91.38 & 85.38 & 59.26 & \textbf{72.19 }& 47.96 & 39.24 & 37.24 & 41.94\\

Hyp-SimGCD & 22.83 & 3.33 & \textbf{75.88} & 27.78 & 32.39 & 21.77 & 57.39 & 72.03 & 45.2 & 44.57 & 63.14 & 22.88 & 35.09 & 43.11 & 24.63 & 23.82 & 12.99 & \underline{53.31}\\

Hyp-{\prevmodel} \cite{dg2net} & \underline{46.79} & 43.75 & 51.26 & \textbf{63.65} & 61.78 & \textbf{65.65} & \textbf{99.61 }& \underline{99.6} & \textbf{99.63} & \underline{89.87} & \underline{93.25} & \underline{85.79} & 57.36 & 60.41 & 54.1 & \underline{46.58} & 43.32 & 51.35\\

\prevmodel         &\underline{46.79}                            & 38.13        & 58.49        & 57.96                             & \underline{73.38}        & 44.48        & \underline{99.34}                           & \textbf{99.7}         & 98.97        & 86.67                           & 91.87        & 82.04        & \textbf{62.97   }                  & \underline{71.18}        & 55.8         & 45.72                    & 36.53        & \textbf{58.13 }       \\
\hline
{\ourmodel}(Ours) & \textbf{48.99} & 43.43 & 57.16 & 62.29 & 59.79 & \underline{64.95} & 99.19 & 99 & \underline{99.48} & \textbf{90.7 }& \textbf{93.61} & \textbf{87.19} & \underline{61.84} & 60.04 &\textbf{ 63.76} & 46.56 & 43.92 & 50.44\\
\bottomrule
\end{tabular}}

\vspace{10px}

\resizebox{0.98\textwidth}{!}{%
\begin{tabular}{>{\bfseries}lccc|ccc|ccc|ccc|ccc|ccc}
\midrule
 \multirow{2}{*}{\textbf{Methods}} & \multicolumn{3}{c|}{\textbf{Sketch $\rightarrow$ Art Painting}} & \multicolumn{3}{c|}{\textbf{Sketch $\rightarrow$ Cartoon}} & \multicolumn{3}{c|}{\textbf{Sketch $\rightarrow$ Photo}} & \multicolumn{3}{c|}{\textbf{Cartoon $\rightarrow$ Art Painting}} & \multicolumn{3}{c|}{\textbf{Cartoon $\rightarrow$ Sketch}} & \multicolumn{3}{c}{\textbf{Cartoon $\rightarrow$ Photo}} \\
 \cline{2-4} \cline{5-7} \cline{8-10} \cline{11-13} \cline{14-16} \cline{17-19}
 & \texttt{All} & \texttt{Old} & \texttt{New} & \texttt{All} & \texttt{Old} & \texttt{New} & \texttt{All} & \texttt{Old} & \texttt{New} & \texttt{All} & \texttt{Old} & \texttt{New} & \texttt{All} & \texttt{Old} & \texttt{New} & \texttt{All} & \texttt{Old} & \texttt{New} \\
\midrule
ViT \cite{vit}                  & 23.93                 & 26.53    & 21.61    & 40.61           & 58.92    & 24.62    & 33.29        & 33.88    & 32.69    & 38.09                  & 47.36 & 29.82 & 33.57           & 35.67 & 30.74 & 41.38         & 39.08 & 43.74  \\
GCD \cite{gcd}                  & 33.25                 & 39.09    & 25.43    & 40.89           & 48.14    & 31.17    & 46.86        & 59.28    & 28.22    & 58.15                  & 78.52 & 30.86 & 36              & \underline{44.83} & 24.04 & 75.75         & 85.88 & 60.55   \\
SimGCD \cite{gcd3}               & 21.19                 & 31.91    & 8.67     & 23.17           & 36.77    & 5.4      & 34.22        & 27.46    & 40.8     & 38.38                  & 42.07 & 34.07 & 34.84           & 33.94 & 37.31 & 53.05         & 45.85 & 59.06  \\
CDAD-Net \cite{rongali2024cdadnetbridgingdomaingaps}                 & 87.99                 & 84.32    & \textbf{92.28}    & 51.88           & 51.77    & 52.02    & 99.04        & \underline{99.21}    & 98.9     & 73.05                  & 76.88 & 68.57 & 41.84           & 42.71 & 39.49 & 99.22         & \underline{99.47} & 99.01   \\
GCD With Synthetic~  & 82.15                 & 85.13    & 79.5     & 44.3            & 48.22    & 40.89    & \textbf{99.49}        & \textbf{99.76}    & 99.21    & 63.01                  & 63.73 & 62.37 & 35.66           & 29.95 & 43.36 & 99.43         & \underline{99.47} & 99.39  \\
CDAD-Net with Synthetic  & 61.91                 & 69.45    & 53.12    & 48.59           & 53.13    & 42.67    & 68.44        & 63.5     & 72.56    & 67.24                  & 65.28 & 69.52 & 42.05           & 39.61 & 48.67 & 99.34         & \underline{99.47} & 99.23   \\
Hyp-GCD \cite{Liu2025HypCD} & 82.15 & 85.13 & 79.5 & 44.3 & 48.22 & 40.89 & 99.49 & 99.76 & 99.21 & 63.01 & 63.73 & 62.37 & 35.66 & 29.95 & 43.36 & \textbf{99.43 }&\underline{ 99.47} & \underline{99.39} \\

Hyp-SelfEx \cite{gcd8} & 88.77 & \underline{94.03} & 84.09 & 57.11 & \textbf{73.52} & 42.76 & 98.16 & 97.82 & 98.51 & 89.11 & 89.1 & \underline{89.12} & 45.31 & 37.02 & \underline{56.51} & 98.34 & 98.06 & 98.64\\

Hyp-SimGCD & 22.82 & 22.77 & 22.87 & 32.26 & 45.48 & 15 & 27.68 & 7.79 & 44.25 & 23.62 & 35.79 & 9.4 & 30.11 & 32.22 & 24.38 & 33.9 & 23.76 & 42.35\\

Hyp-{\prevmodel} \cite{dg2net} & \underline{89.6} & 92.63 & 85.95 & \underline{57.47} & 60.91 & \underline{53.79} & \underline{99.28} & 99.15 & \underline{99.48} & \textbf{93.26} & \textbf{95.89} & \textbf{90.1} & 46.08 & 42.94 & 50.69 & \underline{99.34} & \textbf{99.5 }& 99.11\\

\prevmodel         &88.75                 & 93.52    & 84.49    & 56.76           & \underline{72.14}    & 43.33    & 99.13        & 98.7     & \textbf{99.57 }   & \underline{90.77}                  &  \underline{93.37} &   88.46 &  \textbf{ 49.2}            & 43.18 & \textbf{ 57.33} & 95.57         & 91.62 &  \textbf{99.64 } \\
\hline
{\ourmodel}(Ours) & \textbf{92.36} & \textbf{94.59} & \underline{89.67} &\textbf{ 65.7} & 68.39 & \textbf{62.83} & 98.74 & 99.05 & 98.29 & 88.57 & 92.14 & 84.28 & \underline{47.07} & \textbf{53.12} & 38.19 & 98.86 & 99.45 & 97.99 \\
\bottomrule
\end{tabular}}
\vspace{10px}
\caption{Detailed comparison of our proposed {\ourmodel} on DG-GCD with respect to referred literature for PACS Dataset.}
\label{tab:PACS_comparison2}
\end{table}

\begin{table}[!htbp]
\centering
\resizebox{0.98\textwidth}{!}{%
\begin{tabular}{>{\bfseries}lccc|ccc|ccc|ccc|ccc|ccc}
\toprule
 \multicolumn{19}{c}{\textbf{Office-Home}} \\ 
\midrule
 \multirow{2}{*}{\textbf{Methods}} & \multicolumn{3}{c|}{\textbf{Art $\rightarrow$ Clipart}} & \multicolumn{3}{c|}{\textbf{Art $\rightarrow$ Product}} & \multicolumn{3}{c|}{\textbf{Art $\rightarrow$ Real World}} & \multicolumn{3}{c|}{\textbf{Clipart $\rightarrow$ Art}} & \multicolumn{3}{c|}{\textbf{Clipart $\rightarrow$ Real World}} & \multicolumn{3}{c}{\textbf{Clipart $\rightarrow$ Product}} \\
 \cline{2-4} \cline{5-7} \cline{8-10} \cline{11-13} \cline{14-16} \cline{17-19}
 & \texttt{All} & \texttt{Old} & \texttt{New} & \texttt{All} & \texttt{Old} & \texttt{New} & \texttt{All} & \texttt{Old} & \texttt{New} & \texttt{All} & \texttt{Old} & \texttt{New} & \texttt{All} & \texttt{Old} & \texttt{New} & \texttt{All} & \texttt{Old} & \texttt{New} \\
\midrule
ViT \cite{vit}                  & 18.88                 & 20.86 & 15.79 & 30.34             & 35.42 & 21.83 & 29.52                 & 32.76 & 24.85 & 14.96             & 15.6  & 14.12 & 18.59                     & 20.12 & 16.4  & 30.39                 & 32.51 & 26.84 \\
GCD \cite{gcd}                  & 31.65                 & 32.11 & 30.93 & 63.18             & 64.35 & 61.22 & 63.85                 & 66.56 & 59.96 & 51.96             & 52.7  & 51    & 62.62                     & 65.29 & 58.79 & 60.59                 & 67.13 & 49.61 \\
SimGCD \cite{gcd3}               & 24.54                 &34.35& 8.09  & 41.95             & 57.92 & 13.54 & 46.78                 & 65.54 & 14.73 & 31.11             & 39.56 & 11.88 & 25.66                     & 37.66 & 5.15  & 28.88                 & 41.38 & 12.96 \\
CDAD-Net \cite{rongali2024cdadnetbridgingdomaingaps}             & 30.95 & 33.65 & 26.43           & 64.99 & 68.04 & 59.32                   & 67.5  &  \textbf{ 70.89} & 61.72               & 53.36 &  56.05 & 47.23              & 64.7  & \underline{ 69.4}  & 55.25                  &   \textbf{67.02} &  \underline{68.8}  & \textbf{63.7 }                   \\
GCD With Synthetic~  & 29.86                 & 31.04 & 28.02 & 57.92             & 63.12 & 49.19 & 59.47                 & 59.59 & 59.29 & 53.3              & 52.84 &  \underline{53.89} & 61.46                     & 58.27 & 66.06 & 63.84                 & 64.04 & \underline{63.51} \\
CDAD-Net with Synthetic  &  \textbf{31.97} &  \textbf{35.1}  & 26.71           &  65.39 & \underline{  68.94 }& 62.51                   &   \textbf{67.83} &  \underline{70.87} & 62.64               &   \underline{53.51} &   \underline{56.65} & 46.37              &   \underline{66.97} &  \textbf{ 69.76} & 62.2                   & 61.4  & 65.55 & 57.4  \\
Hyp-GCD \cite{Liu2025HypCD} &28.99 & 29.1 & 29.28 & \underline{67.3} & 66.38 & \underline{64.99} & 63.79 & 63.14 & 62.05 & 45.88 & 43.92 & 40.44 & 60.27 & 58.07 & 54.36 & 63.33 & 62.02 & 60.04 \\

Hyp-SelfEx \cite{gcd8} & 30.42 & 28.84 & \underline{32.89} & 64.26 & 67.7 & 58.5 & 64.53 & 60.91 & \underline{69.73} & 49.8 & 49.18 & 50.61 & 63.78 & 59.62 & 69.76 & 64.23 & 66.35 & 60.67\\

Hyp-SimGCD \cite{gcd3} & 23.34 & \underline{34.69} & 4.31 & 38.5 & 54.96 & 7.84 & 46.37 & 70.58 & 5.03 & 21.82 & 29.42 & 4.55 & 25.98 & 37.79 & 5.83 & 29.93 & 41.55 & 8.28
\\

Hyp-{\prevmodel} \cite{dg2net} & 27.89 & 26.84 & 29.68 & 66.02 & \textbf{70.03 }& 59.95 & 63.16 & 62.55 & 64.2 & 45.94 & 47.43 & 43.3 & 55.96 & 56.65 & 54.79 & 62.71 & 67.75 & 55.09
 \\
\prevmodel         & 31.51                 & 31.96 & 30.81 &   \textbf{67.46}             &  68.73 &  \textbf{ 65.32} & 64.45                 & 60.25 &  \textbf{70.48} & 50.76             & 48.76 & 53.36 & 64.77                     & 60.58 & \textbf{ 70.79} & 65.34                 & 67.48 & 61.76 \\
\hline
{\ourmodel}(Ours) & \underline{31.91} & 30.46 & \textbf{34.18} & 64.81 & 68.19 & 59.14 & \underline{66.28} & 65.6 & 67.26 &\textbf{ 58.65 }& \textbf{60.68} & \textbf{56.02 }& \textbf{67.82} & 66.36 & \underline{69.92} &\underline{ 66.6} & \textbf{70.19 }& 60.59 \\
\bottomrule
\end{tabular}}

\vspace{10px}

\resizebox{0.98\textwidth}{!}{%
\begin{tabular}{>{\bfseries}lccc|ccc|ccc|ccc|ccc|ccc}
\midrule
 \multirow{2}{*}{\textbf{Methods}} & \multicolumn{3}{c|}{\textbf{Product $\rightarrow$ Art}} & \multicolumn{3}{c|}{\textbf{Product $\rightarrow$ Real World}} & \multicolumn{3}{c|}{\textbf{Product $\rightarrow$ Clipart}} & \multicolumn{3}{c|}{\textbf{Real World $\rightarrow$ Art}} & \multicolumn{3}{c|}{\textbf{Real World $\rightarrow$ Product}} & \multicolumn{3}{c}{\textbf{Real World $\rightarrow$ Clipart}} \\
 \cline{2-4} \cline{5-7} \cline{8-10} \cline{11-13} \cline{14-16} \cline{17-19}
 & \texttt{All} & \texttt{Old} & \texttt{New} & \texttt{All} & \texttt{Old} & \texttt{New} & \texttt{All} & \texttt{Old} & \texttt{New} & \texttt{All} & \texttt{Old} & \texttt{New} & \texttt{All} & \texttt{Old} & \texttt{New} & \texttt{All} & \texttt{Old} & \texttt{New} \\
\midrule
ViT \cite{vit}                  & 23.2                 & 24.64 & 21.33 & 31.21               & 35.45 & 25.13 & 19.27           & 20.52 & 17.31 & 32.22          & 35.79 & 27.58 & 44.67              & 52.21 & 32.03 & 20.8               & 23.71 & 16.26 \\
GCD \cite{gcd}                  & 50.27                & 48.18 & 52.99 & 65.07               & 63.09 & \underline{67.91} & 29.08           & 29.22 & 28.87 & 54.26          & 54.05 & 54.55 &   69.04              &  72.76 &  \underline{62.79} & 31.04              & 34.93 & 24.97 \\
SimGCD \cite{gcd3}               & 38.28                & 50.42 & 10.66 & 48.36               & 67.07 & 16.41 & 22.45           & 32.37 & 11.34 & 48.95          &  \underline{66.79} & 8.36  & 57.19                   & 69.23       & 44.15       & 21.7               & 31.46 & 5.33  \\
CDAD-Net \cite{rongali2024cdadnetbridgingdomaingaps}                & 50.1  & 52.43 & 44.67           & 66.47 &   \textbf{72.13} & 56.81                   & 31.36 & \underline{ 34.6}  & 25.94               &  54.68 & 58.07 & 46.96              & 61.39 & 64.79 & 55.06                  & \underline{31.78} &  \textbf{36.02} & 24.69                   \\

GCD With Synthetic~  & 49.18                & 46.54 & 52.61 & 63.4                & 59.67 & \textbf{ 68.77} & 28.43           & 27.72 & 29.55 & 51.71          & 61.55 & 38.91 & 61.14              & 65.34 & 54.1  & 26.38              & 28.11 & 23.68 \\
CDAD-Net with Synthetic &   \underline{54.12} &   \underline{57.67} & 46.04           &  66.97 &  70.2  & 61.46                   &   \underline{32.34} &  \textbf{ 35.13} & 28.68               & 53.72 & 56.89 & 46.5               & 56.47 & 62.33 & 45.62                  & 31.19 & 33.67 & 27.02  \\

Hyp-GCD \cite{Liu2025HypCD} &45.43 & 44.46 & 42.73 & 61.63 & 59.64 & 56.3 & 27.15 & 28.35 & 30.39 & 45.82 & 45.28 & 44.33 & 64.34 & 63.74 & \textbf{62.84 }& 27.66 & 28.25 & 29.25 \\

Hyp-SelfEx \cite{gcd8} & 51.97 & 48.53 &\textbf{ 56.44} & 66.06 & 65.07 & 67.49 & 29.01 & 28.93 & 29.13 & \underline{55.43} & 53.52 & \textbf{57.91} & 65.3 & \underline{73.33} & 51.83 & 30.12 & 29.83 & \underline{30.57} \\

Hyp-SimGCD \cite{gcd3} & 33.29 & 45.26 & 6.07 & 40.77 & 60.68 & 6.75 & 16.62 & 23.46 & 5.15 & 46.39 & 63.59 & 7.24 & 44.88 & 66.32 & 4.96 & 20.98 & 30.97 & 4.24
 \\

Hyp-{\prevmodel} \cite{dg2net} & 44.56 & 41.99 & 49.14 & 61.2 & 62.49 & 59.04 & 26.45 & 26.55 & 26.28 & 46.99 & 48.71 & 43.93 & 63.96 & 64.73 & 62.78 & 27.94 & 27.8 & 28.17
 \\

\prevmodel         & 52.45                & 50.51 &   \underline{54.98} &   \underline{67.87}               & 69.88 & 64.97 & 30.71           & 30.05 & \underline{ 31.75} & 52.31          & 49.42 &  \underline{56.07} &  \underline{67.37}              & 71.65 & 60.19 & 31.28              & 31.13 & \textbf{31.51} \\
\hline
{\ourmodel}(Ours) & \textbf{60.12} &\textbf{ 65.52} & 53.08 & \textbf{68.95 }& \underline{71.13} & 65.81 & \textbf{32.63 }& 32.21 & \textbf{33.3} & \textbf{61.54} & \textbf{71.14 }& 49.05 & \textbf{69.22} & \textbf{73.41} & 62.18 & \textbf{32.79} & \underline{35.85} & 28.02 \\
\bottomrule
\end{tabular}}
\vspace{10px}
\caption{Detailed comparison of our proposed {\ourmodel} on DG-GCD with respect to referred literature for Office-Home Dataset}
\label{tab:office_home_comparison2}
\end{table}
\newpage
\begin{table}[!htbp]
\centering
\resizebox{0.98\textwidth}{!}{%
\begin{tabular}{>{\bfseries}lccc|ccc|ccc|ccc|ccc}
\toprule
 \multicolumn{16}{c}{\textbf{DomainNet}} \\ 
\midrule
 \multirow{2}{*}{\textbf{Methods}} & \multicolumn{3}{c|}{\textbf{Sketch $\rightarrow$ Real}} & \multicolumn{3}{c|}{\textbf{Sketch $\rightarrow$ Quickdraw}} & \multicolumn{3}{c|}{\textbf{Sketch $\rightarrow$ Infograph}} & \multicolumn{3}{c|}{\textbf{Sketch $\rightarrow$ Painting}} & \multicolumn{3}{c}{\textbf{Sketch $\rightarrow$ Clipart}} \\
 \cline{2-4} \cline{5-7} \cline{8-10} \cline{11-13} \cline{14-16}
 & \texttt{All} & \texttt{Old} & \texttt{New} & \texttt{All} & \texttt{Old} & \texttt{New} & \texttt{All} & \texttt{Old} & \texttt{New} & \texttt{All} & \texttt{Old} & \texttt{New} & \texttt{All} & \texttt{Old} & \texttt{New} \\
\midrule
ViT \cite{vit}                  & 47.17              & 47.92 & 44.95 & 12.13              & 12.1  & 12.21 & 11.99              & 12.68 & 10.28 & 30.95             & 33.02 & 25.75 &   \underline{32.64}             &  \underline{34.29} &  \underline{28.64} \\
GCD \cite{gcd}                  & 51.13              & 51.88 & 48.92 &   \textbf{16.08}              &   \underline{15.65} &  \textbf{ 17.2}  & 12.6               & 12.57 &  \underline{12.68} & 35.25             & 35.96 &  \underline{33.46} &  31.22             & 30.85 &   32.1  \\
SimGCD \cite{gcd3}               & 3.11               & 3.47  & 2.32  & 2.31               & 2.4   & 2.1   & 3.16               & 2.27  & 5.24  & 4.1               & 2.57  & 5.62  & 3.02              & 2.3   & 4.07  \\
CDAD-Net \cite{rongali2024cdadnetbridgingdomaingaps}                 & 48.21              & 47.7  &  49.77 & 12.27              & 11.52 & 14.24 & 12.07              & 12.69 & 11.34 & 35.47             & 36.39 & 32.86 & 18.63             & 17.52 & 20.39 \\
GCD With Synthetic~  & 53                 & 51.71 & 47.64 & 13.71              &  13.79 & 13.99 & 12.24              & 11.99 & 11.37 & 35.43             & 34.12 & 30.83 & 22.49             & 22.2  & 21.49 \\
CDAD-Net with Synthetic  & 47.11              & 46.09 & 49.4  & 12.75              & 13.1  & 14.05 & 12.52              & 13.04 & 11.92 & 35.87             & 36.73 & 33.35 & 18.99             & 17.68 & 21.07 \\

Hyp-GCD \cite{Liu2025HypCD}  & 47.08 & 49.78 & \textbf{56.61} & 12.88 & 13.37 & 14.66 & 11.04 & 11.32 & 11.95 & 31.33 & 31.84 & 33.1 & 17.87 & 18.85 & 21.23
 \\

Hyp-SelfEx \cite{gcd8}  & 52.34 & 53.02 & 50.35 & 12.91 & 12.76 & 13.32 & 12.27 & 12.42 & 11.89 & 33.82 & 34.49 & 32.14 & 19.63 & 19.11 & 20.87
 \\

Hyp-SimGCD \cite{gcd3}  & 0.54 & 0.64 & 0.25 & 0.29 & 0.4 & 0.001 & 1.52 & 0.001 & 3.3 & 1.19 & 0.001 & 4.54 & 1.07 & 0.76 & 1.55

 \\

Hyp-{\prevmodel} \cite{dg2net} &36.33 & 34.13 & 41.88 & 13.51 & 12.96 & 14.96 & 9.74 & 9.3 & 10.73 & 24.22 & 23.33 & 26.41 & 14.17 & 14.17 & 14.16
 \\

\prevmodel         &   \underline{53.67}              &   \textbf{55.48 }& 48.35 & \underline{15.9}               & \textbf{16  }  &  15.63 & \underline{14.63}              &  \underline{15.66} & 12.06 &  \underline{37.44}             &   \underline{39.53} & 32.19 & 30.47             & 32.89 & 24.58 \\
\hline
{\ourmodel}(Ours) & \textbf{55.27} & \underline{54.89} & \underline{56.24} & 14.61 & 13.83 & \underline{16.68} & \textbf{15.35 }& \textbf{16.3} &\textbf{ 13.2} & \textbf{39.8} & \textbf{41.15} & \textbf{36.49} & \textbf{35.74} & \textbf{38.61} & \textbf{28.8}
 \\
\bottomrule
\end{tabular}}

\vspace{10px}

\resizebox{0.98\textwidth}{!}{%
\begin{tabular}{>{\bfseries}lccc|ccc|ccc|ccc|ccc}
\midrule
 \multirow{2}{*}{\textbf{Methods}} & \multicolumn{3}{c|}{\textbf{Clipart $\rightarrow$ Infograph}} & \multicolumn{3}{c|}{\textbf{Clipart $\rightarrow$ Quickdraw}} & \multicolumn{3}{c|}{\textbf{Clipart $\rightarrow$ Sketch}} & \multicolumn{3}{c|}{\textbf{Clipart $\rightarrow$ Real}} & \multicolumn{3}{c}{\textbf{Clipart $\rightarrow$ Painting}} \\
 \cline{2-4} \cline{5-7} \cline{8-10} \cline{11-13} \cline{14-16}
 & \texttt{All} & \texttt{Old} & \texttt{New} & \texttt{All} & \texttt{Old} & \texttt{New} & \texttt{All} & \texttt{Old} & \texttt{New} & \texttt{All} & \texttt{Old} & \texttt{New} & \texttt{All} & \texttt{Old} & \texttt{New} \\
\midrule
ViT \cite{vit}                  & 12.18                   & 12.64 & 11.03 & 12.13                   & 12.1  & 12.21 & 24.76                & 26.24 &   21.27 & 44.14              & 45.43 & 40.34 & 26.76                  & 28.7  & 21.91 \\
GCD \cite{gcd}                  & 14.03                   & 14.64 & 12.49 &   14.94                   &  14.67 &  15.65 & 25.33                & 27.68 & 19.78 & 53.23              & 55.48 & 46.62 & 34.82                  & 36.82 & 29.83 \\
SimGCD \cite{gcd3}               & 2.03                    & 0.4   & 3.94  & 0.5                     & 0.3   & 1     & 1                    & 0.02  & 3.842 & 1.64               & 1.07  & 2.42  & 2.07                   & 2.05  & 2.13  \\
CDAD-Net \cite{rongali2024cdadnetbridgingdomaingaps}                 & 12.79                   & 12.96 & \underline{12.87} & 12.06                   & 11.59 & 12.78 & 19                   & 19.17 & 18.76 & 47.06              & 44.62 & 49.2  & 34.45                  & 36.02 & 32.85 \\
GCD With Synthetic~  & 11.46                   & 12.03 & 10.04 & 12.68                   & 12.57 & 12.95 & 18.74                & 20.54 & 14.47 & 50.11              & 52.26 & 43.79 & 32.67                  & 34.91 & 27.06 \\
CDAD-Net with Synthetic  & 13                      & 13.37 & 12.56 & 12.07                   & 11.76 & 12.89 & 17.46                & 18.03 & 16.67 & 48.25              & 47.51 & 49.6  & 33.23                  & 32.79 &  34.2  \\

Hyp-GCD \cite{Liu2025HypCD}  &11.52 & 11.6 & 11.79 & 12.98 & 13.48 & 14.79 & 15.33 & 15.84 & 17.11 & 44.82 & 46.52 & 50.82 & 29.4 & 29.92 & 31.21
 \\

Hyp-SelfEx \cite{gcd8}  & 11.6 & 11.7 & 11.35 & 12.89 & 12.61 & 13.64 & 17.1 & 17.92 & 15.16 & 50.55 & 51.11 & 48.92 & 33.46 & 34.31 & 31.34
 \\

Hyp-SimGCD \cite{gcd3}  & 1.57 & 0.03 & 3.38 & 0.29 & 0.4 & 0.001 & 1.06 & 0.044 & 2.46 & 0.73 & 0.77 & 0.61 & 1.24 & 0.214 & 4.15

 \\

Hyp-{\prevmodel} \cite{dg2net} & 11.95 & 11.81 & 12.27 & 14.1 & 13.43 & 15.88 & 16.21 & 15.76 & 17.32 & 48.04 & 45.92 & 53.42 & 31.28 & 30.73 & 32.65
 \\

\prevmodel         &   \textbf{15.81}                   &   \textbf{17.09} & 12.63 & \textbf{14.53  }                 & 14.14 & 15.58 & 26.86                & 29.49 & 20.64 &   54.54              &  56.03 & 50.17 & 36.81                  &  38.87 & 31.67 \\
\hline
{\ourmodel}(Ours) & \underline{15.7} &\underline{16.34} & \textbf{14.25} & \ 14.35 & 13.83 & 15.72 & 26.6 & 27.56 & 24.24 & 55.04 & 54.49 & 56.42 & 39.08 & 40.13 & 36.5
\\
\bottomrule
\end{tabular}}

\vspace{10px}
\resizebox{0.98\textwidth}{!}{%
\begin{tabular}{>{\bfseries}lccc|ccc|ccc|ccc|ccc}
\midrule
 \multirow{2}{*}{\textbf{Methods}} & \multicolumn{3}{c|}{\textbf{Painting $\rightarrow$ Infograph}} & \multicolumn{3}{c|}{\textbf{Painting $\rightarrow$ Quickdraw}} & \multicolumn{3}{c|}{\textbf{Painting $\rightarrow$ Sketch}} & \multicolumn{3}{c|}{\textbf{Painting $\rightarrow$ Real}} & \multicolumn{3}{c}{\textbf{Painting $\rightarrow$ Clipart}} \\
 \cline{2-4} \cline{5-7} \cline{8-10} \cline{11-13} \cline{14-16}
 & \texttt{All} & \texttt{Old} & \texttt{New} & \texttt{All} & \texttt{Old} & \texttt{New} & \texttt{All} & \texttt{Old} & \texttt{New} & \texttt{All} & \texttt{Old} & \texttt{New} & \texttt{All} & \texttt{Old} & \texttt{New} \\
\midrule
ViT \cite{vit}                  & 12.2                     & 13.1  & 9.94  & 12.13                    & 12.1  & 12.21 & 23                    & 24.78 & 18.79 & 51.53               & 54.16 & 43.8  & 26.57                  & 28.08 & 22.92 \\
GCD \cite{gcd}                  & 12.87                    & 12.67 &  13.37 & 10.74                    & 10.56 & 11.21 & 21.49                 & 22.26 &   19.68 & 52.12               & 51.86 &   52.86 & 25.32                  & 24.79 &   26.6  \\
SimGCD \cite{gcd3}               & 3.2                      & 2.6   & 3.8   & 3.5                      & 2.32  & 4.65  & 4.23                  & 3.56  & 4.86  & 4.2                 & 3.52  & 5     & 4.49                   & 3.6   & 5.23  \\
CDAD-Net \cite{rongali2024cdadnetbridgingdomaingaps}                 & 11.65                    & 12.49 & 10.66 & 11.98                    & 11.2  & 12.44 & 17.11                 & 17.68 & 16.32 & 49.04               & 48.63 & 50.27 & 20.06                  & 19.74 & 20.57 \\
GCD With Synthetic  & 10.86                    & 10.56 & 9.84  & 11.81                    & 11.8  & 11.77 & 17.26                 & 16.25 & 13.83 & 49.1                & 47.3  & 42.04 & 19.3                   & 19.45 & 18.04 \\
CDAD-Net with Synthetic  & 11.53                    & 12.32 & 10.59 & 11.86                    & 10.71 & 12.32 & 17.29                 & 18.45 & 15.7  & 48.4                & 50.23 & 49.7  & 17.44                  & 15.92 & 19.86 \\

Hyp-GCD \cite{Liu2025HypCD}  & 12.12 & 12.38 & 12.96 & 12.32 & 13.05 & 14.97 & 16.78 & 17.54 & 19.42 & 48.4 & 50.39 & 55.41 & 19.34 & 20.23 & 22.39
\\

Hyp-SelfEx \cite{gcd8}  & 12.43 & 12.24 & 12.89 & 12.98 & 12.76 & 13.55 & 18.28 & 18.72 & 17.22 & 51 & 51.59 & 49.27 & 20.42 & 20.11 & 21.18
 \\
Hyp-SimGCD \cite{gcd3}  & 1.52 & 0.002 & 3.3 & 0.35 & 0.43 & 0.17 & 1.034 & 0.002 & 2.46 & 0.48 & 0.63 & 0.05 & 0.96 & 0.001 & 2.5
 \\

Hyp-{\prevmodel} \cite{dg2net} & 12.94 & 12.65 & 13.59 & 14.14 & 13.45 & 15.96 & 17.54 & 16.84 & 19.25 & 50.59 & 49.07 & 54.45 & 19.95 & 19.31 & 21.49
 \\

\prevmodel         &   15.71                    &   16.72 & 13.22 &  12.9                     &  12.66 &  13.53 & 23.14                 & 25.23 & 18.19 & 55.07               & 56.97 & 49.5  &   27.6                   &  29.07 & 24.03 \\
\hline
{\ourmodel}(Ours) & 15.98 & 17.15 & 13.36 & 13.03 & 12.62 & 14.1 & 25.78 & 27.37 & 21.9 & 57.1 & 58.76 & 52.9 & 34.29 & 37.99 & 25.34
 \\

\bottomrule
\end{tabular}}
\vspace{10px}
\caption{Detailed comparison of our proposed {\ourmodel} on DG-GCD with respect to referred literature for DomainNet Dataset}
\label{tab:domain_net_comparison3}
\end{table}

\newpage
\subsection{Extended Tables for additional Baselines}

\begin{table*}[htbp]
\centering
\vspace{-5px}
\caption{
\textbf{Extended clustering accuracy comparison} (\%) including Hyp-SimGCD, {\ourmodel}, and Upper-Bound (UB) on PACS, Office-Home, and DomainNet. (\textbf{Bold}: best, \underline{underline}: second best).
}
\label{tab:dg-gcd-supp-ext}
\scriptsize
\setlength{\tabcolsep}{3pt}
\renewcommand{\arraystretch}{1.05}
\rowcolors{3}{gray!5}{white}
\begin{tabular}{lccccccccccccc}
\toprule
\textbf{Method} & \textbf{Venue} & \multicolumn{3}{c}{\textbf{PACS}} 
                               & \multicolumn{3}{c}{\textbf{Office-Home}} 
                               & \multicolumn{3}{c}{\textbf{DomainNet}} 
                               & \multicolumn{3}{c}{\textbf{Avg.}} \\
\cmidrule(lr){3-5}\cmidrule(lr){6-8}\cmidrule(lr){9-11}\cmidrule(lr){12-14}
 & & All & Old & New & All & Old & New & All & Old & New & All & Old & New \\
\midrule
\textcolor{purple!60!black}{Hyp-SelfEx} \cite{gcd8} & ECCV'24 
  & \underline{72.44} & \underline{74.70} & \underline{71.20}  
  & \underline{52.91} & \underline{52.65} & \underline{52.96}  
  & \underline{29.30} & \underline{30.45} & \underline{26.37}  
  & \underline{51.55} & \underline{52.60} & \underline{50.18} \\
\midrule
\textcolor{purple!60!black}{Hyp-SimGCD} \cite{gcd3} & ICCV'23
  & 31.82 & 32.90 & 33.49  
  & 32.41 & 46.61 & 5.85  
  & 0.92	& 0.29	&1.91
  & 21.72 & 26.6 & 13.75  \\
\midrule
\textcolor{red!80!black}{\textbf{HiDISC (Ours)}} (2 Synth) & -- 
  & \textbf{75.07} & \textbf{75.54} & \textbf{74.52}  
  & \textbf{56.78} & \textbf{59.23} & \textbf{53.21}  
  & \textbf{30.51} & \textbf{31.40} & \textbf{28.41}  
  & \textbf{54.12} & \textbf{55.39} & \textbf{52.05} \\
\midrule
\textcolor{gray!70!black}{CDAD-Net (Upper-Bound})\cite{rongali2024cdadnetbridgingdomaingaps} & --  
  & 83.25 & 87.58 & 77.35  
  & 67.55 & 72.42 & 63.44  
  & 70.28 & 76.46 & 65.19  
  & 73.69 & 78.82 & 68.66 \\
\bottomrule
\end{tabular}
\end{table*}

\textbf{Comparison with Hyperbolic Prototype Methods}: HPDR ~\cite{hu2024rethinking} address domain generalization for face anti-spoofing by leveraging hyperbolic prototypes to improve robustness across visual domains. While effective in this binary closed-set setting, their method (HPDR) does not extend to the more challenging DG-GCD problem, which requires discovering novel categories in unseen domains without target supervision. To clarify this distinction, we re-implemented their embedding learning component within our framework and evaluated it under the DG-GCD protocol. As shown in Table~\ref{tab:hpdr_comparison}, HPDR performs significantly worse than our proposed {\ourmodel} on the Office-Home benchmark, highlighting the importance of discovery-oriented objectives and domain-aware augmentations.

\begin{table}[h]
  \centering
  \begin{tabular}{l|ccc}
  \toprule
  \textbf{Method}      & \textbf{All ↑} & \textbf{Old ↑} & \textbf{New ↑} \\
  \midrule
  HPDR~\cite{hu2024rethinking}   & 15.05 & 16.55 & 12.58 \\
  {\ourmodel}     & 56.78 & 59.23 & 53.21 \\
  \bottomrule
  \end{tabular}
  \vspace{10px}
  \caption{\small
    Comparison of Hu et al.'s method (HPDR) and {\ourmodel} under the DG-GCD setting on Office-Home.}
  \label{tab:hpdr_comparison}
\end{table}

The performance gap can be attributed to three factors: (i) HPDR is designed for binary closed-set classification and lacks mechanisms to create prototypes for unseen classes; (ii) it optimizes closed-set objectives without clustering or discovery losses, yielding near-random novel-class assignments; and (iii) it does not incorporate domain-generalization strategies. In contrast, HiDISC couples prototype anchoring with domain-generalization regularizers and discovery losses, enabling robust alignment and clustering across unseen domains and categories.

\end{document}